\documentclass{article}

\usepackage[accepted]{icml2024}
\usepackage{soul}
\def\title{\model{}: Unlocking the Potential of Transformers in Time Series Forecasting with Sharpness-Aware Minimization and Channel-Wise Attention}
\def\shorttitle{Unlocking the Potential of Transformers}

\icmltitlerunning{\shorttitle}

\usepackage{amsmath,amsfonts,bm}

\usepackage{mathtools}

\usepackage{bbm}

\usepackage{amsthm}
\newtheorem{thm}{Theorem}[section]

\newtheorem{rmk}[thm]{Remark}

\usepackage{mdframed}
\newmdtheoremenv[topline=false, bottomline=false, leftline=false, rightline=false, backgroundcolor=aliceblue,%
innertopmargin=\topskip, splittopskip=\topskip, skipbelow=\baselineskip, skipabove=\baselineskip]{boxthm}{Theorem}[section]
\newmdtheoremenv[topline=false, bottomline=false, leftline=false, rightline=false, backgroundcolor=aliceblue,%
innertopmargin=\topskip, splittopskip=\topskip, skipbelow=\baselineskip, skipabove=\baselineskip]{boxprop}[boxthm]{Proposition}
\newmdtheoremenv[topline=false, bottomline=false, leftline=false, rightline=false, backgroundcolor=aliceblue,%
innertopmargin=\topskip, splittopskip=\topskip, skipbelow=\baselineskip, skipabove=\baselineskip]{boxexample}[boxthm]{Example}
\newmdtheoremenv[topline=false, bottomline=false, leftline=false, rightline=false, backgroundcolor=aliceblue,%
innertopmargin=\topskip, splittopskip=\topskip, skipbelow=\baselineskip, skipabove=\baselineskip]{boxcor}[boxthm]{Corollary}
\newmdtheoremenv[topline=false, bottomline=false, leftline=false, rightline=false, backgroundcolor=aliceblue,%
innertopmargin=\topskip, splittopskip=\topskip, skipbelow=\baselineskip, skipabove=\baselineskip]{boxlem}[boxthm]{Lemma}
\newmdtheoremenv[topline=false, bottomline=false, leftline=false, rightline=false, backgroundcolor=aliceblue,%
innertopmargin=\topskip, splittopskip=\topskip, skipbelow=\baselineskip, skipabove=\baselineskip]{boxdef}[boxthm]{Definition}

\usepackage{stmaryrd}

\definecolor{rightblue}{RGB}{76,114,176} 
\definecolor{rightorange}{RGB}{221,132,82} 
\definecolor{aliceblue}{rgb}{0.94, 0.97, 1.0} 
\definecolor{darkcerulean}{rgb}{0.03, 0.27, 0.49} 
\definecolor{iris}{rgb}{0.35, 0.31, 0.81} 
\definecolor{carmine}{rgb}{0.59, 0.0, 0.09} 
\definecolor{green(munsell)}{rgb}{0.0, 0.66, 0.47} 
\definecolor{celadon}{rgb}{0.67, 0.88, 0.69} 
\definecolor{bluerow}{rgb}{0.0, 0.53, 0.74} 

\usepackage{xcolor}
\definecolor{color1}{HTML}{D8D5F2}
\definecolor{color2}{HTML}{84A2C6}
\definecolor{color3}{HTML}{3A7376}
\definecolor{color4}{HTML}{0F3222}

\definecolor{orangeintro}{HTML}{EC5800}
\definecolor{blueintro}{HTML}{4682B4}

\usepackage{booktabs}
\usepackage{multirow}
\usepackage{graphicx}
\usepackage{caption} 
\def\thick{0.8}

\usepackage{hyperref}
  \hypersetup{
    final=true,
    plainpages=false,
    pdfstartview=FitV,
    pdftoolbar=true,
    pdfmenubar=true,
    bookmarksopen=true,
    bookmarksnumbered=true,
    breaklinks=true,
    linktocpage=true,
    colorlinks=true,
    linkcolor=red, 
    urlcolor=iris, 
    citecolor=darkcerulean, 
    anchorcolor=black
  }

\usepackage{dirtytalk}
\MakeRobust{\say} 

\usepackage{makecell}

\usepackage{algorithm2e}
\RestyleAlgo{ruled}

\usepackage{subcaption}

\usepackage{tikz}
\usepackage{amsmath, amsfonts}
\usetikzlibrary{arrows.meta}

\usepackage{wrapfig}

\usepackage{amssymb}

\newcommand{\sreparam}{$\sigma$Reparam}
\newcommand{\model}{\texttt{SAMformer}}
\newcommand{\transformer}{\texttt{Transformer}}
\newcommand{\tsmixer}{\texttt{TSMixer}}
\newcommand{\transam}{\texttt{Transformer+SAM}}
\usepackage{mleftright} 

\newcommand{\tr}[1]{\operatorname{Tr}\mleft(#1\mright)}
\newcommand{\rk}[1]{\operatorname{rank}\mleft(#1\mright)}

\newcommand{\softmax}[1]{\operatorname{\mathrm{softmax}}\mleft(#1\mright)}

\DeclareMathOperator*{\argmax}{arg\,max}

\newcommand{\RR}{\mathbb{R}}

\newcommand{\mbf}[1]{\mathbf{#1}}

\newcommand{\bA}{\mathbf{A}}
\newcommand{\bP}{\mathbf{P}}
\newcommand{\bI}{\mathbf{I}}
\newcommand{\bX}{\mathbf{X}}
\newcommand{\bW}{\mathbf{W}}
\newcommand{\bM}{\mathbf{M}}

\newcommand{\bomega}{\bm{\omega}}
\newcommand{\bepsilon}{\bm{\epsilon}}

\begin{document}

\addtocontents{toc}{\protect\setcounter{tocdepth}{0}}

\twocolumn[
\icmltitle{\title}



\icmlsetsymbol{equal}{*}
\begin{icmlauthorlist}
\icmlauthor{Romain Ilbert}{equal,hua,uni}
\icmlauthor{Ambroise Odonnat}{equal,hua}
\icmlauthor{Vasilii Feofanov}{hua}
\icmlauthor{ Aladin Virmaux}{hua}
\icmlauthor{Giuseppe Paolo}{hua}
\icmlauthor{Themis Palpanas}{uni}
\icmlauthor{Ievgen Redko}{hua}
\end{icmlauthorlist}

\icmlaffiliation{hua}{Huawei Noah’s Ark Lab, Paris, France}
\icmlaffiliation{uni}{LIPADE, Paris Descartes University, Paris, France}

\icmlcorrespondingauthor{Romain Ilbert}{\href{mailto:romain.ilbert@hotmail.fr}{romain.ilbert@hotmail.fr}}
\icmlcorrespondingauthor{Ambroise Odonnat}{\href{mailto:ambroiseodonnattechnologie@gmail.com}{ambroiseodonnattechnologie@gmail.com}}

\icmlkeywords{Machine Learning, ICML}

\vskip 0.3in
]



\printAffiliationsAndNotice{\icmlEqualContribution} 

\begin{abstract}
Transformer-based architectures achieved breakthrough performance in natural language processing and computer vision, yet they remain inferior to simpler linear baselines in multivariate long-term forecasting. To better understand this phenomenon, we start by studying a toy linear forecasting problem for which we show that transformers are incapable of converging to their true solution despite their high expressive power. 
We further identify the attention of transformers as being responsible for this low generalization capacity. 
Building upon this insight, we propose a shallow lightweight transformer model that successfully escapes bad local minima when optimized with sharpness-aware optimization.
We empirically demonstrate that this result extends to all commonly used real-world multivariate time series datasets. In particular, \model{} surpasses current state-of-the-art methods and is on par with the biggest foundation model \texttt{MOIRAI} while having significantly fewer parameters. The code is available at \href{https://github.com/romilbert/samformer}{https://github.com/romilbert/samformer}. 
\end{abstract}

\section{Introduction}
\label{sec:intro}
Multivariate time series forecasting is a classical learning problem that consists of analyzing time series to predict future trends based on historical information. In particular, long-term forecasting is notoriously challenging due to feature correlations and long-term temporal dependencies in time series. This learning problem is prevalent in those real-world applications where observations are gathered sequentially, such as medical data~\citep{cepulionis2016electro}, electricity consumption~\citep{electricity}, temperatures~\citep{weather}, or stock prices~\citep{sonkavde2023stock}. A plethora of methods have been developed for this task, from classical mathematical tools~\citep{sorjamaa2007methodology, chen2021hamiltonian} and statistical approaches like ARIMA~\citep{box1990arima, box1974forecasting} to more recent deep learning ones~\citep{casolaro2023survey}, including recurrent and convolutional neural networks~\citep{rangapuram2018deep, salinas2020deepar, fan2017multistep, lai2018lstnet, sen2019tcn}. 

\begin{figure}[t!]
   \centering
   \includegraphics[width=0.41\textwidth]{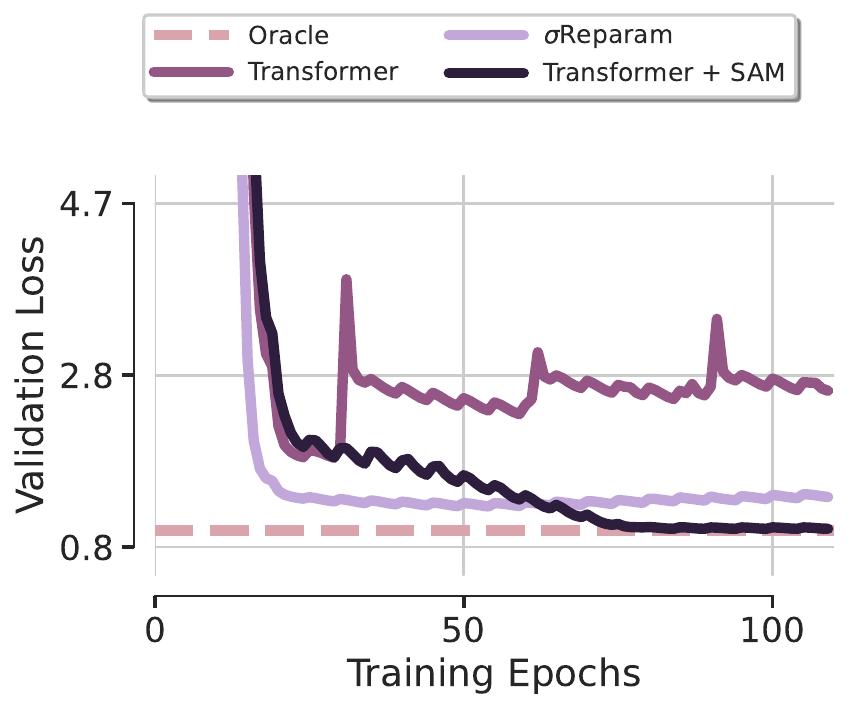}
   \caption{Illustration of our approach on synthetic data. \texttt{Oracle} is the optimal solution, \transformer{} is a base transformer, \sreparam{} is a \transformer{} with weight rescaling~\citep{zhai2023collapse} and \texttt{Transformer + SAM} is \transformer{} trained with sharpness-aware minimization. \transformer{} overfits, \sreparam{} improves slightly but fails to reach \texttt{Oracle} while \transam{} generalizes perfectly. This motivates \model{}, a shallow transformer combining SAM and best practices in time series forecasting.}
   \label{fig:toy_exp_with_SAM}
\end{figure}

Recently, the transformer architecture~\citep{vaswani2017attention} became ubiquitous in natural language processing (NLP)~\citep{devlin2018bert, radford2018improving, touvron2023llama, openai2023gpt4} and computer vision~\citep{dosovitskiy2021vit, caron2021emerging, touvron2021efficient}, achieving breakthrough performance in both domains. Transformers are known to be particularly efficient in dealing with sequential data, a property that naturally calls for their application on time series. Unsurprisingly, many works attempted to propose time series-specific transformer architectures to benefit from their capacity to capture temporal interactions~\citep{haoyi2021informer, wu2021autoformer, zhou2022fedformer, nie2023patchtst}. However, the current state-of-the-art in multivariate time series forecasting is achieved with a simpler MLP-based model \citep{chen2023tsmixer}, which significantly outperforms transformer-based methods. Moreover, \citet{zeng2022effective} have recently found that linear networks can be on par or better than transformers for the forecasting task, questioning their practical utility. This curious finding serves as a starting point for our work.

\paragraph{Limitation of current approaches.} Recent works applying transformers to time series data have mainly focused on either (i) efficient implementations reducing the quadratic cost of attention ~\citep{li2019logtrans, liu2022pyraformer, cirstea2022triformer, kitaev2020reformer, haoyi2021informer, wu2021autoformer} or (ii) decomposing time series to better capture the underlying patterns in them~\citep{wu2021autoformer, zhou2022fedformer}. Surprisingly, none of these works have specifically addressed a well-known issue of transformers related to their training instability, particularly present in the absence of large-scale data
~\citep{liu2020understanding,dosovitskiy2021vit}.
 
\paragraph{Trainability of transformers.}
In computer vision and NLP, it has been found that attention matrices can suffer from entropy or rank collapse \citep{dong2021attentionrank}. Then, several approaches have been proposed to overcome these issues \cite{chen2022vitwithsam,zhai2023collapse}. However, in the case of time series forecasting, open questions remain about how transformer architectures can be trained effectively without a tendency to overfit.  We aim to show that by eliminating training instability, transformers can excel in multivariate long-term forecasting, contrary to previous beliefs of their limitations.

\paragraph{Summary of our contributions.} Our proposal puts forward the following contributions: 
\begin{enumerate}
    \item We show that even when the transformer architecture is tailored to solve a simple toy linear forecasting problem, it still generalizes poorly and converges to sharp local minima. We further identify that attention is mainly responsible for this phenomenon;
    \item We propose a shallow transformer model, termed \model{}, that incorporates the best practices proposed in the research community including reversible instance normalization (RevIN, \citeauthor{kim2021reversible} \citeyear{kim2021reversible}) and channel-wise attention \cite{Zhang_2022_CVPR, zamir2021Restormer} recently introduced in computer vision community. We show that optimizing such a simple transformer with sharpness-aware minimization (SAM) allows convergence to local minima with better generalization;
    \item We empirically demonstrate the superiority of our approach on common multivariate long-term forecasting datasets.  \model{} surpasses current state-of-the-art methods and is on par with the biggest foundation model \texttt{MOIRAI} while having significantly fewer parameters.
\end{enumerate}

\section{Proposed Approach}
\paragraph{Notations.} We represent scalar values with regular letters (e.g., parameter $\lambda$), vectors with bold lowercase letters (e.g., vector $\mbf{x}$), and matrices with bold capital letters (e.g., matrix $\mbf{M}$). We denote by $\mbf{M}^\top$ the transpose of $\mbf{M}$ and likewise for vectors. The rank of a matrix $\mbf{M}$ is denoted by $\rk{\mbf{M}}$, and its Frobenius norm by $\lVert \mbf{M} \rVert_\mathrm{F}$.
We let $\tilde{n} = \min\{n,m\}$, and denote by $\| \bM \|_{*} = \sum_{i=1}^{\tilde{n}} \sigma_i(\bM)$ the nuclear norm of $\bM$ with $\sigma_i(\bM)$ being its singular values, and by $\| \bM \|_{2} = \sigma_{\max}(\bM)$ its spectral norm. The identity matrix of size $n\times n$ is denoted by $\mbf{I}_n$. The notation $\mbf{M} \succcurlyeq \mbf{0}$ indicates that $\mbf{M}$ is positive semi-definite.
\label{sec:contrib}
\subsection{Problem Setup}
We consider the multivariate long-term forecasting framework: given a $D$-dimensional time series of length $L$ (\emph{look-back window}), arranged in a matrix $\mathbf{X}\in\mathbb{R}^{D\times L}$ to facilitate channel-wise attention, our objective is to predict its next $H$ values (\emph{prediction horizon}), denoted by $\mathbf{Y}\in\mathbb{R}^{D\times H}$. We assume that we have access to a training set that consists of $N$ observations $(\mathcal{X},\mathcal{Y}) = (\{\mbf{X}^{(i)}\}_{i=0}^N,\ \{\mbf{Y}^{(i)}\}_{i=0}^N)$, and denote by $\mbf{X}^{(i)}_d\in\mathbb{R}^{1\times L}$ (respectively \,$\mbf{Y}^{(i)}_d\in\mathbb{R}^{1\times H}$) the $d$-th feature of the $i$-th input (respectively target) time series.
We aim to train a predictor $f_{\bm{\omega}}: \RR^{D\times L}\to\RR^{D\times H}$ parameterized by $\bm{\omega}$ that minimizes the mean squared error (MSE) on the training set:
\begin{equation}
\label{eq:training_loss}
    \mathcal{L}_\mathrm{train}(\bm{\omega}) = \frac{1}{ND}\sum_{i=0}^N \lVert \mbf{Y}^{(i)} - f_{\bm{\omega}}(\mbf{X}^{(i)})\rVert_\mathrm{F}^2\,.
\end{equation}

\subsection{Motivational Example} Recently, \citet{zeng2022effective} showed that transformers perform on par with, or are worse than, simple linear neural networks trained to directly project the input to the output. We use this observation as a starting point by considering the following generative model for our toy regression problem mimicking a time series forecasting setup considered later:
\begin{equation}
    \label{eq:toy_exp}
    \mbf{Y} = \mbf{X}\mbf{W}_\mathrm{toy} + \bm{\varepsilon}.
\end{equation}
We let $L\!=\!512, H\!=\!96, D\!=\!7$ and $\mbf{W}_\mathrm{toy} \in \mathbb{R}^{L \times H}, \bm{\epsilon} \in \mathbb{R}^{D \times H}$ having random normal entries and generate $15000$ input-target pairs $\mleft( \mbf{X}, \mbf{Y}\mright)$ ($10000$ for train and $5000$ for validation), with $\mbf{X} \in \RR^{D \times L}$ having random normal entries. 

Given this generative model, we would like to develop a transformer architecture that can efficiently solve the problem in Eq.~\eqref{eq:toy_exp} without unnecessary complexity. To achieve this, we propose to simplify the usual transformer encoder by applying attention to $\bX$ and incorporating a residual connection that adds $\bX$ to the attention's output. Instead of adding a feedforward block on top of this residual connection, we directly employ a linear layer for output prediction. Formally, our model is defined as follows: 
\begin{equation}
    \label{eq:transformer_model}
    f(\bX) = \mleft[\bX + \bA(\bX)\bX\bW_V\bW_O\mright]\bW,
\end{equation}
with $\mbf{W} \in \RR^{L \times H}, \mbf{W}_V \in \RR^{L \times d_\mathrm{m}}$, $\mbf{W}_O \in \RR^{d_\mathrm{m} \times L}$ and 
$\bA(\mbf{X})$ being the \emph{attention matrix} of an input sequence $\bX \in \mathbb{R}^{D \times L}$ defined as
\begin{equation}
    \label{eq:attention_matrix}
    \bA(\bX) = \softmax{\frac{\mbf{X}\mbf{W}_Q\mbf{W}_K^\top\mbf{X}^\top}{\sqrt{d_\mathrm{m}}}} \in \mathbb{R}^{D \times D}
\end{equation}
where the softmax is row-wise, $\mbf{W}_Q \in \RR^{L \times d_\mathrm{m}}, \mbf{W}_K \in \RR^{L \times d_\mathrm{m}}$, and $d_\mathrm{m}$ is the dimension of the model. The softmax makes $\bA(\bX)$ right stochastic, with each row describing a probability distribution. To ease the notations, in contexts where it is unambiguous, we refer to the attention matrix simply as $\bA$, omitting $\bX$.

We term this architecture \transformer{} and briefly comment on it. First, the attention matrix is applied channel-wise, which simplifies the problem and reduces the risk of overparametrization, as the matrix $\bW$ has the same shape as in Eq.~\eqref{eq:toy_exp} and the attention matrix becomes much smaller due to $L>D$. In addition,
channel-wise attention is more relevant than temporal attention in this scenario, as data generation follows an i.i.d. process according to Eq.~\eqref{eq:toy_exp}. We formally establish the identifiability of $\bW_\text{toy}$ by our model below. The proof is deferred to Appendix~\ref{app:optimal_transformer}.
\begin{boxprop}[Existence of optimal solutions]
\label{prop:optimal_transformer}
    Assume $\bW_{Q}, \bW_K, \bW_V$ and $\bW_O$ are fixed and let $\bP = \bX + \bA(\bX) \bX \bW_V \bW_O \in \RR^{D \times L}$. Then, there exists a matrix $\bW \in \RR^{L \times H}$ such that $\bP \bW = \bX \bW_\text{toy}$  
    if, and only if, $\rk{[\bP\quad \bX\bW_{\mathrm{toy}}]} = \rk{\bP}$
    where $[\bP\quad \bX\bW_{\mathrm{toy}}] \in \RR^{D \times (L+H)}$ is a block matrix.
\end{boxprop}
The assumption made above is verified if $P$ is full rank and $D < H$, which is the case in this toy experiment. Consequently, the optimization problem of fitting a transformer on data generated with Eq.~\eqref{eq:toy_exp} theoretically admits infinitely many optimal classifiers $\bW$. 

We would now like to identify the role of attention in solving the problem from Eq.~\eqref{eq:transformer_model}. To this end, we consider a model, termed \texttt{Random Transformer}, where only $\bW$ is optimized, while self-attention weights $\bW_Q, \bW_K, \bW_V, \bW_O$ are fixed during training and initialized following~\citet{glorot2010initialization}. This effectively makes the considered transformer act like a linear model. Finally, we compare the local minima obtained by these two models after their optimization using Adam with the \texttt{Oracle} model that corresponds to the least squares solution of Eq.~\eqref{eq:toy_exp}.

\begin{figure}[!ht]
   \centering
   \includegraphics[width=0.49\textwidth]{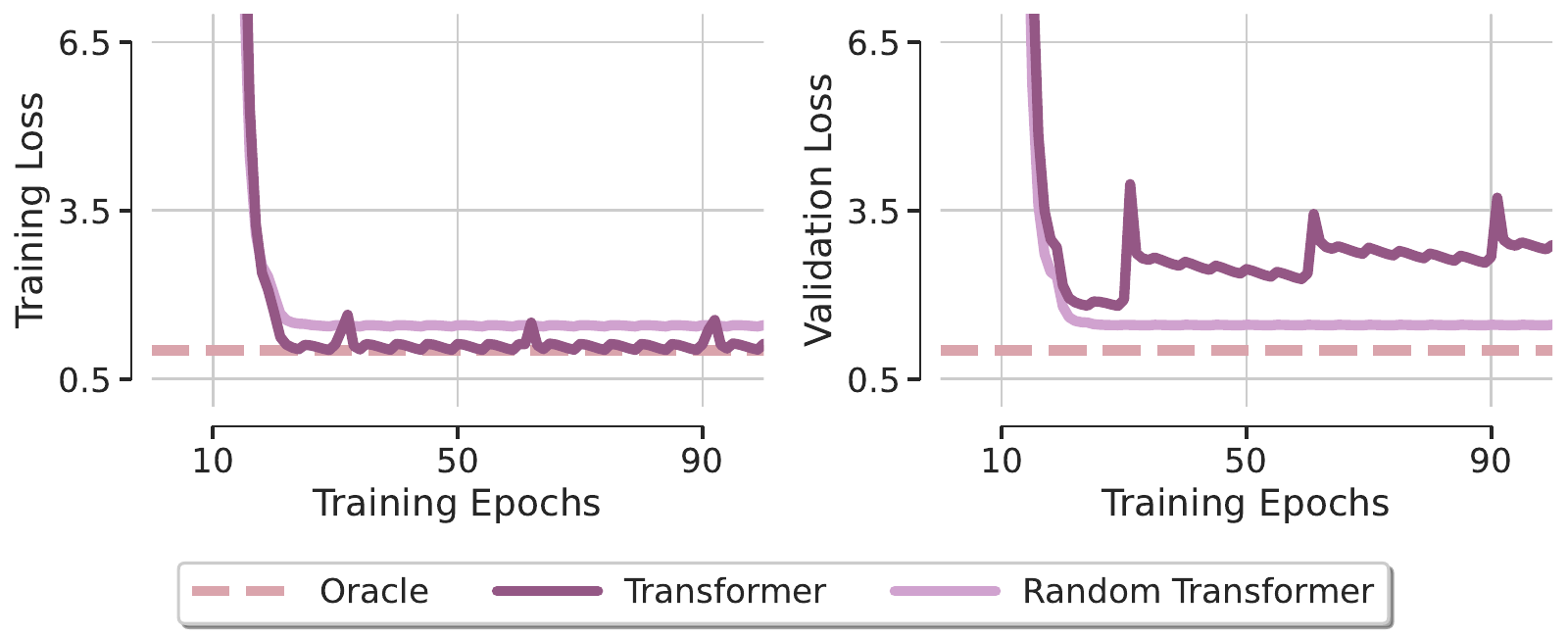}
   \caption{\textbf{Poor generalization.} Despite its simplicity, \transformer{} suffers from severe overfitting. Fixing the attention weights in \texttt{Random Transformer} improves the generalization, hinting at the role of attention in preventing convergence to optimal local minima.}
   \label{fig:toy_exp_without_SAM}
\end{figure}

We present the validation loss for both models in Figure~\ref{fig:toy_exp_without_SAM}. A first surprising finding is that both transformers fail to recover $\bW_{\mathrm{toy}}$, highlighting that optimizing even such a simple architecture with a favorable design exhibits a strong lack of generalization. When fixing the self-attention matrices, the problem is alleviated to some extent, although \texttt{Random Transformer} remains suboptimal. This observation remains consistent across various optimizers (see Figure~\ref{fig:sensitivity_optim} in Appendix~\ref{app:ablation_sensitivity}) and values of learning rate, suggesting that this phenomenon is not attributable to suboptimal optimizer hyperparameters or the specific choice of the optimizer. As there is only a $2\%$ increase in the number of parameters between the \texttt{Random Transformer} and the \transformer{}, it is not due to overfitting either. Hence, we deduce from Figure~\ref{fig:toy_exp_with_SAM} that the poor generalization capabilities of \transformer{} are mostly due to the trainability issues of the attention module.

\subsection{Transformer's Loss Landscape}

\begin{figure*}[htbp]
\centering
\begin{subfigure}{.49\textwidth}
  \centering
  \includegraphics[width=1\linewidth]{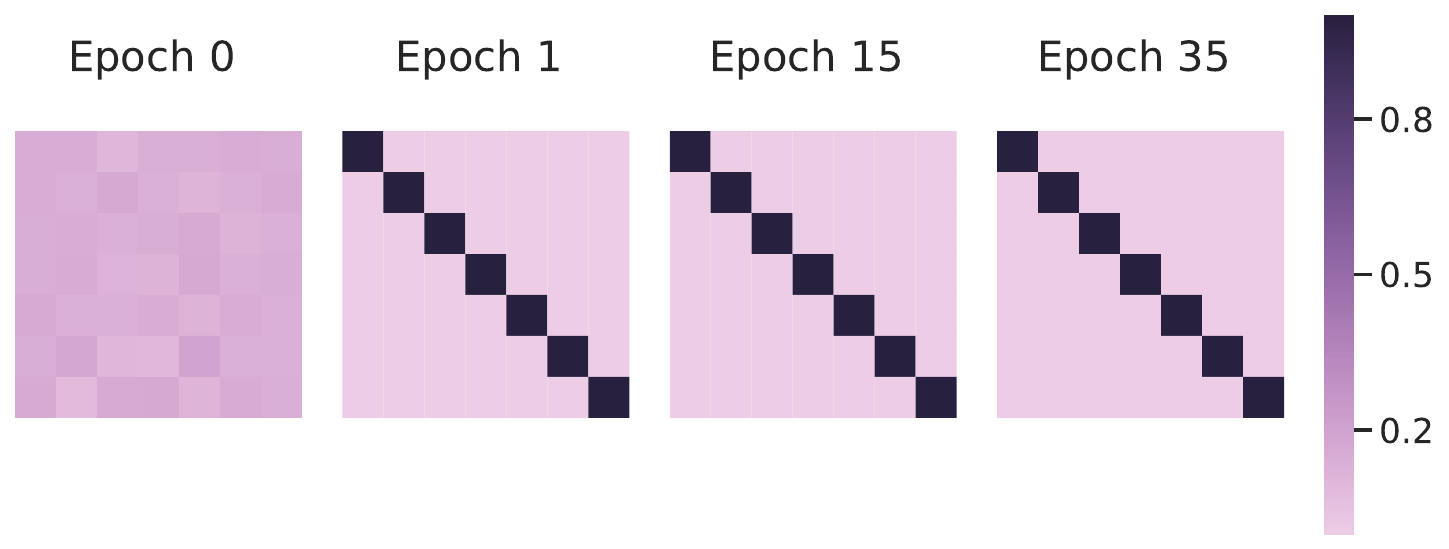}
  \caption{Attention matrices of \transformer{} along the training.}
  \label{fig:attention_matrix_toy}
\end{subfigure}%
\begin{subfigure}{.49\textwidth}
  \centering
  \includegraphics[width=1\linewidth]{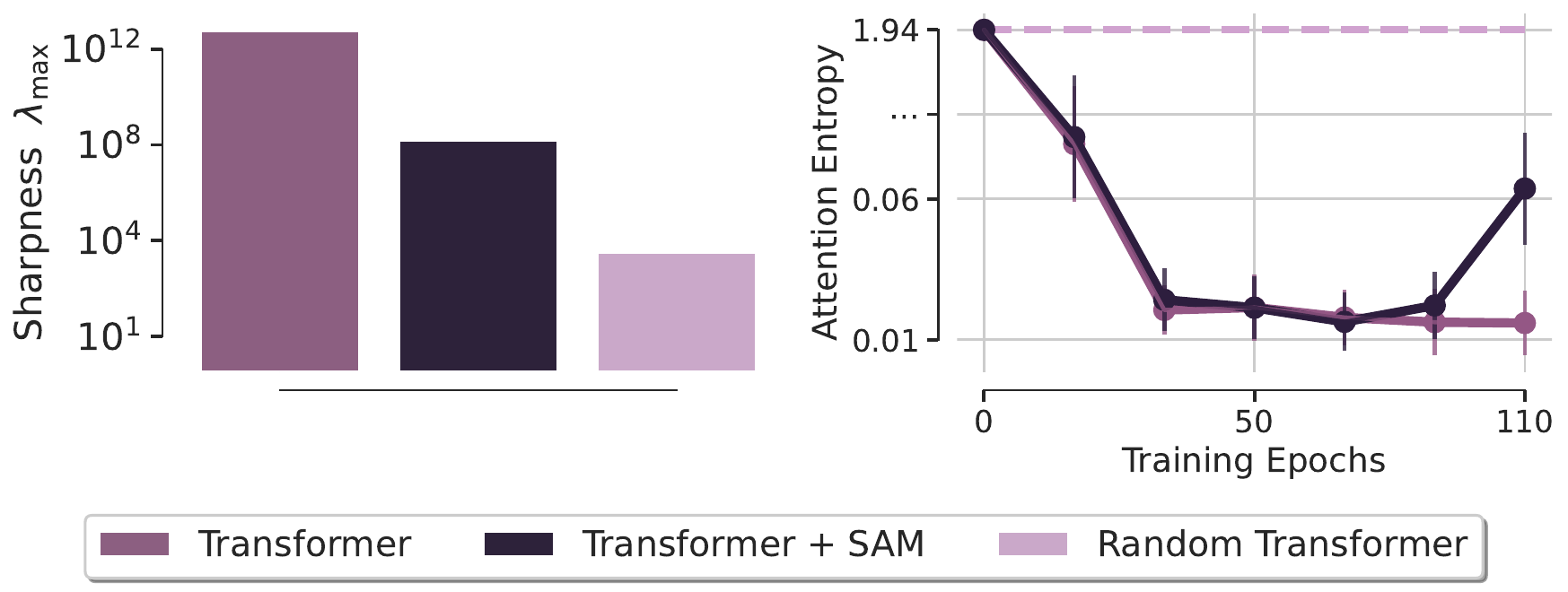}
  \caption{Sharpness at the end of the training, Entropy collapse.}
\label{fig:sharpness_entropy_collapse_toy}
\end{subfigure}
\caption{\textbf{Transformer's loss landscape analysis for linear regression}. \textbf{(a)} The attention matrices of \transformer{} get stuck to identity from the first epoch. \textbf{(b, left)} \transformer{} converges to sharper minimum than \texttt{Transformer+SAM} with much larger $\lambda_\mathrm{max}$ ($\sim \times 10^4)$, while \texttt{Random Transformer} has a smooth loss landscape. \textbf{(b, right)} \transformer{} suffers from entropy collapse during training confirming the high sharpness of its loss landscape.}
\label{fig:toy_exp}
\end{figure*}

\begin{figure*}[h]
\centering
\begin{subfigure}{.49\textwidth}
  \centering
  \includegraphics[width=\linewidth]{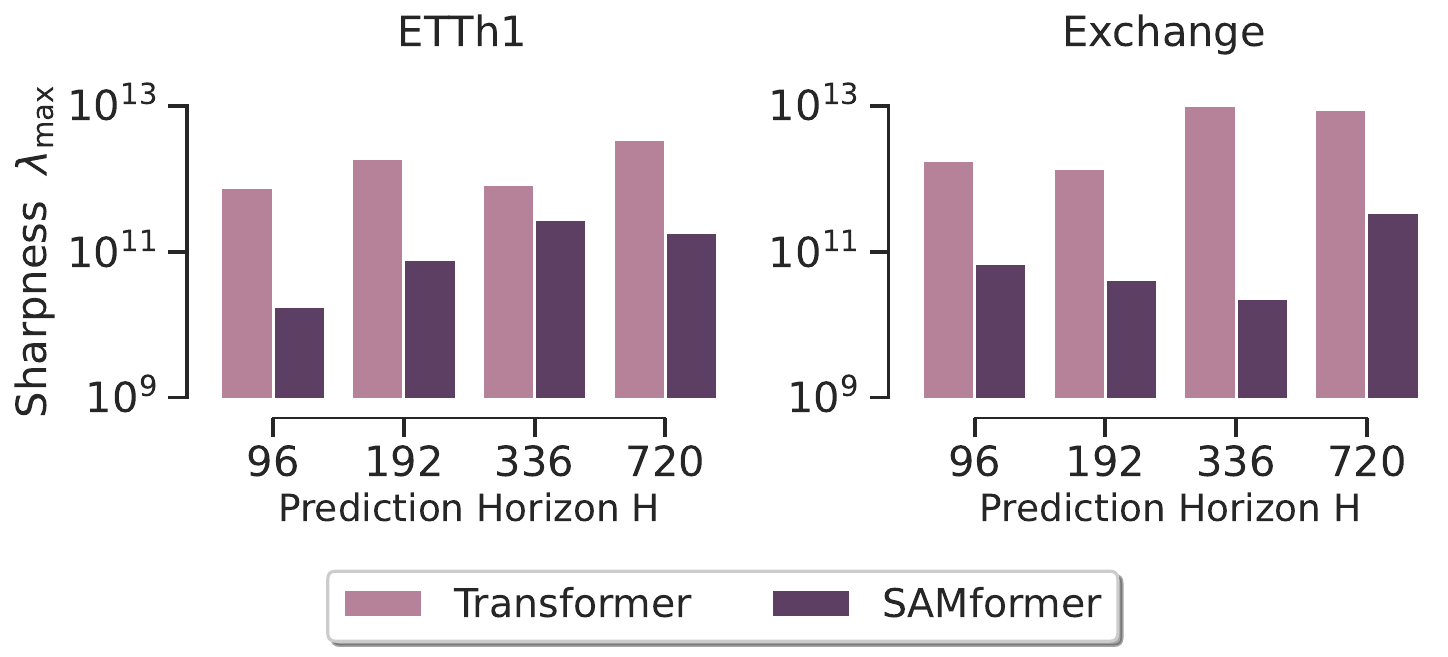}
  \caption{Sharpness of \model{} and \transformer{}.}
  \label{fig:sharpness_2_datasets}
\end{subfigure}%
\begin{subfigure}{.49\textwidth}
  \centering
  \includegraphics[width=\linewidth]{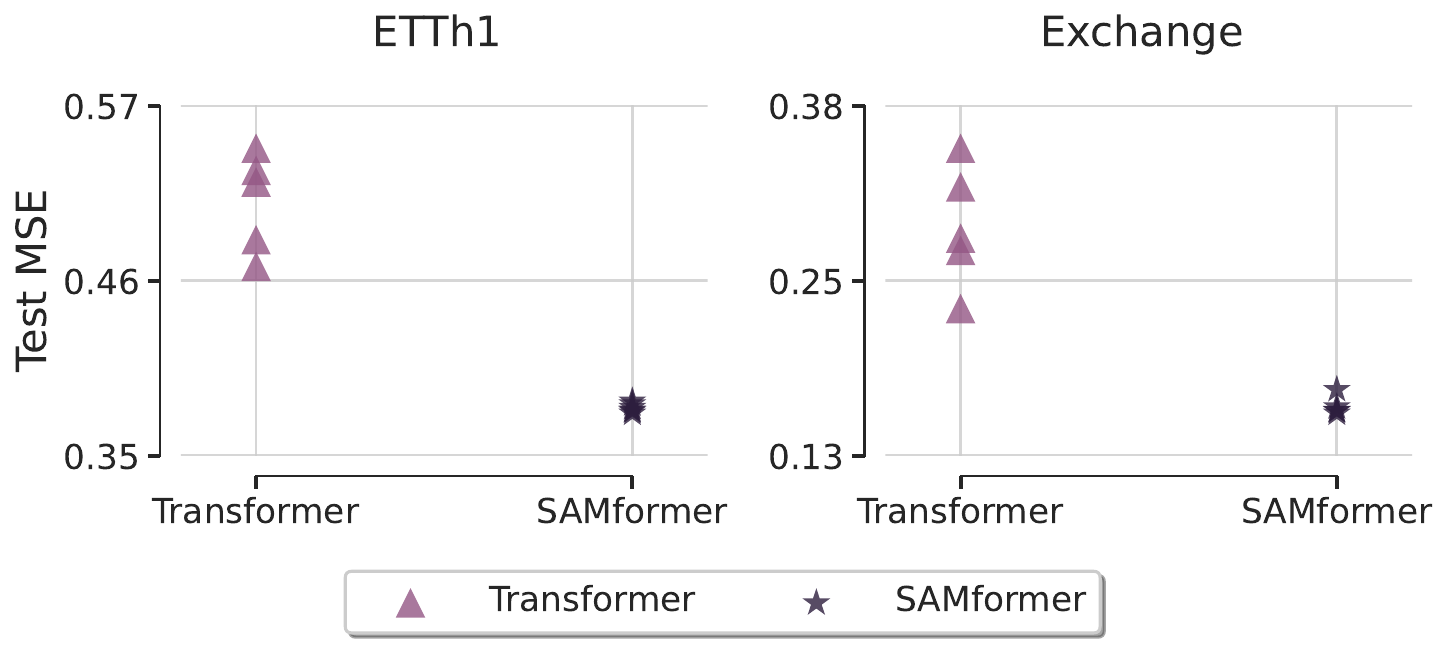}
  \caption{Performance across runs of \model{} and \transformer{}.}
\label{fig:stabilized_performance_3_datasets}
\end{subfigure}
\caption{\textbf{(a)} \model{} has a smoother loss landscape than \transformer{}. \textbf{(b)} \model{} consistently generalize well for every initialization while \transformer{} is unstable and heavily depends on the seed.}
\label{fig:benefits_samformer}
\end{figure*}

\paragraph{Intuition.} In the previous section, we concluded that the attention was at fault for the poor generalization of \transformer{} observed above. To develop our intuition behind this phenomenon, we plot in Figure~\ref{fig:attention_matrix_toy} the attention matrices at different epochs of training. We can see that the attention matrix is close to the identity matrix right after the very first epoch and barely changes afterward, especially with the softmax amplifying the differences in the matrix values. It shows the emergence of \emph{attention's entropy collapse} with a full-rank attention matrix, which was identified in~\citet{zhai2023collapse} as one of the reasons behind the hardness of training transformers. This work also establishes a relationship between entropy collapse and the sharpness of the transformers' loss landscape which we confirm in Figure~\ref{fig:sharpness_entropy_collapse_toy} (a similar behavior is obtained on real data in Figure~\ref{fig:sharpness_2_datasets}. The \transformer{} converges to a sharper minimum than the \texttt{Random Transformer} while having a significantly lower entropy (the attention being fixed at initialization for the latter, its entropy remains constant along training). These pathological patterns suggest that the \transformer{} fails because of the entropy collapse and the sharpness of its training loss. In the next paragraph, we investigate the existing solutions in the literature to alleviate those issues.

\paragraph{Existing solutions.}
Recent studies have demonstrated that the loss landscape of transformers is sharper compared to other residual architectures \citep{chen2022vitwithsam,zhai2023collapse}. This may explain training instability and subpar performance of transformers, especially when trained on small-scale datasets. The sharpness of transformers was observed and quantified differently: while \citet{chen2022vitwithsam} computes $\lambda_\mathrm{max}$, the largest eigenvalue of the loss function's Hessian, \citet{zhai2023collapse} gauges the entropy of the attention matrix to demonstrate its collapse with high sharpness. Both these metrics are evaluated, and their results are illustrated in Figure~\ref{fig:sharpness_entropy_collapse_toy}. This visualization confirms our hypothesis, revealing both detrimental phenomena at once. On the one hand, the sharpness of the transformer with fixed attention is orders of magnitude lower than the sharpness of the transformer that converges to the identity attention matrix. On the other hand, the entropy of the transformer's attention matrix is dropping sharply along the epochs when compared to the initialization. 

To identify an appropriate solution allowing a better generalization performance and training stability, we explore both remedies proposed by \citet{chen2022vitwithsam} and \citet{zhai2023collapse}. The first approach involves utilizing the recently proposed sharpness-aware minimization framework \citep{foret2021sharpnessaware} which replaces the training objective $\mathcal{L}_\mathrm{train}$ of Eq.~\eqref{eq:training_loss} by 
\begin{align*}
\mathcal{L}_\mathrm{train}^{\textrm{SAM}}(\bomega) &= \max_{\|\bepsilon\| < \rho} \mathcal{L}_\mathrm{train}(\bomega + \bepsilon)\,,
\end{align*}
where $\rho>0$ is an hyper-parameter (see Remark~\ref{rmk:sam_rho} of Appendix~\ref{app:additional_background}), and $\bomega$ are the parameters of the model. More details on SAM can be found in Appendix~\ref{app:sam}. The second approach involves reparameterizing all weight matrices with spectral normalization and an additional learned scalar, a technique termed \sreparam{} by \citet{zhai2023collapse}. More formally, we replace each weight matrix $\bW$ as follows 
\begin{equation}
\label{eq:sigma_reparam}
    \widehat{\mbf{W}} = \frac{\gamma}{\lVert \mbf{W} \rVert_2} \mbf{W},
\end{equation}
where $\gamma \in \mathbb{R}$ is a learnable parameter initialized at $1$.

The results depicted in Figure~\ref{fig:toy_exp_with_SAM} highlight our transformer's successful convergence to the desired solution. Surprisingly, this is only achieved with SAM, as \sreparam{} doesn't manage to approach the optimal performance despite maximizing the entropy of the attention matrix. In addition, one can observe in Figure~\ref{fig:sharpness_entropy_collapse_toy} that the sharpness with SAM is several orders of magnitude lower than the \transformer{} while the entropy of the attention obtained with SAM remains close to that of a base \transformer{} with a slight increase in the later stages of the training. It suggests that entropy collapse as introduced in \citet{zhai2023collapse} is benign in this scenario.

To better understand the failure of \sreparam{}, it can be useful to recall how Eq.~\eqref{eq:sigma_reparam} was derived. \citet{zhai2023collapse} departed from a tight lower bound
on the attention entropy and showed that it increases
exponentially fast when $\lVert\mbf{W}_Q\mbf{W}_K^\top\rVert_2$ is minimized \citep[see Theorem 3.1]{zhai2023collapse}. Eq.~\eqref{eq:sigma_reparam} was proposed as a simple way to minimize this quantity.
In the case of channel-wise attention, however, it can be shown that this has a detrimental effect on the rank of the attention matrix, which would consequently exclude certain features from being considered by the attention mechanism.
We formalize this intuition in the following Proposition~\ref{thm:upper_bound_nuclear_norm}, where we consider the nuclear norm, a sum of the singular values, as a smooth proxy of the algebraic rank, which is a common practice ~\citep{daneshmand2020bncollapse, dong2021attentionrank}. The proof is deferred to Appendix~\ref{app:upper_bound_nuclear_norm}.
\begin{boxprop}[Upper bound on the nuclear norm]
\label{thm:upper_bound_nuclear_norm}
    Let $\mbf{X} \in \mathbb{R}^{D \times L}$ be an input sequence. Assuming $\mbf{W}_Q\mbf{W}_K^\top = \mbf{W}_K\mbf{W}_Q^\top \succcurlyeq \mbf{0}$, we have
    \begin{equation*}
        \lVert \mbf{X}\mbf{W}_Q\mbf{W}_K^\top\mbf{X}^\top\rVert_* \leq \lVert\mbf{W}_Q\mbf{W}_K^\top\rVert_2 \lVert\mbf{X}\rVert_\mathrm{F}^2.
    \end{equation*}
\end{boxprop}
Note that the assumption made above holds when ${\bW_Q\!=\!\bW_K}$ and has been previously studied by ~\citet{kim2021lipschitz}.
The theorem confirms that employing \sreparam{} to decrease $\lVert\mbf{W}_Q\mbf{W}_K^\top\rVert_2$ reduces the nuclear norm of the numerator of attention matrix defined by Eq.~\eqref{eq:attention_matrix}. While the direct link between matrix rank and this nuclear norm does not always hold, nuclear norm regularization is commonly used to encourage a low-rank structure in compressed sensing~\citep{recht2010guaranteed, rechet2011matrixcompletion, candes2012matrixcompletion}.

Although Proposition~\ref{thm:upper_bound_nuclear_norm} cannot be directly applied to the attention matrix $\bA(\bX)$, we point out that in the extreme case when \sreparam{} leads to the attention scores $\mbf{X}\mbf{W}_Q\mbf{W}_K^\top\mbf{X}^\top$ to be rank-$1$ with identical rows as studied in~\citep{anagnostidis2022signal}, that the attention matrix stays rank-$1$ after application of the row-wise softmax.
Thus, \sreparam{} may induce a collapse of the attention rank that we empirically observe in terms of nuclear norm in~Figure~\ref{figure:nuclear.norm}. With these findings, we present a new simple transformer model with high performance and training stability for multivariate time series forecasting.

\subsection{\model{}: Putting It All Together}
The proposed \model{} is based on Eq.~\eqref{eq:transformer_model} with two important modifications. 
First, we equip it with Reversible Instance Normalization (RevIN, \citet{kim2021reversible}) applied to $\bX$ as this technique was shown to be efficient in handling the shift between the training and testing data in time series. Second, as suggested by our explorations above, we optimize the model with SAM to make it converge to flatter local minima. Overall, this gives the shallow transformer model with one encoder in Figure \ref{fig:samformer-diagram}.

\setlength{\intextsep}{2pt}%
\setlength{\columnsep}{6pt}%
\begin{wrapfigure}[14]{r}{0.14\textwidth}
\centering
\scalebox{.55}{
\begin{tikzpicture}
\definecolor{emb_color}{RGB}{252,224,225}
\definecolor{multi_head_attention_color}{RGB}{252,226,187}
\definecolor{add_norm_color}{RGB}{242,243,193}
\definecolor{ff_color}{RGB}{194,232,247}
\definecolor{softmax_color}{RGB}{203,231,207}
\definecolor{linear_color}{RGB}{220,223,240}
\definecolor{gray_bbox_color}{RGB}{243,243,244}

\draw[fill=gray_bbox_color, line width=0.046875cm, rounded corners=0.300000cm] (-1.2000, 4.2) -- (2.82000, 4.2) -- (2.82000, 1.30000) -- (-1.2000, 1.30000) -- cycle;

\node[text width=2.500000cm, anchor=north, align=center] at (1.250000,-0.2500000) {Input};
\draw[line width=0.046875cm, -latex] (1.250000, -0.300000) -- (1.250000, 0.300000);

\draw[line width=0.046875cm, fill=ff_color, rounded corners=0.100000cm] (0.000000, 0.800000) -- (2.500000, 0.800000) -- (2.500000, 0.300000) -- (0.000000, 0.300000) -- cycle;
\node[text width=2.500000cm, align=center] at (1.250000,0.550000) {RevIN};
\draw[line width=0.046875cm, -latex] (1.250000, 0.800000) -- (1.250000, 2.130000);

\draw[-latex, line width=0.046875cm, rounded corners=0.200000cm] (1.250000, 1.530000) -- (-0.750000, 1.530000) -- (-0.750000, 3.430000) -- (0.000000, 3.430000);
\draw[-latex, line width=0.046875cm, rounded corners=0.200000cm] (1.250000, 1.730000) -- (0.312500, 1.730000) -- (0.312500, 2.130000);
\draw[-latex, line width=0.046875cm, rounded corners=0.200000cm] (1.250000, 1.730000) -- (2.187500, 1.730000) -- (2.187500, 2.130000);

\draw[line width=0.046875cm, fill=multi_head_attention_color, rounded corners=0.100000cm] (0.000000, 3.030000) -- (2.500000, 3.030000) -- (2.500000, 2.130000) -- (0.000000, 2.130000) -- cycle;
\node[text width=2.500000cm, align=center] at (1.250000,2.580000) {Channel-Wise \vspace{-0.05cm} \linebreak Self-Attention};
\draw[line width=0.046875cm] (1.250000, 3.030000) -- (1.250000, 3.180000);

\draw[line width=0.046875cm, fill=add_norm_color, rounded corners=0.100000cm] (0.000000, 3.680000) -- (2.500000, 3.680000) -- (2.500000, 3.180000) -- (0.000000, 3.180000) -- cycle;
\node[text width=2.500000cm, align=center] at (1.250000,3.430000) {Add};
\draw[line width=0.046875cm, -latex] (1.250000, 3.680000) -- (1.250000, 4.70000);


\draw[line width=0.046875cm, fill=linear_color, rounded corners=0.100000cm] (0.000000, 5.20000) -- (2.500000, 5.20000) -- (2.500000, 4.70000) -- (0.000000, 4.70000) -- cycle;
\node[text width=2.500000cm, align=center] at (1.250000,4.95000) {Linear};
\draw[line width=0.046875cm, -latex] (1.250000, 5.20000) -- (1.250000, 5.80000);

\draw[line width=0.046875cm, fill=ff_color, rounded corners=0.100000cm] (0.000000, 6.30000) -- (2.500000, 6.30000) -- (2.500000, 5.80000) -- (0.000000, 5.80000) -- cycle;
\node[text width=2.500000cm, align=center] at (1.250000,6.050000) {RevIN$^{-1}$};
\draw[line width=0.046875cm, -latex] (1.250000, 6.30000) -- (1.250000, 6.90000);

\node[text width=2.500000cm, anchor=south, align=center] at (1.250000,6.85) {Output };

\end{tikzpicture}
}
\caption{\texttt{SAM} -\texttt{former}}
\label{fig:samformer-diagram}
\end{wrapfigure}

We highlight that \model{} keeps the channel-wise attention represented by a matrix $D \times D$ as in Eq.~\eqref{eq:transformer_model}, contrary to spatial (or temporal) attention given by $L \times L$ matrix used in other models. This brings two important benefits: (i) it ensures feature permutation invariance, eliminating the need for positional encoding, commonly preceding the attention layer; (ii) it leads to a reduced time and memory complexity as $D \leq L$ in most of the real-world datasets. Our channel-wise attention examines the average impact of each feature on the others throughout all timesteps. An ablation study, detailed in Appendix~\ref{app:choice_implementation}, validates the effectiveness of this implementation. We are now ready to evaluate \model{} on common multivariate time series forecasting benchmarks, demonstrating its superior t

\section{Experiments}
\label{section:experiments}

In this section, we empirically demonstrate the quantitative and qualitative superiority of \model{} in multivariate long-term time series forecasting on common benchmarks. We show that \model{} surpasses the current multivariate state-of-the-art \tsmixer{}~\citep{chen2023tsmixer} by $14.33\%$ while having $\sim4$ times fewer parameters. All the implementation details are provided in Appendix~\ref{app:arch_train}.

\paragraph{Datasets.} We conduct our experiments on $8$ publicly available datasets of real-world multivariate time series, commonly used for long-term forecasting~\citep{wu2021autoformer, chen2023tsmixer, nie2023patchtst, zeng2022effective}: the four Electricity Transformer Temperature datasets ETTh1, ETTh2, ETTm1 and ETTm2~\cite{haoyi2021informer}, Electricity~\citep{electricity}, Exchange~\citep{lai2018modeling}, Traffic~\citep{traffic}, and Weather~\citep{weather} datasets. All time series are segmented with input length $L=512$, prediction horizons $H \in \{96, 192, 336, 720\}$, and a stride of $1$, meaning that each subsequent window is shifted by one step.
A more detailed description of the datasets and time series preparation can be found in Appendix~\ref{app:datasets}.

\paragraph{Baselines.} We compare \model{} with \transformer{} presented earlier and \tsmixer{}~\citep{chen2023tsmixer}, a state-of-the-art multivariate baseline entirely built on MLPs. It should be noted that \citet{chen2023tsmixer} displayed the performance of \tsmixer{} for a fixed seed while in Table~\ref{tab:all_results}, we report the performance over several runs with different seeds, resulting in a more reliable evaluation. For a fair comparison, we also include the performance of \tsmixer{} trained with SAM, along with results reported by~\citet{liu2024itransformer} and \citet{chen2023tsmixer} for other recent SOTA multivariate transformer-based models: \texttt{iTransformer}~\citep{liu2024itransformer}, \texttt{PatchTST}~\citep{nie2023patchtst}, \texttt{FEDformer}~\citep{zhou2022fedformer}, \texttt{Informer}~\citep{haoyi2021informer}, and \texttt{Autoformer}~\citep{wu2021autoformer}. All the reported results are obtained using RevIN~\citep{kim2021reversible} for a more equitable comparison between \model{} and its competitors. More detailed information on these baselines can be found in Appendix~\ref{app:baselines}. 

\paragraph{Evaluation.} All models are trained to minimize the MSE loss defined in Eq.~\eqref{eq:training_loss}. The average MSE on the test set, together with the standard deviation over $5$ runs with different seeds is reported. Additional details and results, including the Mean Absolute Error (MAE), can be found in Table~\ref{tab:all_results_mae} of Appendix~\ref{app:mae_results}. Except specified otherwise, all our results are also obtained over $5$ runs with different seeds.

\subsection{Main Takeaways}

\paragraph{\model{} improves over state-of-the-art.}
The experimental results are detailed in Table~\ref{tab:all_results}, with a Student's t-test analysis available in Appendix Table~\ref{tab:significance_test}.
\model{} outperforms its competitors on $\mbf{7}$ \textbf{out of} $\mbf{8}$ datasets by a large margin. In particular, it improves over its best competitor \tsmixer{}+SAM by $\mbf{5.25\%}$, surpasses the standalone \tsmixer{} by $\mbf{14.33\%}$ and the best multivariate transformer-based model \texttt{FEDformer} by $\mbf{12.36\%}$. In addition, it improves over \transformer{} by $\mbf{16.96\%}$. \model{} also outperforms the very recent \texttt{iTransformer}, a transformer-based approach that uses both temporal and spatial attention, and \texttt{PatchTST} which was tailored for univariate time series forecasting. We notice that \texttt{iTransformer} has mixed global performance and gets beaten by \model{} on all datasets, except Exchange on which it significantly outperforms all competitors. This explains that \model{} improves it only by $\mbf{3.94}\%$ overall but up to $\mbf{8.38\%}$ without it. Finally, \model{} outperforms \texttt{PatchTST} by $\mbf{11.13\%}$. For every horizon and dataset (except Exchange), \model{} is ranked either first or second. Notably, SAM's integration improves the generalization capacity of \tsmixer{}, resulting in an average enhancement of $9.58\%$. A similar study with the MAE in Table~\ref{tab:all_results_mae} leads to the same conclusions. As \tsmixer{} trained with SAM is the second-best baseline almost always ranked second, it serves as a primary benchmark for further discussion in this section. It should be noted that \model{} has $~4$ times fewer parameters than \tsmixer{}, and several orders of magnitude fewer than the transformer-based methods.

\begin{figure}[t]
  \centering
\includegraphics[width=\linewidth]{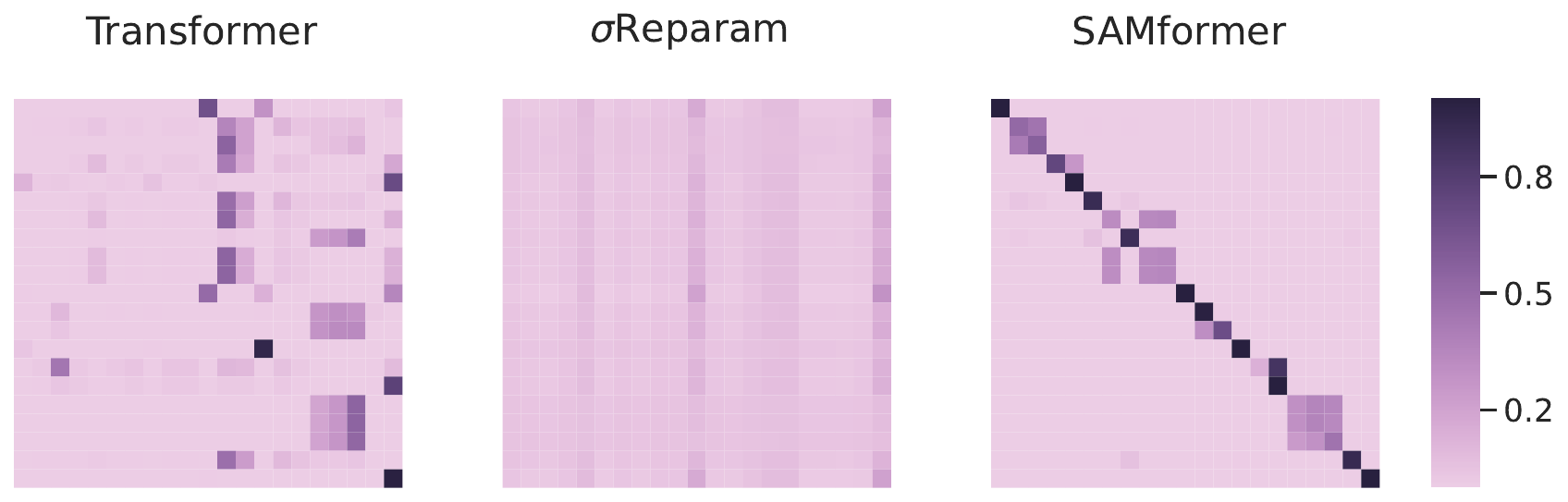}
  \caption{Attention matrices on Weather dataset. \model{} preserves self-correlation among features while \sreparam{} degrades the rank, hindering the propagation of information.
  }
\label{fig:attention_matrix_3_models}
\end{figure}

\begin{figure}[t]
\centering
\includegraphics[width=0.48\textwidth]{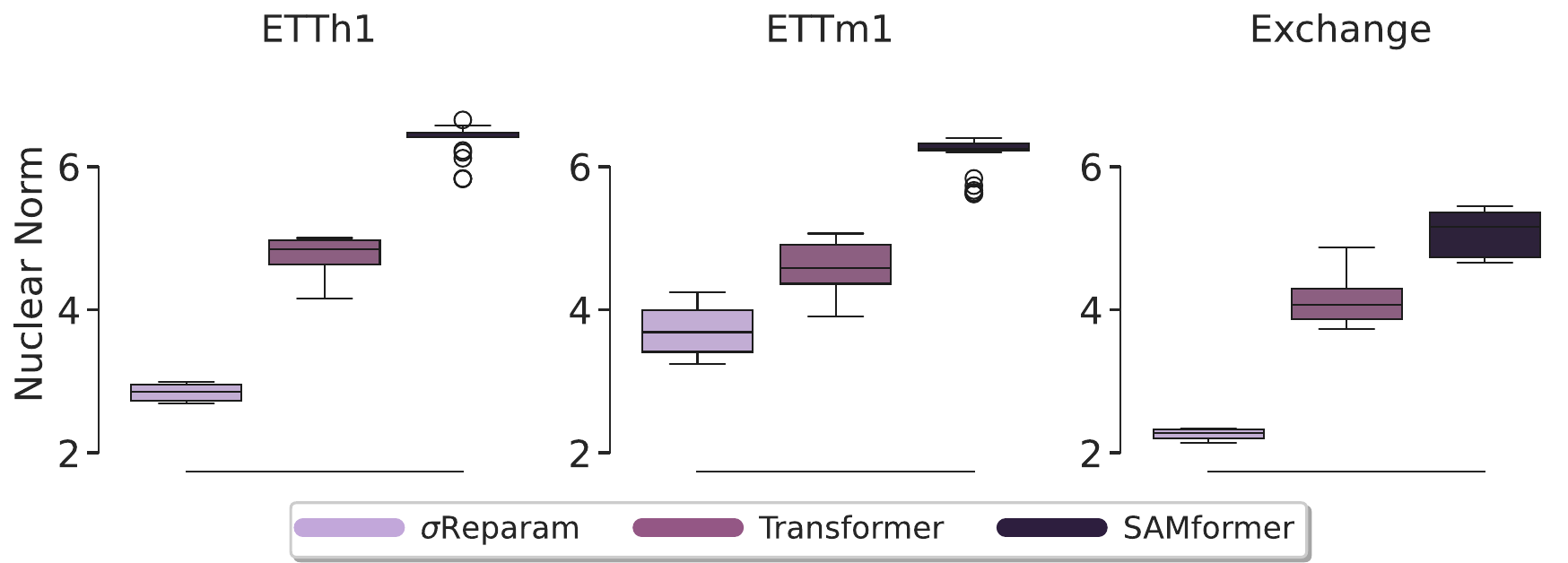}
\caption{Nuclear norm of the attention matrix for different models: $\sigma$Reparam induces lower nuclear norm in accordance with Proposition~\ref{thm:upper_bound_nuclear_norm}, while \model{} keeps the expressiveness of the attention over \transformer{}.}
\label{figure:nuclear.norm}
\end{figure}

\setlength{\tabcolsep}{0.3em}
\begin{table*}[!ht] 
\centering
\caption{Performance comparison between our model (\textcolor{bluerow}{\textbf{\model}}) and baselines for multivariate long-term forecasting with different horizons $H$. Results marked with \say{$^\dagger$} are obtained from~\citet{liu2024itransformer} and those marked with \say{$^*$} are obtained from \citet{chen2023tsmixer}, along with the publication year of the respective methods. Transformer-based models are abbreviated by removing the ``former" part of their name. We display the average test MSE with standard deviation obtained on $5$ runs with different seeds. \textbf{Best} results are in bold, \underline{second best} are underlined.}
\label{tab:all_results}
\scalebox{0.83}{
\begin{tabular}{ccccccccccc}
\toprule[\thick pt]%
\multicolumn{1}{c}{\multirow{3}{*}{Dataset}} & \multicolumn{1}{c}{\multirow{3}{*}{$H$}} & \multicolumn{2}{c}{with SAM} & \multicolumn{7}{c}{without SAM}\\ 
\cmidrule(r{10pt}l{5pt}){3-4} \cmidrule(r{10pt}l{5pt}){5-11} 
 \multicolumn{1}{c}{} & \multicolumn{1}{c}{} & \multicolumn{1}{c}{\textcolor{bluerow}{\textbf{\model}}} &  \multicolumn{1}{c}{\tsmixer{}} & \multicolumn{1}{c}{\transformer{}} & \multicolumn{1}{c}{\tsmixer{}} & \multicolumn{1}{c}{$\texttt{iTrans}^\dagger$} &\multicolumn{1}{c}{$\texttt{PatchTST}^\dagger$} &\multicolumn{1}{c}{$\texttt{In}^*$} & \multicolumn{1}{c}{$\texttt{Auto}^*$} & \multicolumn{1}{c}{$\texttt{FED}^*$}\\
\multicolumn{1}{c}{} & \multicolumn{1}{c}{} & \multicolumn{1}{c}{-} &  \multicolumn{1}{c}{-} & \multicolumn{1}{c}{-} & \multicolumn{1}{c}{\small 2023} & \multicolumn{1}{c}{\small 2024} &\multicolumn{1}{c}{\small 2023} &\multicolumn{1}{c}{\small 2021} & \multicolumn{1}{c}{\small 2021} & \multicolumn{1}{c}{\small 2022}\\
\midrule[\thick pt]
\multirow{4}{*}{\rotatebox[origin=c]{90}{ETTh1}} & $96$ & $\underline{0.381}_{\pm 0.003}$ & $0.388_{\pm 0.001}$ & $0.509_{\pm 0.031}$ & $0.398_{\pm 0.001}$& $0.386$ & $0.414$ & $0.941$ & $0.435$ & $\mbf{0.376}$ \\
& $192$ & $\mbf{0.409}_{\pm 0.002}$ & $\underline{0.421}_{\pm 0.002}$ & $0.535_{\pm 0.043}$ & $0.426_{\pm 0.003}$ &$0.441$ & $0.460$ & $1.007$ &$ 0.456$ & $0.423$\\
& $336$ & $\mbf{0.423}_{\pm 0.001}$ & $\underline{0.430}_{\pm 0.002}$ & $0.570_{\pm 0.016}$ & $0.435_{\pm 0.003}$ &$0.487$ & $0.501$ & $1.038$ & $0.486$ & $0.444$ \\ 
& $720$ & $\mbf{0.427}_{\pm 0.002}$ & $\underline{0.440}_{\pm 0.005}$ & $0.601_{\pm 0.036}$ & $0.498_{\pm 0.076}$ & $0.503$& $0.500$&  $1.144$ & $0.515$ & $0.469$\\
\midrule[\thick pt]
\multirow{4}{*}{\rotatebox[origin=c]{90}{ETTh2}} & $96$ & $\mbf{0.295}_{\pm 0.002}$ & $0.305_{\pm 0.007}$ & $0.396_{\pm 0.017}$ & $0.308_{\pm 0.003}$ &$\underline{0.297}$&$0.302$& $1.549$ & $0.332$ & $0.332$\\
& $192$ & $\mbf{0.340}_{\pm 0.002}$ & $\underline{0.350}_{\pm 0.002}$ & $0.413_{\pm 0.010}$ & $0.352_{\pm 0.004}$ &$0.380$& $0.388$& $3.792$ & $0.426$ & $0.407$\\
& $336$ & $\mbf{0.350}_{\pm 0.000}$ & $\underline{0.360}_{\pm 0.002}$ & $0.414_{\pm 0.002}$ & $0.360_{\pm 0.002}$ &$0.428$&$0.426$& $4.215$ & $0.477 $& $0.400$ \\ 
& $720$ & $\mbf{0.391}_{\pm 0.001}$ & $\underline{0.402}_{\pm 0.002}$ & $0.424_{\pm 0.009}$ & $0.409_{\pm 0.006}$ &$0.427$& $0.431$& $3.656 $& $0.453$ & $0.412$ \\
\midrule[\thick pt]
\multirow{4}{*}{\rotatebox[origin=c]{90}{ETTm1}} & $96$ & $0.329_{\pm0.001}$ & $\underline{0.327}_{\pm0.002}$ & $0.384_{\pm 0.022}$ & $0.336_{\pm 0.004}$ &$0.334$&$0.329$& $0.626$ & $0.510$ & $\mbf{0.326}$ \\
& $192$ & $\mbf{0.353}_{\pm0.006}$ & $\underline{0.356}_{\pm0.004}$ & $0.400_{\pm 0.026}$ & $0.362_{\pm 0.006}$ &$ 0.377$&$ 0.367 $& $0.725$ & $0.514$ & $0.365$ \\
& $336$ & $\mbf{0.382}_{\pm0.001}$ & $\underline{0.387}_{\pm0.004}$ & $0.461_{\pm 0.017}$ & $0.391_{\pm 0.003}$ &$0.426$&$ 0.399$& $1.005$ & $0.510$ & $0.392$ \\ 
& $720$ & $\mbf{0.429}_{\pm0.000}$ & $\underline{0.441}_{\pm0.002}$ & $0.463_{\pm 0.046}$ & $0.450_{\pm 0.006}$ &$ 0.491$&$0.454$& $1.133$ & $0.527$ & $0.446$\\
\midrule[\thick pt]
\multirow{4}{*}{\rotatebox[origin=c]{90}{ETTm2}} & $96$ & $\underline{0.181}_{\pm0.005}$ & $0.190_{\pm0.003}$ & $0.200_{\pm 0.036}$ & $0.211_{\pm 0.014}$ &$\mbf{0.180}$&$ 0.175$& $0.355$ & $0.205$ & $\mbf{0.180}$  \\
& $192$ & $\mbf{0.233}_{\pm0.002}$ & $0.250_{\pm0.002}$ & $0.273_{\pm 0.013}$ & $0.252_{\pm 0.005}$ &$0.250$&$\underline{0.241}$& $0.595$ & $0.278$ & $0.252$ \\
& $336$ & $\mbf{0.285}_{\pm0.001}$ & $\underline{0.301}_{\pm0.003}$ & $0.310_{\pm 0.022}$ & $0.303_{\pm 0.004}$ &$0.311$&$0.305$& $1.270$ & $0.343$ & $0.324$ \\ 
& $720$ & $\mbf{0.375}_{\pm0.001}$ & $\underline{0.389}_{\pm0.002}$ & $0.426_{\pm 0.025}$ & $0.390_{\pm 0.003}$ &$0.412$ &$0.402$& $3.001$ & $0.414$ & $0.410$ \\
\midrule[\thick pt]
\multirow{4}{*}{\rotatebox[origin=c]{90}{\text{\small Electricity}}} & $96$ & $\mbf{0.155}_{\pm0.002}$ & $\underline{0.171}_{\pm0.001}$ & $0.182_{\pm 0.006}$ & $0.173_{\pm 0.004}$ &-&- &$0.304$ & $0.196$ & $0.186$ \\
& $192$ & $\mbf{0.168}_{\pm0.001}$ & $\underline{0.191}_{\pm0.010}$ & $0.202_{\pm 0.041}$ & $0.204_{\pm 0.027}$& -&- &$0.327$ & $0.211$ & $0.197$ \\
& $336$ & $\mbf{0.183}_{\pm0.000}$ & $\underline{0.198}_{\pm0.006}$ & $0.212_{\pm 0.017}$ & $0.217_{\pm 0.018}$ &-&-& $0.333$ & $0.214$ & $0.213$ \\ 
& $720$ & $\mbf{0.219}_{\pm0.000}$ & $\underline{0.230}_{\pm0.005}$ & $0.238_{\pm 0.016}$ & $0.242_{\pm 0.015}$& -&- &$0.351$ & $0.236$ & $0.233$ \\
\midrule[\thick pt]
\multirow{4}{*}{\rotatebox[origin=c]{90}{Exchange}} 
& $96$ & $0.161_{\pm0.007}$ & ${0.233}_{\pm0.016}$ & $0.292_{\pm0.045}$ & $0.343_{\pm0.082}$ &$\mbf{0.086}$&$\underline{0.088}$& $0.847$ & $0.197$ & $0.139$ \\
& $192$ & $0.246_{\pm0.009}$ & $0.342_{\pm0.031}$ & $0.372_{\pm0.035}$ & $0.342_{\pm0.031}$ &$\underline{0.177}$&$ \mbf{0.176}$&$1.204$ & $0.300$ & $0.256$\\
& $336$ & $0.368_{\pm0.006}$ & $0.474_{\pm0.014}$ & $0.494_{\pm0.033}$ & $0.484_{\pm0.062}$ &$\underline{0.331}$&$\mbf{0.301}$&$1.672$ & $0.509$ & $0.426$\\ 
& $720$ & $1.003_{\pm0.018}$ & $1.078_{\pm0.179}$ & $1.323_{\pm0.192}$ & $1.204_{\pm0.028}$&$\mbf{0.847}$ &$\underline{0.901}$ &$2.478$ & $1.447$ & $1.090$\\
\midrule[\thick pt]
\multirow{4}{*}{\rotatebox[origin=c]{90}{Traffic}} 
& $96$ & $\underline{0.407}_{\pm0.001}$ & $0.409_{\pm0.016}$ & $0.420_{\pm0.041}$ & $0.409_{\pm0.016}$ &$\mbf{0.395}$&$0.462$& $0.733$ & $0.597$ & $0.576$ \\
& $192$ & $\mbf{0.415}_{\pm0.005}$ & $0.433_{\pm0.009}$ & $0.441_{\pm0.039}$ & $0.637_{\pm0.444}$ &$\underline{0.417}$&$0.466$& $0.777$ & $0.607$ & $0.610$ \\
& $336$ & $\mbf{0.421}_{\pm0.001}$ & $\underline{0.424}_{\pm0.000}$ & $0.501_{\pm0.154}$ & $0.747_{\pm0.277}$ &$0.433$&$0.482$& $0.776$ & $0.623$ & $0.608$ \\ 
& $720$ & $\mbf{0.456}_{\pm0.003}$ & $0.488_{\pm0.028}$ & $0.468_{\pm0.021}$ & $0.688_{\pm0.287}$ &$\underline{0.467}$&$0.514$& $0.827$ & $0.639$ & $0.621$ \\
\midrule[\thick pt]
\multirow{4}{*}{\rotatebox[origin=c]{90}{Weather}} 
& $96$ & $\underline{0.197}_{\pm0.001}$ & $\mbf{0.189}_{\pm0.003}$ & $0.227_{\pm0.012}$ & $0.214_{\pm0.004}$ &$0.174$&$0.177$& $0.354$ & $0.249$ & $0.238$ \\
& $192$ & $\underline{0.235}_{\pm0.000}$ & $\mbf{0.228}_{\pm0.004}$ & $0.256_{\pm0.018}$ & $0.231_{\pm0.003}$ &$0.221$&$0.225$& $0.419$ & $0.325$ & $0.275$ \\
& $336$ & $\underline{0.276}_{\pm0.001}$ & $\mbf{0.271}_{\pm0.001}$ & $0.278_{\pm0.001}$ & $0.279_{\pm0.007}$ &$0.278$&$0.278$& $0.583$ & $0.351$ & $0.339$ \\ 
& $720$ & $\underline{0.334}_{\pm0.000}$ & $\mbf{0.331}_{\pm0.001}$ & $0.353_{\pm0.002}$ & $0.343_{\pm0.024}$& $ 0.358$& $0.354$& $0.916$ & $0.415$ & $0.389$\\
\midrule[\thick pt]%
\multicolumn{3}{c}{\textcolor{bluerow}{\textbf{Overall MSE improvement}}}  &\textcolor{bluerow}{$\mbf{5.25\%}$} & \textcolor{bluerow}{$\mbf{16.96\%}$} & \textcolor{bluerow}{$\mbf{14.33\%}$} & \textcolor{bluerow}{$\mbf{3.94\%}$}
& \textcolor{bluerow}{$\mbf{11.13\%}$}
& \textcolor{bluerow}{$\mbf{72.20\%}$} & \textcolor{bluerow}{$\mbf{22.65\%}$} & \textcolor{bluerow}{$\mbf{12.36\%}$}\\
\bottomrule[\thick pt]%
\end{tabular}
}
\end{table*}

\paragraph{Smoother loss landscape.} The introduction of SAM in the training of \model{} makes its loss smoother than that of \transformer{}. We illustrate this in Figure~\ref{fig:sharpness_2_datasets} by comparing the values of $\lambda_\mathrm{max}$ for \transformer{} and \model{} after training on ETTh1 and Exchange.
Our observations reveal that \transformer{} exhibits considerably higher sharpness, while \model{} has a desired behavior with a loss landscape sharpness that is an order of magnitude smaller.

\paragraph{Improved robustness.}
\model{} demonstrates robustness against random initialization. Figure~\ref{fig:stabilized_performance_3_datasets} illustrates the test MSE distribution of \model{} and \transformer{} across $5$ different seeds on ETTh1 and Exchange with a prediction horizon of $H=96$. \model{} consistently maintains performance stability across different seed choices, while \transformer{} exhibits significant variance and, thus, a high dependency on weight initialization. This observation holds across all datasets and prediction horizons as shown in Appendix~\ref{app:stabilized_performance}.

\subsection{Qualitative Benefits of Our Approach} 

\paragraph{Computational efficiency.} \model{} is computationally more efficient than \tsmixer{} and usual transformer-based approaches, benefiting from a shallow lightweight implementation, i.e., a single layer with one attention head. The number of parameters of \model{} and \tsmixer{} is detailed in Appendix Table~\ref{tab:model_params}. We observe that, on average, \model{} has $\sim4$ times fewer parameters than \tsmixer{}, which makes this approach even more remarkable. Importantly, \tsmixer{} itself is recognized as a computationally efficient architecture compared to the transformer-based baselines~\citep[Table 6]{chen2023tsmixer}.

\paragraph{Fewer hyperparameters and versatility.}
\model{} requires minimal hyperparameters tuning, contrary to other baselines, including \tsmixer{} and \texttt{FEDformer}. In particular, \model{}'s architecture remains the same for all our experiments (see Appendix~\ref{app:arch_train} for details), while \tsmixer{} varies in terms of the number of residual blocks and feature embedding dimensions, depending on the dataset. This versatility also comes with better robustness to the prediction horizon $H$.
In Appendix~\ref{app:sensitivity_horizon} Figure~\ref{fig:sensitivity_horizon}, we display the evolution forecasting accuracy on all datasets for $H \in \{96, 192, 336, 720\}$ for \model{} and \tsmixer{} (trained with SAM). We observe that \model{} consistently outperforms its best competitor \tsmixer{} (trained with SAM) for all horizons.

\begin{figure*}[!ht]
\centering
\begin{subfigure}{.49\textwidth}
  \centering
  \includegraphics[width=\linewidth]{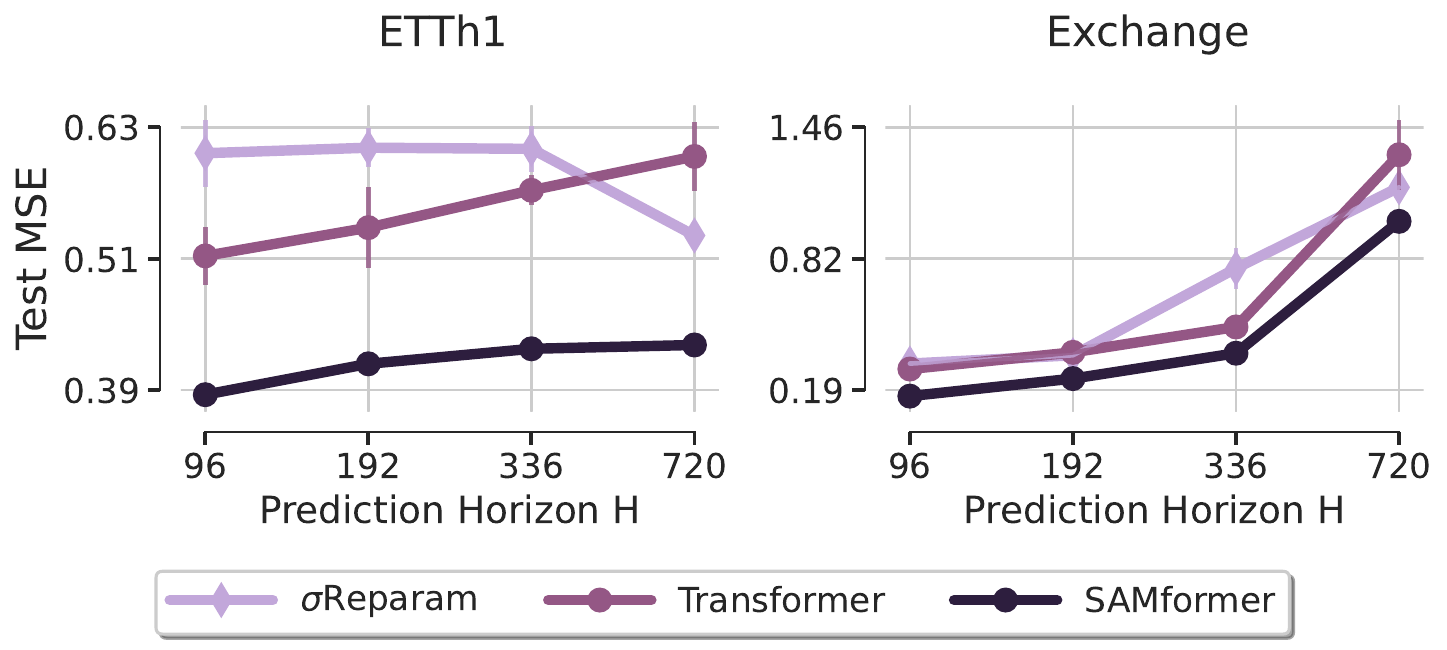}
  \caption{Comparison of \transformer{}, \sreparam{} and \model{}.}
  \label{fig:sreparam_failure}
\end{subfigure}%
\begin{subfigure}{.49\textwidth}
  \centering
  \includegraphics[width=\linewidth]{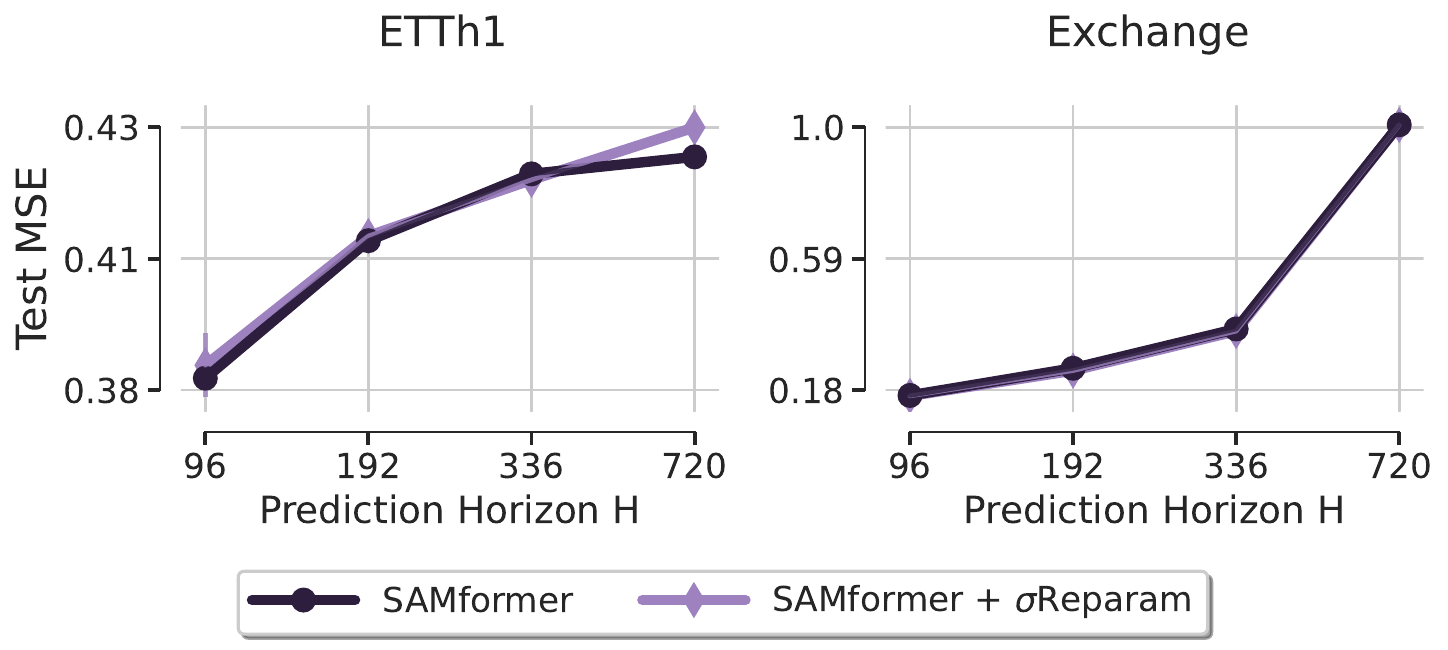}
  \caption{Comparison of \model{} and \model{} + \sreparam{}.}
\label{fig:sreparam_sam_failure}
\end{subfigure}
\caption{\textbf{Suboptimality of \sreparam{}.} (a) \sreparam{} alone does not bring improvement on \transformer{} and is clearly outperformed by \model{}. Combining \sreparam{} with \model{} does not bring significant improvement but heavily increases the training time (see Figure~\ref{fig:computational_time_sreparam}).}
\label{fig:sreparam_sreparam_sam_failure}
\end{figure*}

\paragraph{Better attention.} We display the attention matrices after training on Weather with the prediction horizon $H=96$ for \transformer{}, \model{} and \transformer{} + \sreparam{} in Figure~\ref{fig:attention_matrix_3_models}. We note that \transformer{} excludes self-correlation between features, having low values on the diagonal, while \model{} strongly promotes them. This pattern is reminiscent of~\citet{he2023deepshortcut} and \citet{trockman2023mimetic}: both works demonstrated the importance of diagonal patterns in attention matrices for signal propagation in transformers used in NLP and computer vision. Our experiments reveal that these insights also apply to time-series forecasting.
Note that freezing the attention to $\bA(\bX) = \bI_D$ is largely outperformed by \model{} as shown in Table~\ref{tab:identity_attention}, Appendix~\ref{app:choice_implementation}, which confirms the importance of learnable attention. The attention matrix given by \sreparam{} at Figure~\ref{fig:attention_matrix_3_models} has almost equal rows, leading to rank collapse.
In Figure~\ref{figure:nuclear.norm}, we display the distributions of nuclear norms of attention matrices after training \transformer{}, \model{} and \sreparam{}.
We observe that \sreparam{} heavily penalizes the nuclear norms of the attention matrix, which is coherent with Proposition~\ref{thm:upper_bound_nuclear_norm}. In contrast, \model{} maintains it above \transformer{}, thus improving the expressiveness of attention.

\subsection{\model{} vs \texttt{MOIRAI}}
In this section, we show that despite its simplicity, \model{} is a strong baseline competing not only with the dedicated time series methods (Table~\ref{tab:all_results}), such as \tsmixer{} but also with the biggest existing time series forecasting foundation model \texttt{MORAI}~\citep{woo2024unified} that was trained on the largest pretraining corpus \texttt{LOTSA} with nearly $\mbf{27}$ \textbf{billion of samples}. \texttt{MOIRAI} was provided in three sizes: small ($14$ million parameters), base ($91$ million) and large ($314$ million). In Table~\ref{tab:comparison_moirai}, we see that \model{} is on-par with \texttt{MOIRAI} on most datasets, superior on $3$ on them, and overall improves \texttt{MOIRAI} by at least $\mbf{1.1\%}$ and up to $\mbf{7.6\%}$. This comparison highlights again the fact that \model{} shows impressive performance, globally superior to its competitors while having much less trainable parameters. 

\setlength{\tabcolsep}{0.3em}
\begin{table}[!t] 
\centering
\caption{Comparison performance of \textcolor{bluerow}{\textbf{\model{}}} and \texttt{MOIRAI}~\citep{woo2024unified} for multivariate long-term forecasting. We display the test MSE averaged over horizons $\{96, 192, 336, 720\}$. \textbf{Best} results are in bold, \underline{second best} are underlined.}
\label{tab:comparison_moirai}
\scalebox{0.75}{
\begin{tabular}{ccccc}
\toprule[\thick pt]%
\multicolumn{1}{c}{\multirow{2}{*}{Dataset}} & \multicolumn{1}{c}{Full-shot} & \multicolumn{3}{c}{Zero-shot~\citep{woo2024unified}.}\\ 
\cmidrule(r{10pt}l{5pt}){2-2} \cmidrule(r{10pt}l{5pt}){3-5} 
 \multicolumn{1}{c}{} & \multicolumn{1}{c}{\textcolor{bluerow}{\textbf{\model{}}}} &  \multicolumn{1}{c}{$\texttt{MOIRAI}_\texttt{Small}$} & \multicolumn{1}{c}{$\texttt{MOIRAI}_\texttt{Base}$} & \multicolumn{1}{c}{$\texttt{MOIRAI}_\texttt{Large}$} \\
\midrule[\thick pt]
ETTh1 & $0.410$ & $\underline{0.400}$ & $\mbf{0.434}$ & $0.510$\\
\midrule[\thick pt]
ETTh2 & $\underline{0.344}$ & $\mbf{0.341}$ & $0.345$ & $0.354$\\
\midrule[\thick pt]
ETTm1 & $\mbf{0.373}$ & $0.448$ & $\underline{0.381}$ & $0.390$ \\
\midrule[\thick pt]
ETTm2 & $\mbf{0.269}$ & $0.300$ & $\underline{0.272}$ & $0.276$ \\
\midrule[\thick pt]
Electricity & $\mbf{0.181}$ & $0.233$ & $\underline{0.188}$ & $\underline{0.188}$\\
\midrule[\thick pt]
Weather & $0.260$ & $\underline{0.242}$ & $\mbf{0.238}$ & $0.259$ \\
\midrule[\thick pt]%
\multicolumn{2}{c}{\textcolor{bluerow}{\textbf{Overall MSE improvement}}}  &\textcolor{bluerow}{$\mbf{6.9\%}$} & \textcolor{bluerow}{$\mbf{1.1\%}$} & \textcolor{bluerow}{$\mbf{7.6\%}$} \\
\bottomrule[\thick pt]%
\end{tabular}
}
\end{table}

\subsection{Ablation Study and Sensitivity Analysis}

\paragraph{Choices of implementation.}
We empirically compared our architecture, which is channel-wise attention (Eq.~\eqref{eq:transformer_model}), with temporal-wise attention. Table~\ref{tab:temporal_attention} of Appendix~\ref{app:choice_implementation} shows the superiority of our approach in the considered setting. We conducted our experiments with Adam~\citep{KingBa15}, the de-facto optimizers for transformers~\citep{ahn2023linear, pan2022toward, zhou2022fedformer, haoyi2021informer, chen2022vitwithsam}.
We provide an in-depth ablation study in Appendix~\ref{app:sensitivity_optim} that motivates this choice. As expected \citep{ahn2023linear, liu2020understanding, pan2022toward, zhang2020adaptattention}, SGD~\citep{nesterov1983sgd} fails to converge and AdamW~\citep{loshchilov2018decoupled} leads to similar performance but is very sensitive to the choice of the weight decay strength.

\paragraph{Sensitivity to the neighborhood size $\rho$.} 
The test MSE of \model{} and \tsmixer{} is depicted in Figure~\ref{fig:sensitivity_rho} of Appendix~\ref{app:sensitivity_rho} as a function of the neighborhood size $\rho$. It appears that \tsmixer{}, with its quasi-linear architecture, exhibits less sensitivity to $\rho$ compared to \model{}. This behavior is consistent with the understanding that, in linear models, the sharpness does not change with respect to $\rho$, given the constant nature of the loss function's Hessian. Consequently, \tsmixer{} benefits less from changes in $\rho$ than \model{}. Our observations consistently show that a sufficiently large $\rho$, generally above $0.7$ enables \model{} to achieve lower MSE than \tsmixer{}.

\paragraph{SAM vs \sreparam{}.} We mentioned previously that \sreparam{} doesn't improve the performance of a transformer on a simple toy example, although it makes it comparable to the performance of a transformer with fixed random attention. To further show that \sreparam{} doesn't provide an improvement on real-world datasets, we show in Figure~\ref{fig:sreparam_failure} that on ETTh1 and Exchange, \sreparam{} alone fails to match \model{}'s improvements, even underperforming \transformer{} in some cases. A potential improvement may come from combining SAM and \sreparam{} to smooth a rather sparse matrix obtained with SAM. However, as Figure~\ref{fig:sreparam_sam_failure} illustrates, this combination does not surpass the performance of using SAM alone. Furthermore, combining SAM and \sreparam{} significantly increases training time and memory usage, especially for larger datasets and longer horizons (see Appendix Figure~\ref{fig:computational_time_sreparam}), indicating its inefficiency as a method.

\section{Discussion and Future Work}
In this work, we demonstrated how simple transformers can reclaim their place as state-of-the-art models in long-term multivariate series forecasting from their MLP-based competitors. Rather than concentrating on new architectures and attention mechanisms, we analyzed the current pitfalls of transformers in this task and addressed them by carefully designing an appropriate training strategy. Our findings suggest that even a simple shallow transformer has a very sharp loss landscape which makes it converge to poor local minima. We analyzed popular solutions proposed in the literature to address this issue and showed which of them work or fail. Our proposed \model{}, optimized with sharpness-aware minimization, leads to a substantial performance gain compared to the existing forecasting baselines, including the current largest foundation model \texttt{MOIRAI}, and benefits from a high versatility and robustness across datasets and prediction horizons. Finally, we also showed that channel-wise attention in time series forecasting can be more efficient -- both computationally and performance-wise -- than temporal attention commonly used previously. We believe that this surprising finding may spur many further works building on top of our simple architecture to improve it even further.

\section*{Acknowledgements}
The authors would like to thank the machine learning community for providing open-source baselines and datasets. The authors thank the anonymous reviewers and meta-reviewers for their time and constructive feedback. This work was enabled thanks to open-source software such as Python~\citep{van1995python}, PyTorch~\citep{pytorch}, TensorFlow~\citep{tensorflow2015-whitepaper}, Scikit-learn~\citep{scikit-learn} and Matplotlib~\citep{hunter2007matplotlib}.

\section*{Impact Statement}
This paper presents work whose goal is to advance the field of Machine Learning. There are many potential societal consequences of our work, none of which we feel must be specifically highlighted here.

\bibliography{references}
\bibliographystyle{icml2024}

\newpage
\appendix
\onecolumn
\textbf{\LARGE Appendix}
\paragraph{Roadmap.} In this appendix, we provide the detailed experimental setup in Section~\ref{app:exp_setup}, additional experiments in Section~\ref{app:add_exp}, and a thorough ablation study and sensitivity analysis in Section~\ref{app:ablation_sensitivity}. Additional background knowledge is available in Section~\ref{app:additional_background} and proofs of the main theoretical results are provided in Section~\ref{app:proofs}. We display the corresponding table of contents below.

\addtocontents{toc}{\protect\setcounter{tocdepth}{2}}

\renewcommand*\contentsname{\Large Table of Contents}

\tableofcontents
\clearpage

\section{Experimental Setup}
\label{app:exp_setup}

\subsection{Architecture and Training Parameters}
\label{app:arch_train}
\paragraph{Architecture.} We follow~\citet{chen2023tsmixer, nie2023patchtst}, and to ensure a fair comparison of baselines, we apply the reversible instance normalization (\texttt{RevIN}) of~\citet{kim2021reversible} (see Appendix~\ref{app:revin} for more details). The network used in \model{} and \transformer{} is a simplified one-layer transformer with one head of attention and without feed-forward. Its neural network function follows Eq.~\eqref{eq:transformer_model}, while \texttt{RevIN} normalization and denormalization are applied respectively before and after the neural network function, see Figure~\ref{fig:samformer-diagram}. We display the inference step of \model{} in great detail in Algorithm~\ref{alg:samformer_algorithm}. For the sake of clarity, we describe the application of the neural network function sequentially on each element of the batches but in practice, the operations are parallelized and performed batch per batch. For \model{} and \transformer{}, the dimension of the model is $d_\mathrm{m} = 16$ and remains the same in all our experiments. For \tsmixer{}, we used the official implementation that can be found at \href{https://github.com/google-research/google-research/tree/master/tsmixer}{here}. 

\begin{algorithm}[!ht] 
\caption{Architecture of the network used in \model{} and \transformer{}}
\label{alg:samformer_algorithm}
    \textbf{Parameters:} Batch size $bs$, input length $L$, prediction horizon $H$, dimension of the model $d_\mathrm{m}$. \\
    \textbf{Network trainable parameters:} $\mbf{W}_Q \in \RR^{L \times d_\mathrm{m}}, \mbf{W}_K \in \RR^{L \times d_\mathrm{m}}$, $\mbf{W}_V \in \RR^{L \times d_\mathrm{m}}$, $\mbf{W}_O \in \RR^{d_\mathrm{m} \times L}$, $\mbf{W} \in \RR^{L \times H}$. \\
    \textbf{\texttt{RevIN} trainable parameters}: $\bm{\beta}, \bm{\gamma}$. \\
    \textbf{Input:} Batch of $bs$ input sequences $\mbf{X} \in \mathbb{R}^{D \times L}$ arranged in a tensor $\mbf{B}_\mathrm{in}$ of dimension $bs \times L \times D$. \\
    \textbf{\texttt{RevIN} normalization:} $\mbf{X} \gets \tilde{\mbf{X}}$ following Eq.~\eqref{eq:revin_first_step}. The output is a tensor $\tilde{\mbf{B}}_\mathrm{in}$ of dimension $bs \times L \times D$. \\
    \textbf{Transposition of the batch:} $\tilde{\mbf{B}}_\mathrm{in}$ is reshaped in dimension $bs \times D \times L$. \\ 
    \textbf{Applying the neural network of Eq.~\eqref{eq:transformer_model}:} \\ 
    \For {each $\tilde{\mbf{X}} \in \tilde{\mbf{B}}_\mathrm{in}$}{
    \textbf{1. Attention layer} \\
    Rescale the input with the attention matrix (Eq.~\eqref{eq:attention_matrix}). \\
    The output $\bA(\tilde{\mbf{X}})\tilde{\mbf{X}}\bW_V\bW_O$ is of dimension $D \times L$ \\
    \textbf{2. Skip connection} \\
    Sum the input $\tilde{\mbf{X}}$ and the output of the attention layer. \\ 
    The output $\tilde{\mbf{X}} + \bA(\tilde{\mbf{X}})\tilde{\mbf{X}}\bW_V\bW_O$ is of dimension $D \times L$. \\
    \textbf{3. Linear layer} \\
    Apply a linear layer on the output of the skip connection. \\
    The output $\tilde{\mbf{Y}} = \mleft[\tilde{\mbf{X}} + \bA(\tilde{\mbf{X}})\tilde{\mbf{X}}\bW_V\bW_O\mright]\bW$ is of dimension $D \times H$. \\
    Unnormalized predictions are arranged in a tensor $\tilde{\mbf{B}}_\mathrm{out}$ of dimension $bs \times D \times H$. \\
    }
    \textbf{Transposition of the batch:} $\tilde{\mbf{B}}_\mathrm{out}$ is reshaped in dimension $bs \times H \times D$. \\ 
    \textbf{\texttt{RevIN} denormalization:} $\tilde{\mbf{Y}} \gets \hat{\mbf{Y}}$ following Eq.~\eqref{eq:revin_second_step}. 
    \\
    \textbf{Output:} Batch of $bs$ prediction sequences $\hat{\mbf{Y}} \in \mathbb{R}^{D \times H}$ arranged in a tensor $\hat{\mbf{B}}_\mathrm{out}$ of dimension $bs \times H \times D$. \\
\end{algorithm}

\paragraph{Training parameters.} For all of our experiments, we train our baselines (\model{}, \transformer{}, \tsmixer{} with SAM, \tsmixer{} without SAM) with the Adam optimizer \citep{KingBa15}, a batch size of $32$, a cosine annealing scheduler \citep{loshchilov2017sgdr} and the learning rates summarized in Table~\ref{tab:lr}. 
\setlength{\tabcolsep}{0.2em}
\begin{wraptable}[6]{r}{0.4\textwidth}
\centering
    \caption{Learning rates used in our experiments. ETT designs ETTh1/ETTh2/ETTm1/ETTm2.}
    \label{tab:lr}
    \scalebox{0.8}{
    \begin{tabular}{cccccc}
        \toprule
         Dataset & ETT & Electricity & Exchange & Traffic & Weather\\
         \midrule
         Learning rate & $0.001$ & $0.0001$ & $0.001$ & $0.0001$ & $0.0001$ \\
         \bottomrule
    \end{tabular}
    }
\end{wraptable}
For \model{} and \tsmixer{} trained with SAM, the values of neighborhood size $\rho^*$ used are reported in Table~\ref{tab:optimal_rho}. The training/validation/test split is $12/4/4$ months on the \texttt{ETT} datasets and $70\%/20\%/10\%$ on the other datasets. We use a look-back window $L=512$ and use a sliding window with stride $1$ to create the sequences. The training loss is the MSE on the multivariate time series (Eq.~\eqref{eq:training_loss}).
Training is performed during $300$ epochs and we use early stopping with a patience of $5$ epochs. For each dataset, baselines, and prediction horizon $H \in \{96, 192, 336, 720\}$, each experiment is run $5$ times with different seeds, and we display the average and the standard deviation of the test MSE and MAE over the $5$ trials. 

\subsection{Datasets}
\label{app:datasets}
We conduct our experiments on $8$ publicly available datasets of real-world time series, widely used for multivariate long-term forecasting~\citep{wu2021autoformer, chen2023tsmixer, nie2023patchtst}. The $4$ Electricity Transformer Temperature datasets ETTm1, ETTm2, ETTh1, and ETTh2~\citep{haoyi2021informer} contain the time series collected by electricity transformers from July 2016 to July 2018. Whenever possible, we refer to this set of $4$ datasets as ETT. Electricity~\citep{electricity} contains the time series of electricity consumption from $321$ clients from 2012 to 2014. Exchange~\citep{lai2018modeling} contains the time series of daily exchange rates between $8$ countries from 1990 to 2016. Traffic~\citep{traffic} contains the time series of road occupancy rates captured by $862$ sensors from January 2015 to December 2016. Last but not least, Weather~\citep{weather} contains the time series of meteorological information recorded by 21 weather indicators in 2020. It should be noted that Electricity, Traffic, and Weather are large-scale datasets. The ETT datasets can be downloaded \href{https://github.com/zhouhaoyi/Informer2020}{here} while the $4$ other datasets can be downloaded \href{https://github.com/thuml/Autoformer}{here}. Table~\ref{tab:dataset_description} sums up the characteristics of the datasets used in our experiments. 
\begin{table}[!t]
\centering
\caption{Neighborhood size $\rho^*$ at which \model{} and \tsmixer{} achieve their best performance on the benchmarks.}
\label{tab:optimal_rho}
\setlength{\tabcolsep}{7pt} 
\scalebox{0.9}{
\begin{tabular}{cccccccccc}
\toprule[\thick pt]%
\multicolumn{1}{c}{H} & \multicolumn{1}{c}{Model} & \multicolumn{1}{c}{ETTh1} & \multicolumn{1}{c}{ETTh2} & \multicolumn{1}{c}{ETTm1} & \multicolumn{1}{c}{ETTm2} & \multicolumn{1}{c}{Electricity} & \multicolumn{1}{c}{Exchange} & \multicolumn{1}{c}{Traffic} & \multicolumn{1}{c}{Weather} \\ 
\midrule[\thick pt]
\multirow{2}{*}{96} & \model{} & 0.5 & 0.5 & 0.6 & 0.2 & 0.5 &0.7&0.8&0.4\\
 & \tsmixer{} & 1.0&0.9&1.0&1.0&0.9&1.0&0.0&0.5\\
\midrule[\thick pt]%
 \multirow{2}{*}{192} & \model{} & 0.6&0.8&0.9&0.9&0.6&0.8&0.1&0.4\\
 & \tsmixer{} & 0.7&0.1&0.6&1.0&1.0&0.0&0.9&0.4\\
\midrule[\thick pt]%
 \multirow{2}{*}{336} & \model{} &0.9&0.6&0.9&0.8&0.5&0.5&0.5&0.6\\
 & \tsmixer{} & 0.7&0.0&0.7&1.0&0.4&1.0&0.6&0.6\\
\midrule[\thick pt]%
 \multirow{2}{*}{720} & \model{} & 0.9&0.8&0.9&0.9&1.0&0.9&0.7&0.5\\
 & \tsmixer{} & 0.3&0.4&0.5&1.0&0.9&0.1&0.9&0.3 \\
\bottomrule[\thick pt]%
\end{tabular}
}
\end{table}

\begin{table}[htbp]
    \centering
    \caption{Characteristics of the multivariate time series datasets used in our experiments with various sizes and dimensions.}
    \label{tab:dataset_description}
    \scalebox{1}{
    \begin{tabular}{lcccccc}
        \toprule
         Dataset &ETTh1/ETTh2 & ETTm1/ETTm2 & Electricity & Exchange & Traffic & Weather\\
         \midrule
         \# features & $7$ & $7$ & $321$ & $8$ & $862$ & $21$\\
         \# time steps & $17420$ & $69680$ & $26304$ & $7588$ & $17544$ & $52696$ \\
         Granularity & 1 hour & 15 minutes & 1 hour & 1 day & 1 hour & 10 minutes \\
         \bottomrule
    \end{tabular}
    }
\end{table}

\subsection{More Details on the Baselines}
\label{app:baselines}
As stated above, we conducted all our experiments with a look-back window $L=512$ and prediction horizons $H \in \{96, 192, 336, 720\}$. Results reported in Table~\ref{tab:all_results} from \model{}, \tsmixer{}, and \transformer{} \textbf{\textit{come from our own experiments}}, conducted over $5$ runs with $5$ different seeds. The reader might notice that the results of \texttt{TSMixer} without SAM slightly differ from the ones reported in the original paper \citep{chen2023tsmixer}. It comes from the fact that the authors reported results from a single seed, while we report average performance with standard deviation on multiple runs for a better comparison of methods. We perform a Student's t-test in Table~\ref{tab:significance_test} for a more thorough comparison of \model{} and \tsmixer{} with SAM. It should be noted that, unlike our competitors including \tsmixer{}, the architecture of \texttt{SAMformer} remains the same for all the datasets. This highlights the robustness of our method and its advantage as no heavy hyperparameter tuning is required. For a fair comparison of models, we also report results from other baselines in the literature that we did not run ourselves. For \texttt{Informer}~\cite{haoyi2021informer}, \texttt{Autoformer}~\citep{wu2021autoformer}, and \texttt{Fedformer} \citep{zhou2022fedformer}, the results on all datasets, except Exchange, are reported from \citet{chen2023tsmixer}. Results on the Exchange dataset for those $5$ baselines come from the original corresponding papers and hence refer to the models without \texttt{RevIN}. For \texttt{iTransformer}~\citep{liu2024itransformer} and \texttt{PacthTST}~\citep{nie2023patchtst}, results are reported from~\citet{liu2024itransformer}. Those baselines also make use of RevIn~\citep{kim2021reversible}. It should be noted that \texttt{iTransformer}~\citep{liu2024itransformer} uses both temporal and channel-wise attention. Our large-scale experimental evaluation ensures a comprehensive and comparative analysis across various established models in multivariate long-term time series forecasting.

\section{Additional Experiments}
\label{app:add_exp}
In this section, we provide additional experiments to showcase, quantitatively and qualitatively, the superiority of our approach. 

\subsection{MAE Results}
\label{app:mae_results}
In this section, we provide the performance comparison of the different baselines with the Mean Absolute Error (MAE). We display the results in Table~\ref{tab:all_results_mae}. The conclusion is similar to the one made in the main paper in Table~\ref{tab:all_results} and confirms the superiority of \model{} compared to its competitors, including very recent baselines like \tsmixer{}~\citep{chen2023tsmixer}, \texttt{iTransformer}~\citep{liu2024itransformer} and \texttt{PatchTST}~\citep{nie2023patchtst}. 
\vspace{10pt}
\setlength{\tabcolsep}{0.3em}
\begin{table*}[!ht] 
\centering
\caption{Performance comparison between our model (\textcolor{bluerow}{\textbf{\model}}) and baselines for multivariate long-term forecasting with different horizons $H$. Results marked with \say{$^\dagger$} are obtained from~\citet{liu2024itransformer} and those marked with \say{$^*$} are obtained from \citet{chen2023tsmixer}, along with the publication year of the respective methods. Transformer-based models are abbreviated by removing the ``former" part of their name. We display the average test MAE with standard deviation obtained on $5$ runs with different seeds. \textbf{Best} results are in bold, \underline{second best} are underlined.}
\label{tab:all_results_mae}
\scalebox{0.83}{
\begin{tabular}{ccccccccccc}
\toprule[\thick pt]%
\multicolumn{1}{c}{\multirow{3}{*}{Dataset}} & \multicolumn{1}{c}{\multirow{3}{*}{$H$}} & \multicolumn{2}{c}{with SAM} & \multicolumn{7}{c}{without SAM}\\ 
\cmidrule(r{10pt}l{5pt}){3-4} \cmidrule(r{10pt}l{5pt}){5-11} 
 \multicolumn{1}{c}{} & \multicolumn{1}{c}{} & \multicolumn{1}{c}{\textcolor{bluerow}{\textbf{\model}}} &  \multicolumn{1}{c}{\tsmixer{}} & \multicolumn{1}{c}{\transformer{}} & \multicolumn{1}{c}{\tsmixer{}} & \multicolumn{1}{c}{$\texttt{iTrans}^\dagger$} &\multicolumn{1}{c}{$\texttt{PatchTST}^\dagger$} &\multicolumn{1}{c}{$\texttt{In}^*$} & \multicolumn{1}{c}{$\texttt{Auto}^*$} & \multicolumn{1}{c}{$\texttt{FED}^*$}\\
\multicolumn{1}{c}{} & \multicolumn{1}{c}{} & \multicolumn{1}{c}{-} &  \multicolumn{1}{c}{-} & \multicolumn{1}{c}{-} & \multicolumn{1}{c}{\small 2023} & \multicolumn{1}{c}{\small 2024} &\multicolumn{1}{c}{\small 2023} &\multicolumn{1}{c}{\small 2021} & \multicolumn{1}{c}{\small 2021} & \multicolumn{1}{c}{\small 2022}\\
\midrule[\thick pt]
\multirow{4}{*}{\rotatebox[origin=c]{90}{ETTh1}} & $96$ & $\mbf{0.402}_{\pm 0.001}$ & $0.408_{\pm 0.001}$ & $0.619_{\pm 0.203}$ & $0.414_{\pm 0.004}$ & $\underline{0.405}$& $0.419$& 0.769 & 0.446 & $0.415$ \\
& $192$ & $\mbf{0.418}_{\pm 0.001}$ & $\underline{0.426}_{\pm 0.002}$ & $0.513_{\pm 0.024}$ & $0.428_{\pm 0.001}$ &$0.436$ &$0.445$&0.786 & 0.457 & 0.446 \\
& $336$ & $\mbf{0.425}_{\pm 0.000}$ & $\underline{0.434}_{\pm 0.001}$ & $0.529_{\pm 0.008}$ & $\underline{0.434}_{\pm 0.001}$ & $0.458$&$0.466$&0.784 & 0.487 & 0.462 \\ 
& $720$ & $\mbf{0.449}_{\pm 0.002}$ & $\underline{0.459}_{\pm 0.004}$ & $0.553_{\pm 0.021}$ & $0.506_{\pm 0.064}$ & $0.491$&$0.488$&0.857 & 0.517 & 0.492 \\
\midrule[\thick pt]
\multirow{4}{*}{\rotatebox[origin=c]{90}{ETTh2}} & $96$ & $0.358_{\pm 0.002}$ & $0.367_{\pm 0.002}$ & $0.416_{\pm 0.025}$ & $\underline{0.367}_{\pm 0.003}$ &$\underline{0.349}$ &$\mbf{0.348}$&0.952 & 0.368 & 0.374 \\
& $192$ & $\mbf{0.386}_{\pm 0.003}$ & $\underline{0.393}_{\pm 0.001}$ & $0.435_{\pm 0.019}$ & $0.395_{\pm 0.003}$ &$0.400$ &$0.400$&1.542 & 0.434 & 0.446 \\
& $336$ & $\mbf{0.395}_{\pm 0.002}$ & $\underline{0.404}_{\pm 0.004}$ & $0.434_{\pm 0.014}$ & $\underline{0.404}_{\pm 0.002}$ &$0.432$ &$0.433$&  1.642 & 0.479 & 0.447 \\ 
& $720$ & $\mbf{0.428}_{\pm 0.001}$ & $\underline{0.435}_{\pm 0.002}$ & $0.448_{\pm 0.006}$ & $0.441_{\pm 0.005}$ &$0.445$ &$0.446$&  1.619 & 0.490 & 0.469 \\
\midrule[\thick pt]
\multirow{4}{*}{\rotatebox[origin=c]{90}{ETTm1}} & $96$ & $\mbf{0.363}_{\pm0.001}$ & $\mbf{0.363}_{\pm0.001}$ & $0.395_{\pm 0.024}$ & $0.371_{\pm 0.002}$ & $0.368$&$ 0.367$& 0.560 & 0.492 & $0.390$ \\
& $192$ & $\mbf{0.378}_{\pm0.003}$ & $\underline{0.381}_{\pm0.002}$ & $0.414_{\pm 0.027}$ & $0.384_{\pm 0.003}$ & $0.391$&$ 0.385$&0.619 & 0.495 & 0.415 \\
& $336$ & $\mbf{0.394}_{\pm0.001}$ & $\underline{0.397}_{\pm0.002}$ & $0.445_{\pm 0.009}$ & $0.399_{\pm 0.003}$ &$0.420$&$ 0.410$& 0.741 & 0.492 & 0.425 \\ 
& $720$ & $\mbf{0.418}_{\pm0.000}$ & $\underline{0.425}_{\pm0.001}$ & $0.456_{\pm 0.035}$ & $0.429_{\pm 0.002}$ &$0.459$&$ 0.439$& 0.845 & 0.493 & 0.458 \\
\midrule[\thick pt]
\multirow{4}{*}{\rotatebox[origin=c]{90}{ETTm2}} & $96$ & $0.274_{\pm0.010}$ & $0.284_{\pm0.004}$ & $0.290_{\pm 0.026}$ & $0.302_{\pm 0.013}$ &$ \underline{0.264}$&$\mbf{0.259}$& 0.462 & 0.293 & $0.271$ \\
& $192$ & $\underline{0.306}_{\pm0.001}$ & $0.320_{\pm0.001}$ & $0.347_{\pm 0.025}$ & $0.323_{\pm 0.005}$ & $0.309$&$\mbf{0.302}$&0.586 & 0.336 & $0.318$ \\
& $336$ & $\mbf{0.338}_{\pm0.001}$ & $0.350_{\pm0.001}$ & $0.360_{\pm 0.017}$ & $0.352_{\pm 0.003}$ &$0.348$&$ \underline{0.343}$& 0.871 & 0.379 & 0.364 \\ 
& $720$ & $\mbf{0.390}_{\pm0.001}$ & $0.402_{\pm0.002}$ & $0.424_{\pm 0.014}$ & $0.402_{\pm 0.003}$ &$0.407$&$ \underline{0.400}$& 1.267 & 0.419 & 0.420 \\
\midrule[\thick pt]
\multirow{4}{*}{\rotatebox[origin=c]{90}{\text{\small Electricity}}} & $96$ & $\mbf{0.252}_{\pm0.002}$ & $\underline{0.273}_{\pm0.001}$ & $0.288_{\pm 0.013}$ & $0.277_{\pm 0.003}$ &-&-& 0.393 & 0.313 & 0.302 \\
& $192$ & $\mbf{0.263}_{\pm0.001}$ & $\underline{0.292}_{\pm0.011}$ & $0.304_{\pm 0.033}$ & $0.304_{\pm 0.027}$ &-&-& 0.417 & 0.324 & 0.311 \\
& $336$ & $\mbf{0.277}_{\pm0.000}$ & $\underline{0.297}_{\pm0.007}$ & $0.315_{\pm 0.018}$ & $0.317_{\pm 0.018}$ &-&-& 0.422 & 0.327 & 0.328 \\ 
& $720$ & $\mbf{0.306}_{\pm0.000}$ & $\underline{0.321}_{\pm0.006}$ & $0.330_{\pm 0.014}$ & $0.333_{\pm 0.015}$ &-&-& 0.427 & 0.342 & 0.344 \\
\midrule[\thick pt]
\multirow{4}{*}{
\rotatebox[origin=c]{90}{Exchange}} & $96$ & $0.306_{\pm0.006}$ & ${0.363}_{\pm0.013}$ & $0.369_{\pm0.049}$ & $0.436_{\pm0.054}$ &$\underline{0.206}$&$\mbf{0.205}$& 0.752 & 0.323 & $0.276$ \\
& $192$ & $0.371_{\pm0.008}$ & $0.437_{\pm0.021}$ & $0.416_{\pm0.041}$ & $0.437_{\pm0.021}$ &$\mbf{0.299}$&$ \mbf{0.299}$&$0.895$ & $\underline{0.369}$ & $\underline{0.369}$ \\
& $336$ & $0.453_{\pm0.004}$ & $0.515_{\pm0.006}$ & $0.491_{\pm0.036}$ & $0.523_{\pm0.029}$ &$\underline{0.417}$&$ \mbf{0.397}$&1.036 & 0.524 & 0.464 \\ 
& $720$ & $0.750_{\pm0.006}$ & $0.777_{\pm0.064}$ & $0.823_{\pm0.040}$ & $0.818_{\pm0.007}$ &$\mbf{0.691}$&$\underline{0.714}$&1.310 & 0.941 & 0.800\\
\midrule[\thick pt]
\multirow{4}{*}{
\rotatebox[origin=c]{90}{Traffic}} & $96$ & $\underline{0.292}_{\pm0.001}$ & $0.300_{\pm0.020}$ & $0.306_{\pm0.033}$ & $\underline{0.300}_{\pm0.020}$ & $\mbf{0.268}$&$0.295$&0.410 & 0.371 & 0.359 \\
& $192$ & $\underline{0.294}_{\pm0.005}$ & $0.317_{\pm0.012}$ & $0.321_{\pm0.034}$ & $0.419_{\pm0.218}$ &$\mbf{0.276}$&$ 0.296$& 0.435 & 0.382 & 0.380 \\
& $336$ & $\underline{0.292}_{\pm0.000}$ & $0.299_{\pm0.000}$ & $0.348_{\pm0.093}$ & $0.501_{\pm0.163}$ &$ \mbf{0.283}$&$0.304$& 0.434 & 0.387 & 0.375 \\ 
& $720$ & $\underline{0.311}_{\pm0.003}$ & $0.344_{\pm0.026}$ & $0.325_{\pm0.023}$ & $0.458_{\pm0.159}$ &$\mbf{0.302}$&$0.322$& 0.466 & 0.395 & 0.375 \\
\midrule[\thick pt]
\multirow{4}{*}{
\rotatebox[origin=c]{90}{Weather}} & $96$ & $0.249_{\pm0.001}$ & $0.242_{\pm0.002}$ & $0.281_{\pm0.018}$ & $0.271_{\pm0.009}$ &$\mbf{0.214}$&$ \underline{0.218}$& 0.405 & 0.329 & 0.314 \\
& $192$ & $0.277_{\pm0.000}$ & $0.272_{\pm0.003}$ & $0.302_{\pm0.020}$ & $0.275_{\pm0.003}$ &$\mbf{0.254}$&$ \underline{0.259}$& 0.434 & 0.370 & 0.329 \\
& $336$ & $0.304_{\pm0.001}$ & $0.299_{\pm0.001}$ & $0.310_{\pm0.012}$ & $0.307_{\pm0.009}$ &$\mbf{0.296}$&$\underline{0.297}$& 0.543 & 0.391 & 0.377 \\ 
& $720$ & $\underline{0.342}_{\pm0.000}$ & $\mbf{0.341}_{\pm0.002}$ & $0.363_{\pm0.002}$ & $0.351_{\pm0.021}$ &$0.347$&$ 0.348$& 0.705 & 0.426 & 0.409 \\
\midrule[\thick pt]
\multicolumn{3}{c}{\textcolor{bluerow}{\textbf{Overall MAE improvement}}}  &\textcolor{bluerow}{$\mbf{3.99\%}$} & \textcolor{bluerow}{$\mbf{11.63\%}$} & \textcolor{bluerow}{$\mbf{9.60\%}$} &\textcolor{bluerow}{$\mbf{2.05\%}$}&\textcolor{bluerow}{$\mbf{2.75\%}$}& \textcolor{bluerow}{$\mbf{53.00\%}$} & \textcolor{bluerow}{$\mbf{15.67\%}$} & \textcolor{bluerow}{$\mbf{9.93\%}$}  \\
\bottomrule[\thick pt]%
\end{tabular}
}
\end{table*}

\subsection{Significance Test for \model{} and \tsmixer{} with SAM}
\label{app:significance_test}
In this section, we perform a Student t-test between \model{} and \tsmixer{} trained with SAM. It should be noted that \tsmixer{} with SAM significantly outperforms vanilla \tsmixer{}.
We report the results in Table~\ref{tab:significance_test}. 
We observe that the \model{} significantly improves upon \tsmixer{} trained with SAM on $7$ out of $8$ datasets. 

\begin{table}[h]
\centering
\caption{Significance test with Student's t-test and performance comparison between  \model{} and \tsmixer{} trained with SAM across various datasets and prediction horizons. We display the average and standard deviation of the test MSE obtained on $5$ runs ($\mathrm{mean}_{\pm \mathrm{std}}$). The performance of the best model is in \textbf{bold} when the improvement is statistically significant at the level $0.05$ ($\text{p-value} < 0.05$).}
\label{tab:significance_test}
\setlength{\tabcolsep}{7pt} 
\scalebox{0.8}{
\begin{tabular}{cccccccccc}
\toprule[\thick pt]%
\multicolumn{1}{c}{H} & \multicolumn{1}{c}{Model} & \multicolumn{1}{c}{ETTh1} & \multicolumn{1}{c}{ETTh2} & \multicolumn{1}{c}{ETTm1} & \multicolumn{1}{c}{ETTm2} & \multicolumn{1}{c}{Electricity} & \multicolumn{1}{c}{Exchange} & \multicolumn{1}{c}{Traffic} & \multicolumn{1}{c}{Weather} \\ 
\midrule[\thick pt]
\multirow{2}{*}{96} & \model{} & $\textbf{0.381}_{\pm0.003}$ & $\textbf{0.295}_{\pm0.002}$ & $0.329_{\pm0.001}$ & $\textbf{0.181}_{\pm0.005}$ & $\textbf{0.155}_{\pm0.002}$ & $\textbf{0.161}_{\pm0.007}$ & $0.407_{\pm0.001}$ & $0.197_{\pm0.001}$  \\
 & \tsmixer{}  & $0.388_{\pm0.001}$ & $0.305_{\pm0.007}$ & $0.327_{\pm0.002}$ & $0.190_{\pm0.003}$ & $0.171_{\pm0.001}$ & $0.233_{\pm0.016}$ & $0.409_{\pm0.016}$ & $\textbf{0.189}_{\pm0.003}$ \\
\midrule[\thick pt]%
 \multirow{2}{*}{192} & \model{} & $\textbf{0.409}_{\pm 0.002}$  & $\textbf{0.340}_{\pm0.002}$ & $0.353_{\pm0.006}$ & $\textbf{0.233}_{\pm0.002}$ & $\textbf{0.168}_{\pm0.001}$ & $\textbf{0.246}_{\pm0.009}$ & $\textbf{0.415}_{\pm0.005}$ & $0.235_{\pm0.000}$  \\
 & \tsmixer{} & $0.421_{\pm0.002}$ & $0.350_{\pm0.002}$ & $0.356_{\pm0.004}$ & $0.250_{\pm0.002}$ & $0.191_{\pm0.010}$ & $0.342_{\pm0.031}$ & $0.433_{\pm0.009}$ & $\textbf{0.228}_{\pm0.004}$ \\
\midrule[\thick pt]%
 \multirow{2}{*}{336} & \model{} & $\textbf{0.423}_{\pm0.001}$ & $\textbf{0.350}_{\pm0.000}$ & $\textbf{0.382}_{\pm0.001}$ & $\textbf{0.285}_{\pm0.001}$ & $\textbf{0.183}_{\pm0.000}$ & $\textbf{0.368}_{\pm0.006}$ & $\textbf{0.421}_{\pm0.001}$ & $0.276_{\pm0.001}$  \\
 & \tsmixer{}  & $0.430_{\pm0.002}$  & $0.360_{\pm0.002}$ & $0.387_{\pm0.004}$  & $0.301_{\pm0.003}$ & $0.198_{\pm0.006}$ & $0.474_{\pm0.014}$ & $0.424_{\pm0.000}$ & $\textbf{0.271}_{\pm0.001}$ \\
\midrule[\thick pt]%
 \multirow{2}{*}{720} & \model{}  & $\textbf{0.427}_{\pm0.002}$  & $\textbf{0.391}_{\pm0.001}$ & $\textbf{0.429}_{\pm0.000}$ & $\textbf{0.375}_{\pm0.001}$ & $\textbf{0.219}_{\pm0.000}$  & $1.003_{\pm0.018}$ & $\textbf{0.456}_{\pm0.003}$ & $0.334_{\pm0.000}$  \\
 & \tsmixer{}  & $0.440_{\pm0.005}$  & $0.402_{\pm0.002}$ & $0.441_{\pm0.002}$ & $0.389_{\pm0.002}$ & $0.230_{\pm0.005}$ & $1.078_{\pm0.179}$ & $0.488_{\pm0.028}$ & $\textbf{0.331}_{\pm0.001}$ \\
\bottomrule[\thick pt]%
\end{tabular}
}
\end{table}

\subsection{Computational Efficiency of \model}
\label{app:train_eff}
In this section, we showcase the computational efficiency of our approach. We compare in Table~\ref{tab:model_params} the number of parameters of \model{} and \tsmixer{} on the several benchmarks used in our experiments. We also display the ratio between the number of parameters of \tsmixer{} and the number of parameters of \model{}. Overall, \model{} has $\sim4$ times fewer parameters than \tsmixer{} while outperforming it by $14.33\%$ on average.

\begin{table}[t]
\centering
\caption{Comparison of the number of parameters between \model{} and \tsmixer{} on the datasets described in Table~\ref{tab:dataset_description} for prediction horizons $H \in \{96, 192, 336, 720\}$. We also compute the \textcolor{bluerow}{ratio} between the number of parameters of \tsmixer{} and the number of parameters of \model{}. A ratio of $10$ means that \tsmixer{} has $10$ times more parameters than \model{}. For each dataset, we display in the last cell of the corresponding row the ratio averaged over all the horizons $H$. The overall ratio over all datasets and horizons is displayed in \textbf{\textcolor{bluerow}{bold}} in the bottom right-hand cell.}
\label{tab:model_params}
\scalebox{0.8}{
\begin{tabular}{lccccccccc}
\toprule[\thick pt]
\multicolumn{1}{c}{\multirow{2}{*}{Dataset}} & \multicolumn{2}{c}{$H=96$} & \multicolumn{2}{c}{$H=192$} & \multicolumn{2}{c}{$H=336$} & \multicolumn{2}{c}{$H=720$} & \multicolumn{1}{c}{\multirow{2}{*}{\textcolor{bluerow}{\textbf{Total}}}} \\
\cmidrule(r{10pt}l{5pt}){2-3} \cmidrule(r{10pt}l{5pt}){4-5} \cmidrule(r{10pt}l{5pt}){6-7} \cmidrule(r{10pt}l{5pt}){8-9}
& \model & \texttt{TSMixer} & \model & \texttt{TSMixer}  & \model & \texttt{TSMixer} & \model & \texttt{TSMixer} &\\
\midrule[\thick pt]
ETT & 50272 & 124142 & 99520 & 173390 & 173392 & 247262 & 369904 & 444254  & - \\
Exchange & 50272 & 349344 & 99520 & 398592 & 173392 & 472464 & 369904 & 669456  & -   \\
Weather & 50272 & 121908 & 99520 & 171156 & 173392 & 245028 & 369904 & 442020  & -   \\
Electricity & 50272 & 280676 & 99520 & 329924 & 173392 & 403796 & 369904 & 600788  & -   \\
Traffic & 50272 & 793424 & 99520 & 842672 & 173392 & 916544 & 369904 & 1113536  & -   \\
\midrule[\thick pt]
\textcolor{bluerow}{\textbf{Avg. Ratio}} & \multicolumn{2}{c}{\textcolor{bluerow}{6.64}} & \multicolumn{2}{c}{\textcolor{bluerow}{3.85}} & \multicolumn{2}{c}{\textcolor{bluerow}{2.64}} & \multicolumn{2}{c}{\textcolor{bluerow}{1.77}} & \multicolumn{1}{c}{\textcolor{bluerow}{\textbf{3.73}}} \\
\bottomrule[\thick pt]
\end{tabular}}
\end{table}

\subsection{Strong Generalization Regardless of the Initialization}
\label{app:stabilized_performance}
In this section, we demonstrate that \model{} has a strong generalization capacity. In particular, \transformer{} heavily depends on the initialization, which might be due to bad local minima as its loss landscape is sharper than the one of \model{}. We display in Figure~\ref{fig:stabilized_performance_96_192} and Figure~\ref{fig:stabilized_performance_336_720} the distribution of the test MSE on $5$ runs on the datasets used in our experiments (Table~\ref{tab:dataset_description}) and various prediction horizons $H \in \{96, 192, 336, 720\}$. We can see that \model{} has strong and stable performance across the datasets and horizons, regardless of the seed. On the contrary, the performance \transformer{} is unstable with a large generalization gap depending on the seed. 

\def\figlength{0.9}
\begin{figure}[ht!]
\centering
\subfloat[Prediction horizon $H=96$.]{\includegraphics[width=\figlength\textwidth]{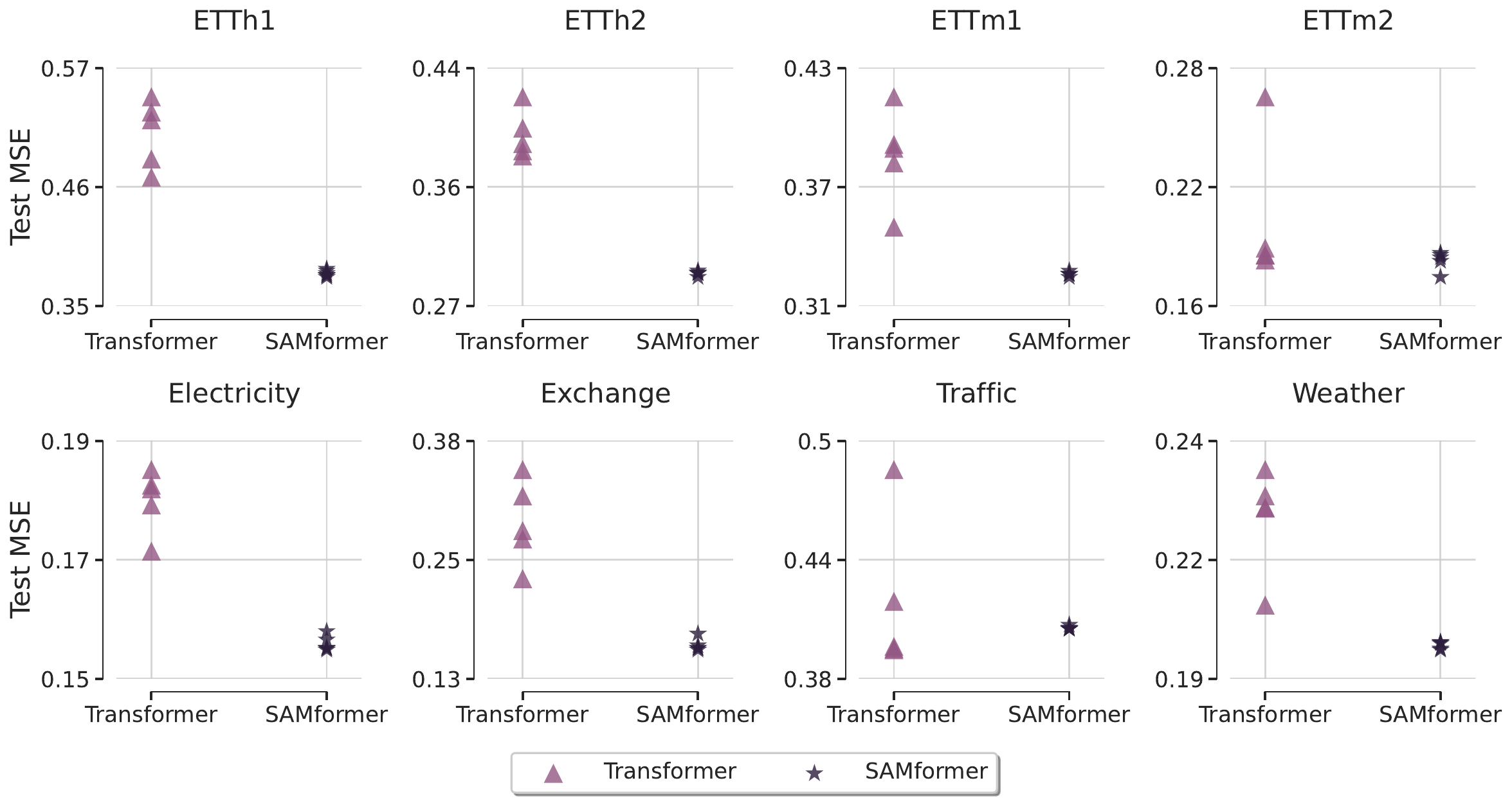}}
\label{fig:stabilized_performance_96} \qquad
\subfloat[Prediction horizon $H=192$.]{\includegraphics[width=\figlength\textwidth]{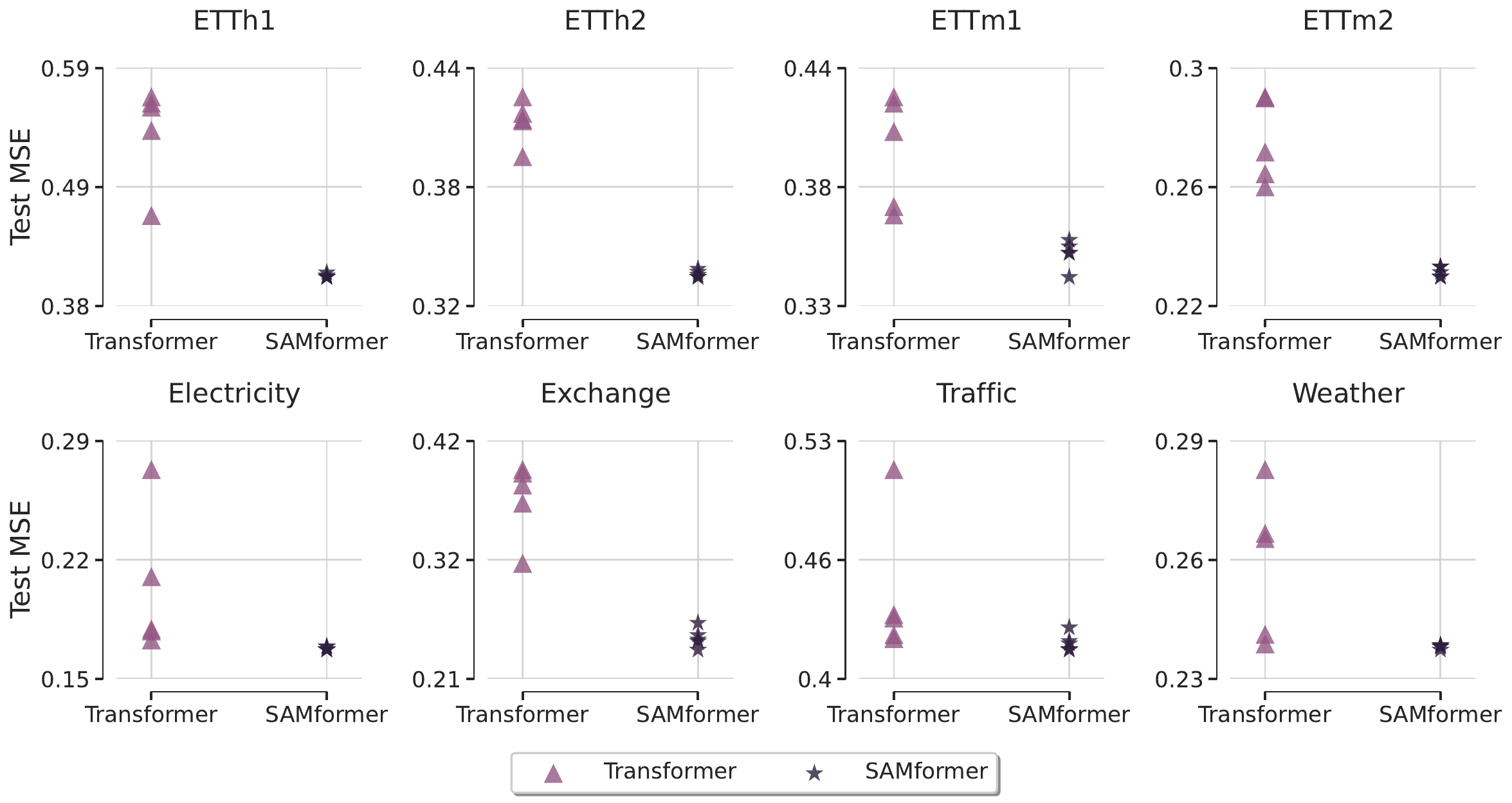}}
\label{fig:stabilized_performance_192} \qquad
\caption{Test Mean Squared error on all datasets for a prediction horizon $H \in \{96, 192\}$ across five different seed values for \transformer{} and \model{}. This plot reveals a significant variance for the \transformer{}, as opposed to the minimal variance of \model{}, showing the high impact of weight initialization on \transformer{} and the high resilience of \model{}.}
\label{fig:stabilized_performance_96_192}
\end{figure}

\begin{figure}[ht!]
\centering
\subfloat[Prediction horizon $H=336$.]{\includegraphics[width=\figlength\textwidth]{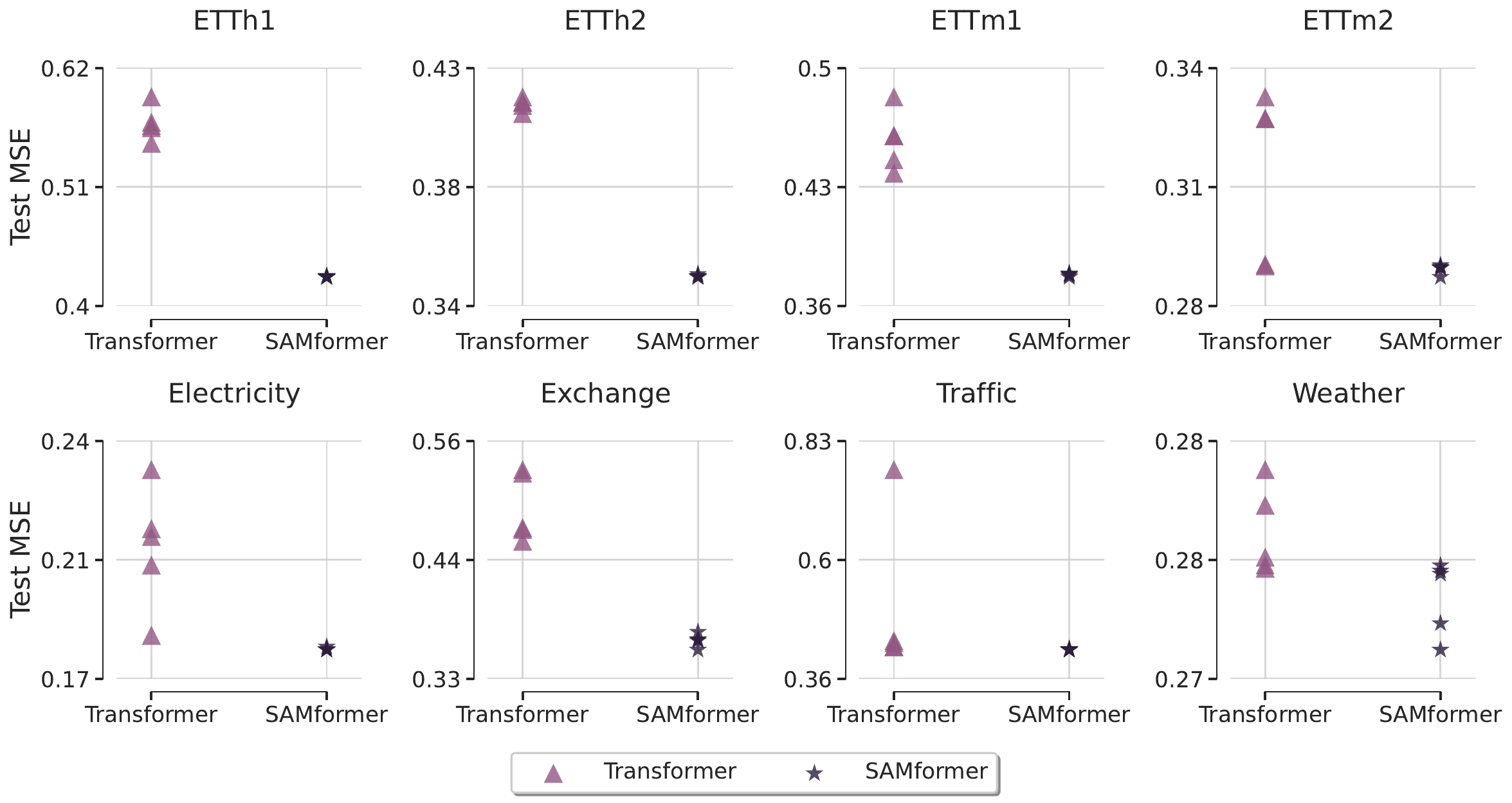}}
\label{fig:stabilized_performance_336} \qquad
\subfloat[Prediction horizon $H=720$.]{\includegraphics[width=\figlength\textwidth]{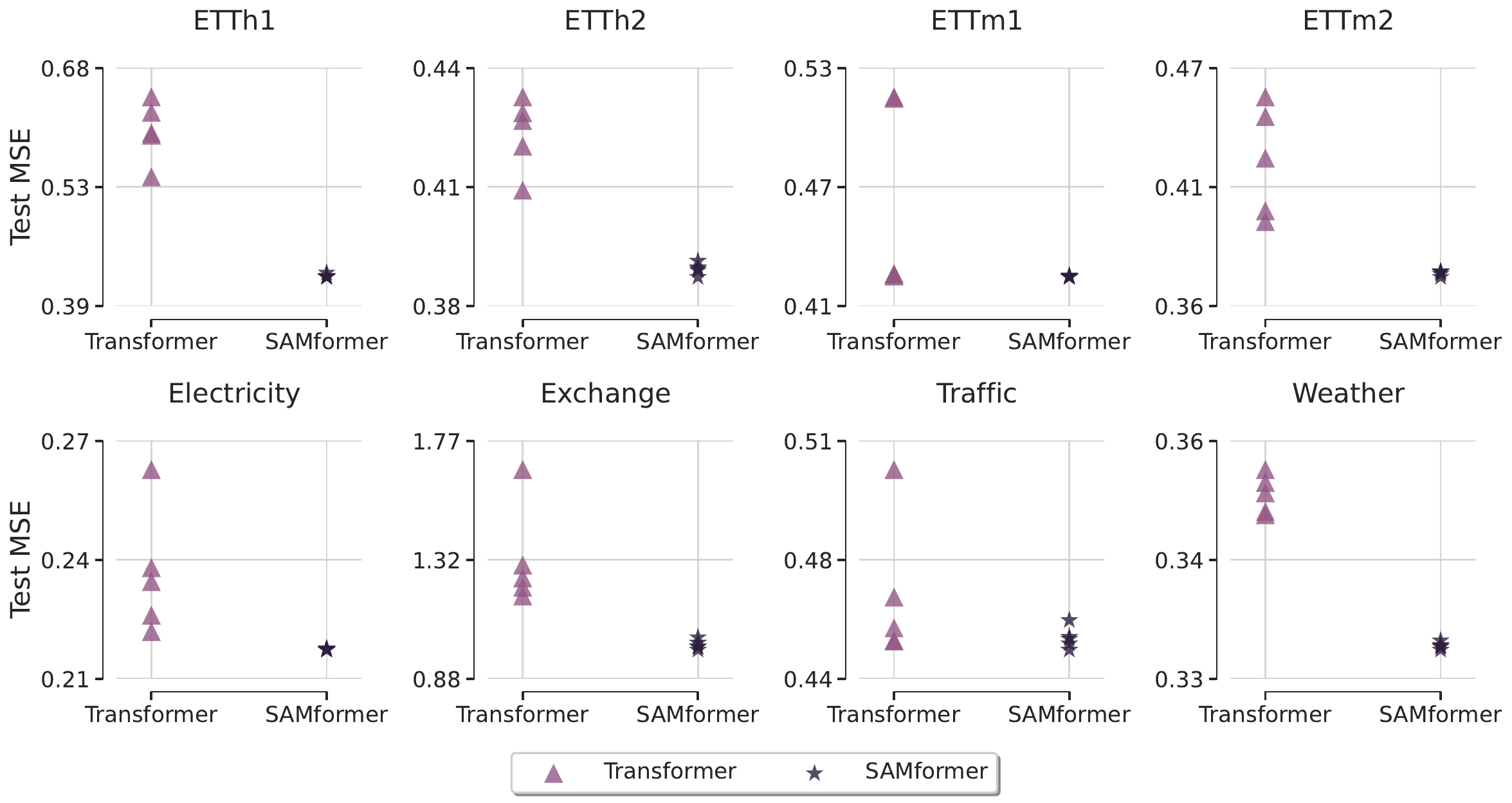}}
\label{fig:stabilized_performance_720} \qquad
\caption{Test Mean Squared error on all datasets for a prediction horizon $H \in \{336, 720\}$ across five different seed values for \transformer{} and \model{}. This plot reveals a significant variance for the \transformer{}, as opposed to the minimal variance of \model{}, showing the high impact of weight initialization on \transformer{} and the high resilience of \model{}.}
 \label{fig:stabilized_performance_336_720}
\end{figure}

\subsection{Faithful Signal Propagation}
\label{app:more_attention_matrices}
\begin{figure}[ht!]
    \centering
    \includegraphics[width=0.6\textwidth]{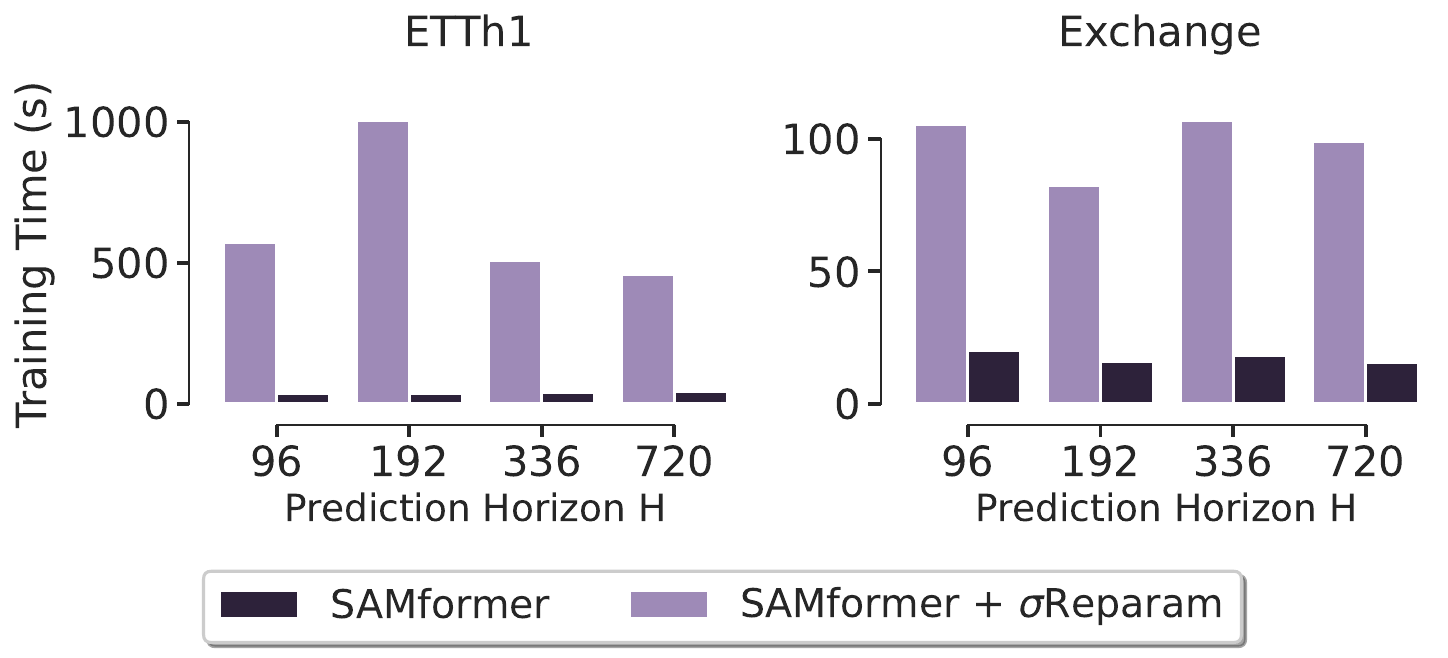}
    \caption{Using \sreparam{} on top of \model{} heavily increases the training time.}
    \label{fig:computational_time_sreparam}
\end{figure}

In this section, we consider \transformer{}, \model{}, \sreparam{}, which corresponds to \transformer{} with the rescaling proposed by~\citet{zhai2023collapse} and \model{} + \sreparam{} which is \model{} with the rescaling proposed by~\citet{zhai2023collapse}. We plot a batch of attention matrices after training with prediction horizon $H=96$ (our primary study does not identify significant changes with the value of horizon) on Weather in Figure~\ref{fig:attention_batch_weather}. While \transformer{} tends to ignore the importance of a feature on itself by having low values on the diagonal, we can see in the bottom left of Figure~\ref{fig:attention_batch_weather} that \model{} strongly encourages these feature-to-feature correlations. A very distinctive pattern is observable: a near-identity attention reminiscent of~\citet{he2023deepshortcut} and~\citet{trockman2023mimetic}. The former showed that pretrained vision models present similar patterns and both identified the benefits of such attention matrices for the propagation of information along the layers of deep transformers in NLP and computer vision. While in our setting, we have a single-layer transformer, this figure indicates that at the end of the training, self-information from features to themselves is not lost. In contrast, we see that \sreparam{} leads to almost rank-$1$ matrices with identical columns. This confirms the theoretical insights from Theorem~\ref{thm:upper_bound_nuclear_norm} that showed how rescaling the trainable weights with \sreparam{} to limit the magnitude of $\lVert \bW_Q\bW_K^\top\rVert_2$ could hamper the rank of $\bX\bW_Q\bW_K^\top\bX^\top$ and of the attention matrix. Finally, we observe that naively combining \model{} with \sreparam{} does not solve the issues: while some diagonal patterns remain, most of the information has been lost. Moreover, combining both \sreparam{} and \model{} heavily increases the training time, as shown in Figure~\ref{fig:computational_time_sreparam}.

\def\figlength{0.45}
\begin{figure}[ht!]
\centering
\subfloat[\transformer{}]{\includegraphics[width=\figlength\textwidth]{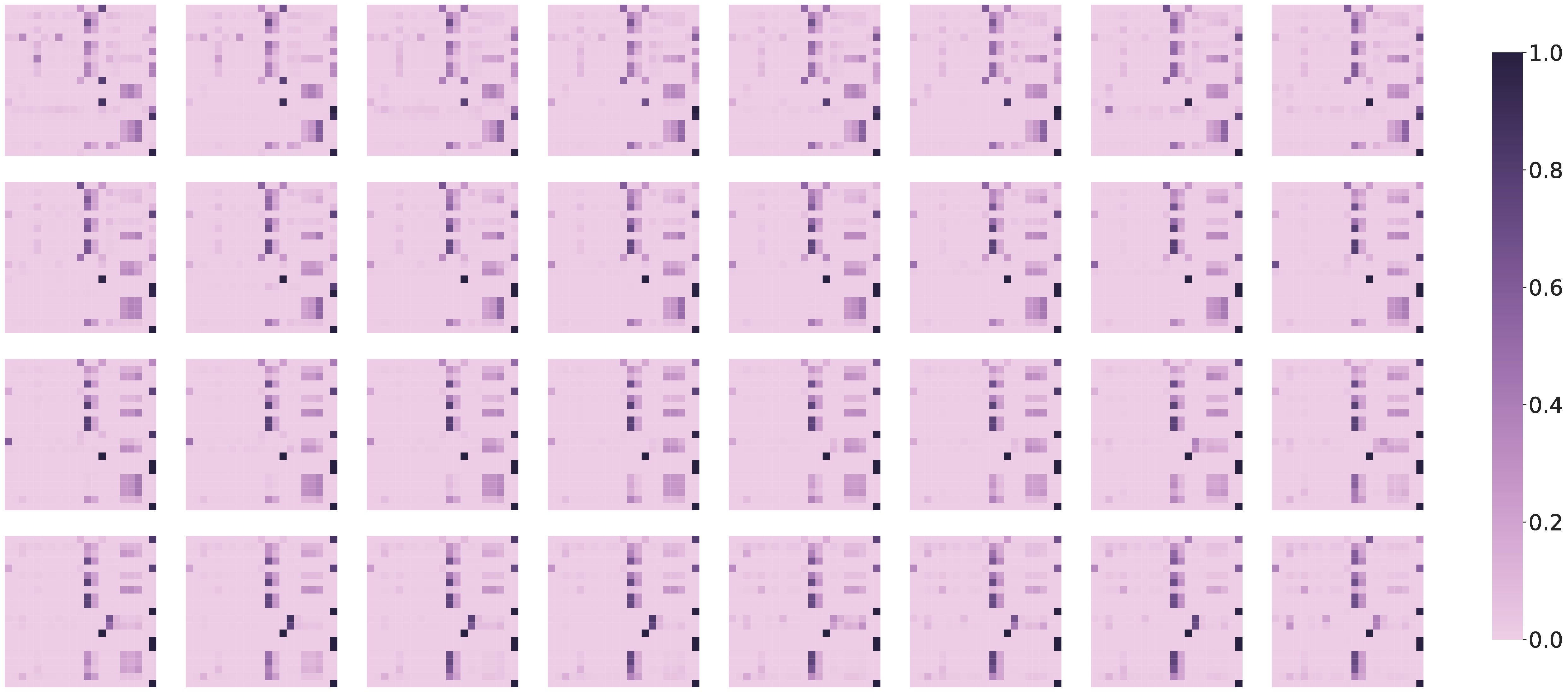}}
\label{fig:attention_batch_weather_transformer} \qquad
\subfloat[\sreparam{}]{\includegraphics[width=\figlength\textwidth]{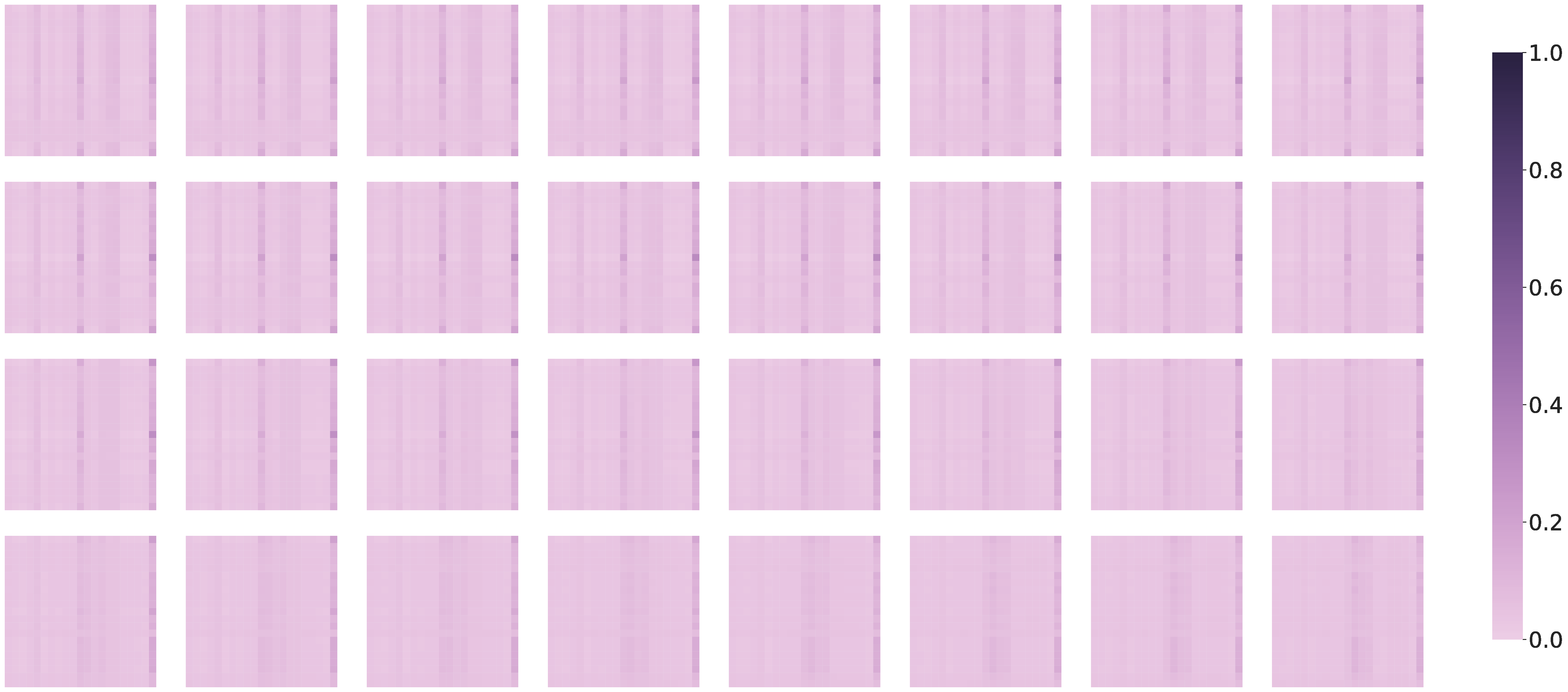}}
\label{fig:attention_batch_weather_sreparam} \qquad

\subfloat[\model{}]{\includegraphics[width=\figlength\textwidth]{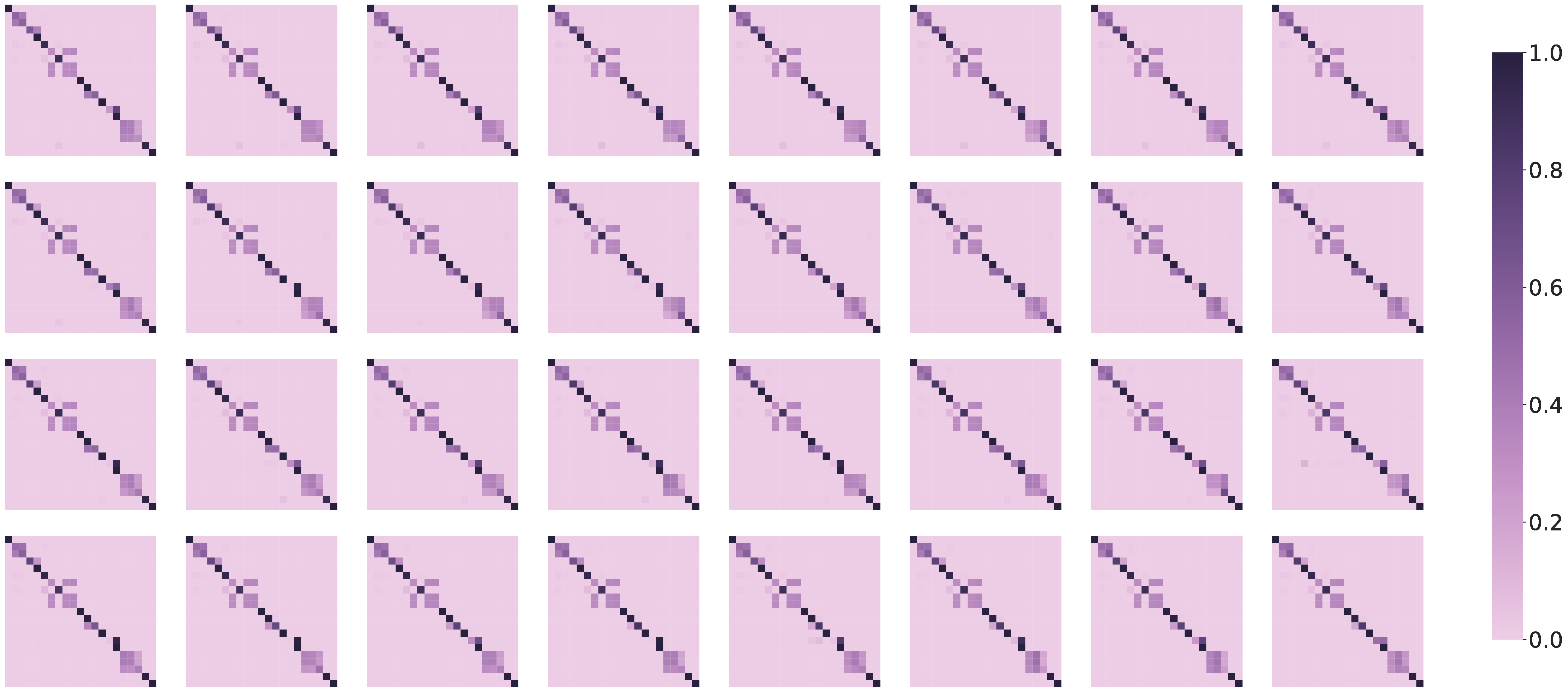}}
\label{fig:attention_batch_weather_samformer} \qquad
\subfloat[\model{} + \sreparam{}]{\includegraphics[width=\figlength\textwidth]{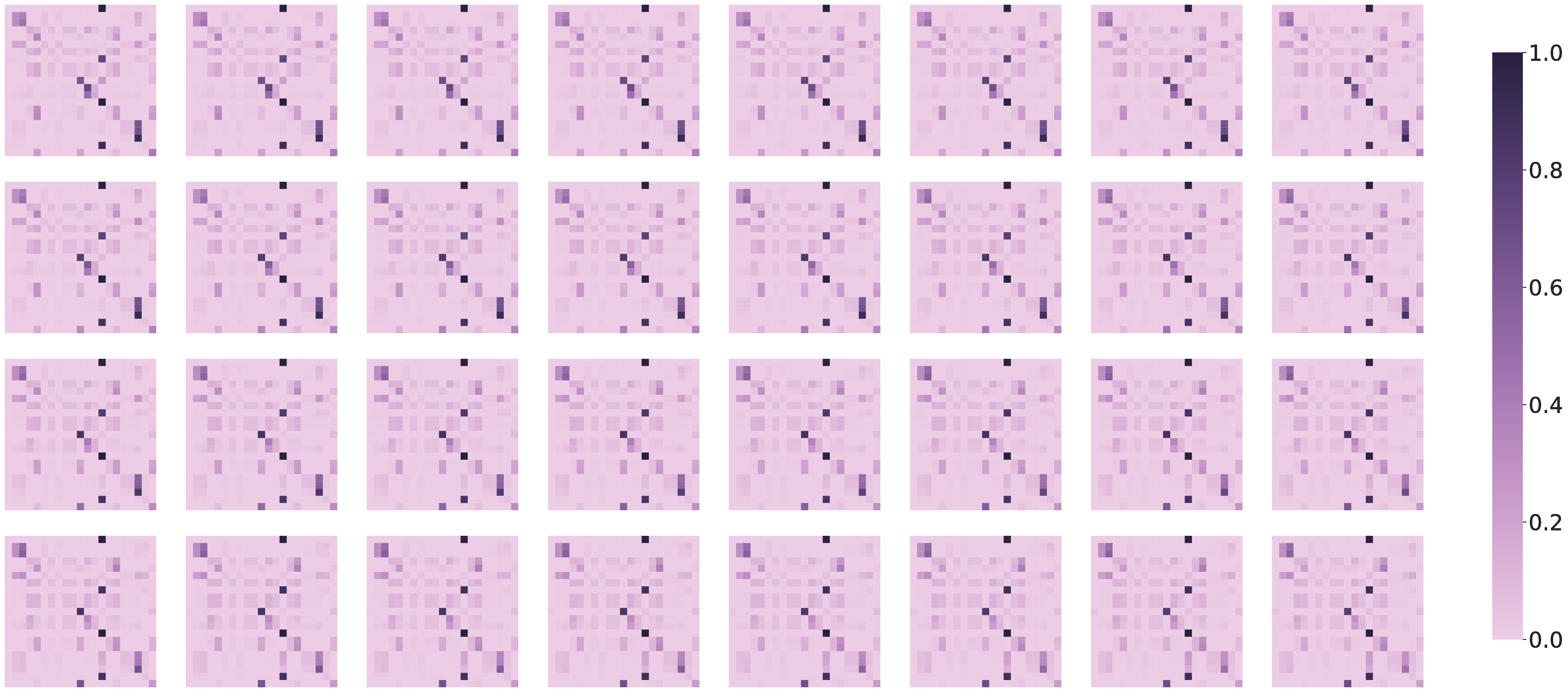}}
\label{fig:attention_batch_weather_sam_sreparan} \qquad
\caption{Batch of $32$ attention matrices on Weather with horizon $H=96$ after training different models. \textbf{(a)} \transformer{}. \textbf{(b)} \sreparam{} \textbf{(c)} \model{}. \textbf{(d)} \model{} + \sreparam{}.}
 \label{fig:attention_batch_weather}
\end{figure}

\section{Ablation Study and Sensitivity Analysis}
\label{app:ablation_sensitivity}

\subsection{Sensitivity to the Prediction Horizon $H$.}
\label{app:sensitivity_horizon}
\begin{figure}[ht!]
   \centering
   \includegraphics[width=.8\textwidth]{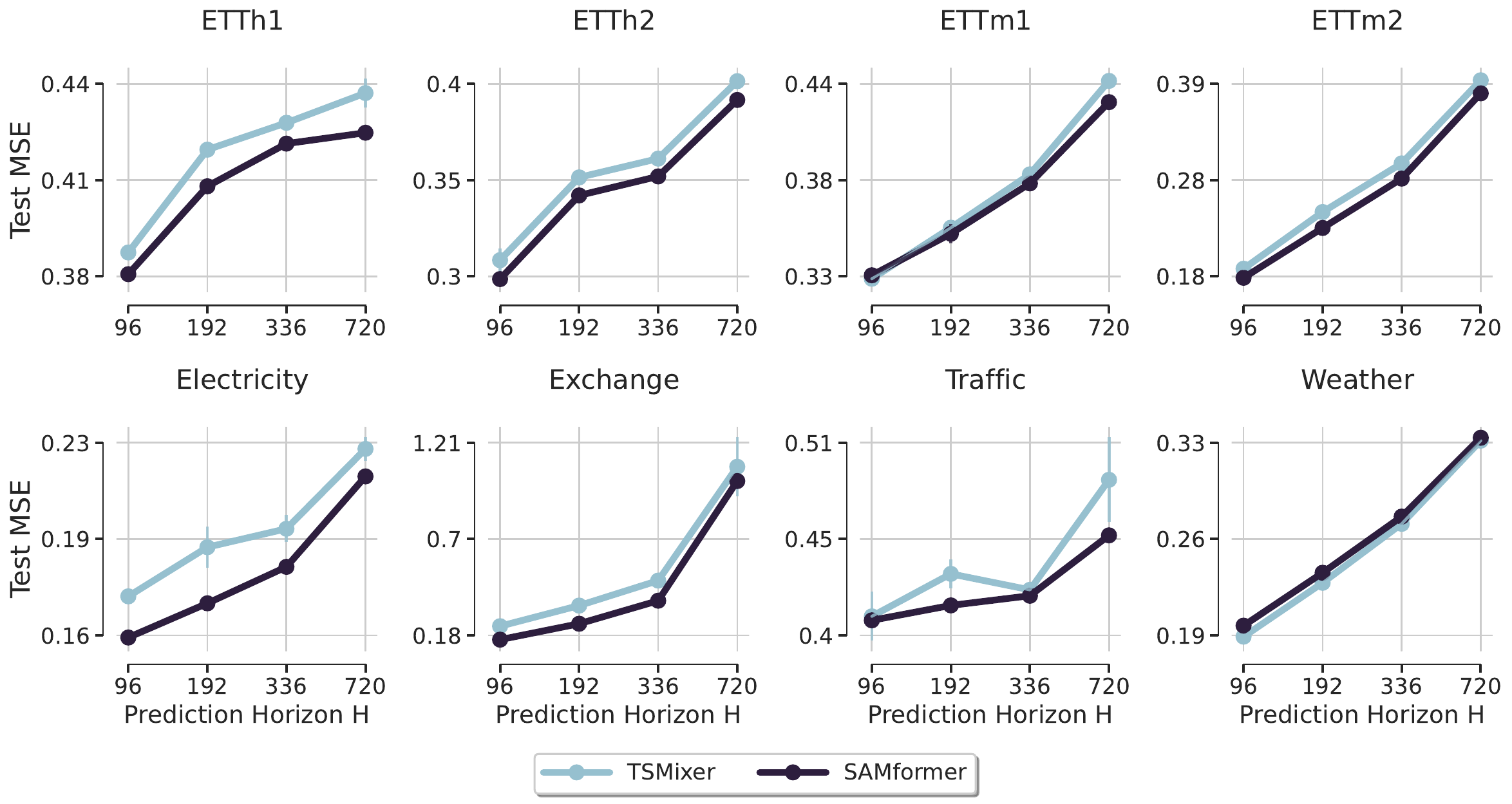}
   \caption{Evolution of the test MSE on all datasets for a prediction horizon $H \in \{96, 192, 336, 720\}$. We display the average test MSE with a $95\%$ confidence interval. We see that \model{} consistently performs well with a low variance. Despite its lightweight (Table~\ref{tab:model_params}), \model{} surpasses \tsmixer{} (trained with SAM) on $7$ out of $8$ datasets as shown in Table~\ref{tab:all_results} and Table~\ref{tab:significance_test}.}
   \label{fig:sensitivity_horizon}
\end{figure}

In Figure~\ref{fig:sensitivity_horizon}, we show that \model{} outperforms its best competitor, \tsmixer{} trained with SAM, on $7$ out of $8$ datasets for all values of prediction horizon $H$. This demonstrates the robustness of \model{}.

\subsection{Sensitivity to the Neighborhood Size $\rho$.}
\label{app:sensitivity_rho}
\begin{figure}[ht!]
   \centering
   \includegraphics[width=\textwidth]{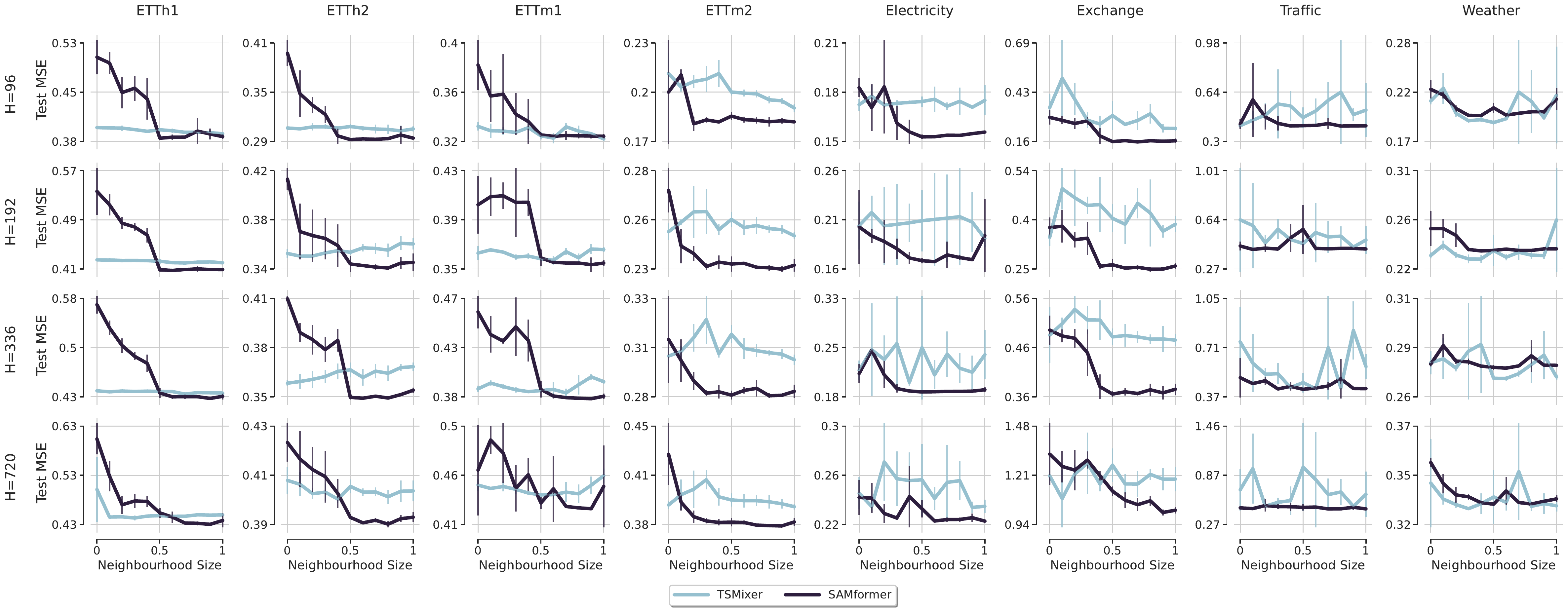}
   \caption{Evolution of the test MSE with the neighborhood size $\rho$ of SAM (Remark~\ref{rmk:sam_rho}). We display the average test MSE with a $95\%$ confidence interval. Overall, \model{} has a smooth behavior with $\rho$, with a decreasing MSE and less variance. On the contrary, \tsmixer{} is less stable and fluctuates more. On most of the datasets, the range of neighborhood seizes $\rho$ such that \model{} is below \tsmixer{} is large. The first value $\rho=0$ amounts to the usual minimization with Adam, which confirms that SAM always improves the performance of \model{}. In addition, and despite its lightweight (Table~\ref{tab:model_params}), \model{} achieves the lowest MSE on $7$ out of $8$ datasets, as shown in Table~\ref{tab:all_results} and Table~\ref{tab:significance_test}. It should be noted that compared to similar studies in computer vision~\citep{chen2022vitwithsam}, values of $\rho$ must be higher to effectively improve the generalization and flatten the loss landscapes.}
   \label{fig:sensitivity_rho}
\end{figure}

In Figure~\ref{fig:sensitivity_rho}, we display the evolution of test MSE of \model{} and \tsmixer{} with the values of neighborhood size $\rho$ for SAM. Overall, \model{} has a smooth behavior with $\rho$, with a decreasing MSE and less variance. On the contrary, \tsmixer{} is less stable and fluctuates more. On most of the datasets, the range of neighborhood seizes $\rho$ such that \model{} is below \tsmixer{} is large. The first value $\rho=0$ amounts to the usual minimization with Adam, which confirms that SAM always improves the performance of \model{}. In addition, and despite its lightweight (Table~\ref{tab:model_params}), \model{} achieves the lowest MSE on $7$ out of $8$ datasets, as shown in Table~\ref{tab:all_results} and Table~\ref{tab:significance_test}. It should be noted that compared to similar studies in computer vision~\citep{chen2022vitwithsam}, values of $\rho$ must be higher to effectively improve the generalization and flatten the loss landscapes. This follows from the high sharpness $\lambda_{max}$ observed in time series forecasting (Figure~\ref{fig:sharpness_2_datasets}) compared to computer vision models~\citep{chen2022vitwithsam}.

\subsection{Sensitivity to the Change of the Optimizer.}
\label{app:sensitivity_optim}
\begin{figure*}[ht!]
\centering
\begin{subfigure}{.49\textwidth}
  \centering
  \includegraphics[width=\linewidth]{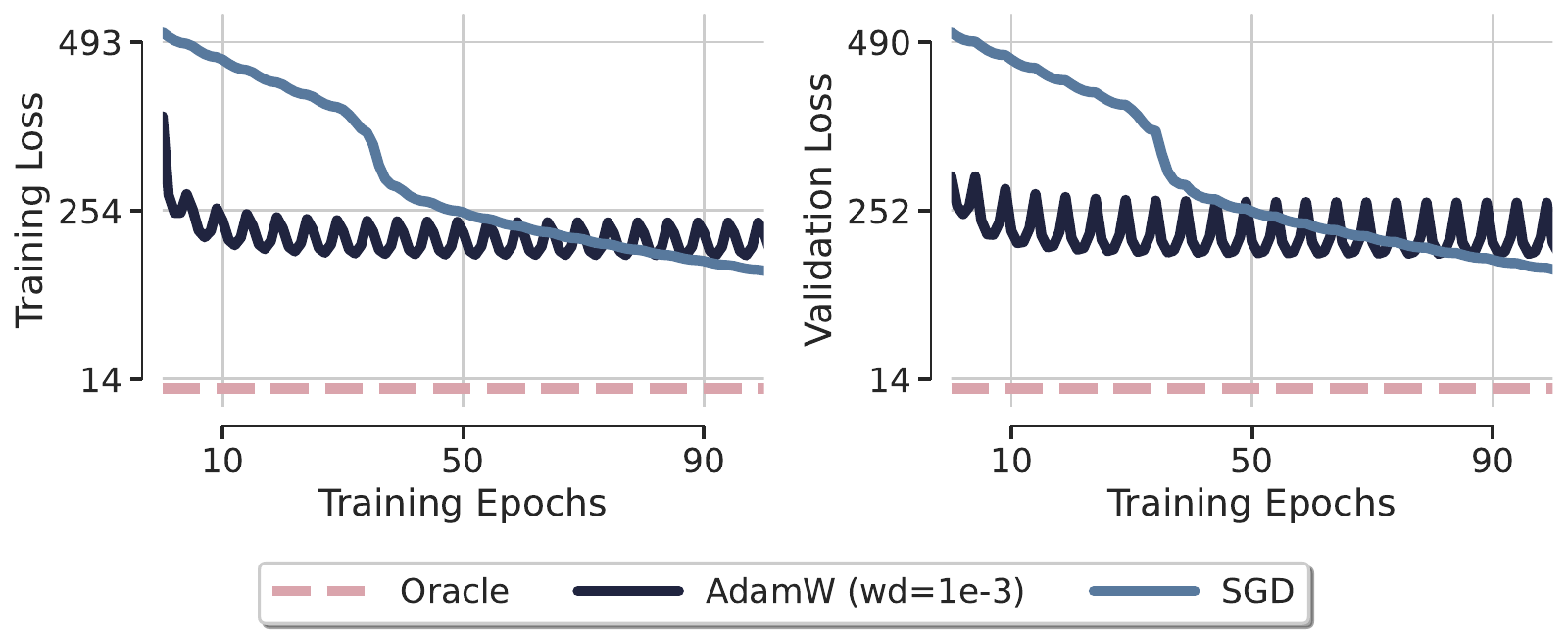}
  \caption{SGD and AdamW with $\mathrm{wd}=1\mathrm{e}{-3}$}
  \label{fig:bad_adamw_sgd}
\end{subfigure}%
\begin{subfigure}{.49\textwidth}
  \centering
  \includegraphics[width=\linewidth]{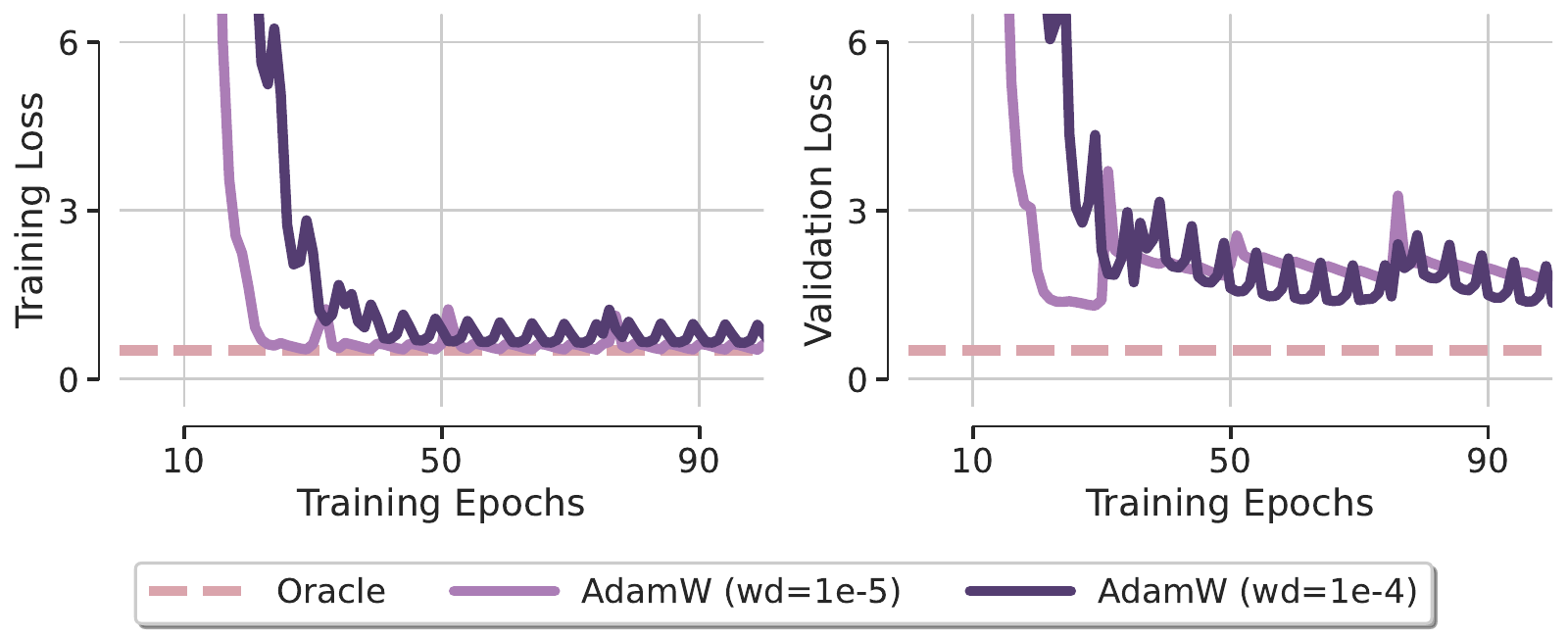}
  \caption{AdamW with $\mathrm{wd} \in \{1\mathrm{e}{-5}, 1\mathrm{e}{-4}\}$.}
\label{fig:good_adamw}
\end{subfigure}
\caption{Illustration of different optimizers on synthetic data generated with Eq.~\eqref{eq:toy_exp} where \texttt{Oracle} is the least-square solution. We saw in Figure~\ref{fig:toy_exp_with_SAM} that with Adam, \transformer{} overfits and has poor performance while \model{} smoothly reaches the oracle. \textbf{(a)} We can see that using SGD and Adam with weight decay $\mathrm{wd}=1\mathrm{e}{-5}$ leads to huge loss magnitudes and fails to converge. \textbf{(b)} With well-chosen weight decays ($\mathrm{wd} \in \{1\mathrm{e}{-3}, 1\mathrm{e}{-4}\}$), training \transformer{} with AdamW leads to similar performance than Adam. The overfitting is noticeable and the training is unstable. AdamW does not bring more stabilization and is very sensitive to the hyperparameters. Hence, this toy example motivates us to conduct our thorough experiments with the optimizer Adam.}
\label{fig:sensitivity_optim}
\end{figure*}

In our work, we considered the Adam optimizer \citep{KingBa15} as it is the de-facto optimizer for transformer-based models \citep{ahn2023linear, pan2022toward, zhou2022fedformer, haoyi2021informer, chen2022vitwithsam}. The superiority of Adam to optimize networks with attention has been empirically and theoretically studied, where recent works show that the SGD \citep{nesterov1983sgd} was not suitable for attention-based models \citep{ahn2023linear, liu2020understanding, pan2022toward, zhang2020adaptattention}. To ensure the thoroughness of our investigation, we conducted experiments on the synthetic dataset introduced in Eq.~\eqref{eq:toy_exp} and reported the results in Figure~\ref{fig:bad_adamw_sgd}. As expected, we see that using SGD leads to high-magnitude losses and divergence. We also conducted the same experiments with the AdamW \citep{loshchilov2018decoupled} that incorporates the weight decay scheme in the adaptive optimizer Adam \citep{KingBa15}. We display the results obtained with weight decay factors $\mathrm{wd}=1\mathrm{e}{-3}$ in Figure~\ref{fig:bad_adamw_sgd} and with $\mathrm{wd} \in \{1\mathrm{e}{-5}, 1\mathrm{e}{-4}\}$ in Figure~\ref{fig:good_adamw}. When $\mathrm{wd}=1\mathrm{e}{-3}$, we observe that it does not converge.
However, with $\mathrm{wd} \in \{1\mathrm{e}{-5}, 1\mathrm{e}{-4}\}$, we observe a similar behavior for \transformer{} than when it is trained with Adam (Figure~\ref{fig:toy_exp_without_SAM}). Hence, using AdamW does not lead to the significant benefits brought by SAM (Figure~\ref{fig:toy_exp_with_SAM}. As the optimization is very sensitive to the value of weight decay $\mathrm{wd}$, it motivates us to conduct our experiments with Adam.

\subsection{Ablation on the Implementation.}
\label{app:choice_implementation}

This ablation study contrasts two variants of our model to showcase the effectiveness of Sharpness-Aware Minimization (SAM) and our attention approach. \texttt{Identity Attention} represents \model{} with an attention weight matrix constrained to identity, illustrating that SAM does not simply reduce the attention weight matrix to identity, as performance surpasses this configuration. \texttt{Temporal Attention} is compared to our Transformer without SAM, highlighting our focus on treating feature correlations in the attention mechanism rather than temporal correlations.

\begin{table}[th!]
\centering
\caption{The \texttt{Temporal Attention} model is benchmarked against our \transformer{} model, which employs feature-based attention rather than time-step-based attention. We report in the last column the \textbf{Overall improvement} in MSE and MAE of \transformer{} over the \texttt{Temporal Attention}. This comparison reveals that channel-wise attention, i.e., focusing on features pairwise correlations, significantly boosts the performance, with a $12.97\%$ improvement in MSE and $18.09\%$ in MAE across all considered datasets. }
\label{tab:temporal_attention}
\setlength{\tabcolsep}{5pt} 
\scalebox{0.7}{
\begin{tabular}{cccccccccccc}
\toprule[\thick pt]%
\multicolumn{1}{c}{Model} & \multicolumn{1}{c}{Metrics} & \multicolumn{1}{c}{H} & \multicolumn{1}{c}{ETTh1} & \multicolumn{1}{c}{ETTh2} & \multicolumn{1}{c}{ETTm1} & \multicolumn{1}{c}{ETTm2} & \multicolumn{1}{c}{Electricity} & \multicolumn{1}{c}{Exchange} & \multicolumn{1}{c}{Traffic} & \multicolumn{1}{c}{Weather} & \multicolumn{1}{c}{\textbf{Overall Improvement}} \\ 
\midrule[\thick pt]
\multirow{8}{*}{\rotatebox[origin=c]{90}{\scriptsize \texttt{Temporal Attention}}} & \multirow{4}{*}{MSE} & 96 & $0.496_{\pm0.009}$ &$0.401_{\pm0.011}$  & $0.542_{\pm0.063}$  & $0.330_{\pm0.034}$ & $0.291_{\pm0.025}$ & $0.684_{\pm0.218}$ & $0.933_{\pm0.188}$ & $0.225_{\pm0.005}$ & \multirow{4}{*}{12.97\%} \\
 &  & 192 & $0.510_{\pm0.014}$  & $0.414_{\pm0.020}$  &$0.615_{\pm0.056}$  &$0.394_{\pm0.033}$  &$0.294_{\pm0.024}$  & $0.434_{\pm0.063}$ &$0.647_{\pm0.131}$  & $0.254_{\pm0.001}$ &  \\
 &  & 336 & $0.549_{\pm0.017}$ &$0.396_{\pm0.014}$  &$0.620_{\pm0.046}$  & $0.436_{\pm0.081}$ &$0.290_{\pm0.016}$  &$0.473_{\pm0.014}$  & $0.656_{\pm0.113}$ & $0.292_{\pm0.000}$ &  \\
 &  & 720 &$0.604_{\pm0.017}$   & $0.396_{\pm0.010}$  & $0.694_{\pm0.055}$ & $0.469_{\pm0.005}$ &$0.307_{\pm0.014}$  & $1.097_{\pm0.084}$  & - & $0.346_{\pm0.000}$ &  \\
\cmidrule{2-12}
 & \multirow{4}{*}{MAE}  & 96 &$0.488_{\pm0.007}$  & $0.434_{\pm0.006}$ & $0.525_{\pm0.040}$ & $0.393_{\pm0.020}$ &$0.386_{\pm0.014}$  & $0.589_{\pm0.096}$ & $0.598_{\pm0.072}$ &  $0.277_{\pm0.004}$&\multirow{4}{*}{18.09\%}  \\
 &  & 192 &$0.492_{\pm0.010}$  & $0.443_{\pm0.015}$  & $0.566_{\pm0.032}$  & $0.421_{\pm0.019}$  & $0.385_{\pm0.014}$ &$0.498_{\pm0.033}$  &$0.467_{\pm0.072}$  & $0.294_{\pm0.001}$ &  \\
 &  & 336 &$0.517_{\pm0.012}$  &$0.440_{\pm0.012}$  &$0.550_{\pm0.024}$  & $0.443_{\pm0.039}$ &$0.383_{\pm0.009}$  &$0.517_{\pm0.008}$  &$0.469_{\pm0.070}$   & $0.320_{\pm0.000}$ &  \\
 &  & 720 &$0.556_{\pm0.009}$  &$0.442_{\pm0.006}$  & $0.584_{\pm0.027}$ & $0.459_{\pm0.004}$ &$0.396_{\pm0.012}$  &$0.782_{\pm0.041}$  & -  & $0.356_{\pm0.000}$ &  \\
\bottomrule[\thick pt]%
\end{tabular}
}
\end{table}

\begin{table}[ht!]
\centering
\caption{\texttt{Identity Attention} represents our \model{} with the attention weight matrix constrained to an identity matrix. We report in the last column the \textcolor{bluerow}{\textbf{Overall improvement}} in MSE and MAE of \model{} over the \texttt{Identity Attention}. This setup demonstrates that naively fixing the attention matrix to the identity does not enable to match the performance of SAM, despite the near-identity attention matrices SAM showcases (see Appendix~\ref{app:more_attention_matrices} for more details). In particular, we observe an overall improvement of $11.93\%$ in MSE and $4.18\%$ in MAE across all the datasets.}
\label{tab:identity_attention}
\setlength{\tabcolsep}{5pt} 
\scalebox{0.7}{
\begin{tabular}{cccccccccccc}
\toprule[\thick pt]%
\multicolumn{1}{c}{Model} & \multicolumn{1}{c}{Metrics} & \multicolumn{1}{c}{H} & \multicolumn{1}{c}{ETTh1} & \multicolumn{1}{c}{ETTh2} & \multicolumn{1}{c}{ETTm1} & \multicolumn{1}{c}{ETTm2} & \multicolumn{1}{c}{Electricity} & \multicolumn{1}{c}{Exchange} & \multicolumn{1}{c}{Traffic} & \multicolumn{1}{c}{Weather} & \multicolumn{1}{c}{\textcolor{bluerow}{\textbf{Overall Improvement}}} \\ 
\midrule[\thick pt]
\multirow{8}{*}{\rotatebox[origin=c]{90}{\footnotesize \texttt{Identity Attention}}} & \multirow{4}{*}{MSE} & 96 & $0.477_{\pm0.059}$ & $0.346_{\pm0.055}$  & $0.345_{\pm0.027}$ & $0.201_{\pm0.035}$ & $0.175_{\pm0.015}$  & $0.179_{\pm0.031}$ & $0.416_{\pm0.037}$ & $0.206_{\pm0.019}$ & \multirow{4}{*}{11.93\%}  \\
 &  & 192 &  $0.467_{\pm0.074}$ & $0.374_{\pm0.031}$ & $0.384_{\pm0.042}$ & $0.248_{\pm0.016}$ & $0.189_{\pm0.022}$ & $0.320_{\pm0.070}$  & $0.437_{\pm0.041}$  & $0.236_{\pm0.002}$ &  \\
 &  & 336 & $0.512_{\pm0.070}$ & $0.372_{\pm0.024}$  & $0.408_{\pm0.032}$  & $0.303_{\pm0.022}$ & $0.211_{\pm0.019}$  & $0.443_{\pm0.071}$  & $0.500_{\pm0.155}$ & $0.277_{\pm0.003}$  &  \\
 &  & 720 & $0.505_{\pm0.107}$  & $0.405_{\pm0.012}$ & $0.466_{\pm0.043}$ & $0.397_{\pm0.029}$ & $0.233_{\pm0.019}$  & $1.123_{\pm0.076}$ &$0.468_{\pm0.021}$  & $0.338_{\pm0.009}$ &  \\
\cmidrule{2-12}
 & \multirow{4}{*}{MAE}  & 96 & $0.473_{\pm0.041}$  & $0.395_{\pm0.033}$ & $0.376_{\pm0.019}$  & $0.294_{\pm0.027}$  & $0.283_{\pm0.023}$  & $0.320_{\pm0.023}$ & $0.301_{\pm0.039}$ & $0.259_{\pm0.021}$ & \multirow{4}{*}{4.18\%}  \\
 &  & 192 & $0.463_{\pm0.055}$ & $0.413_{\pm0.022}$   &$0.399_{\pm0.030}$& $0.321_{\pm0.012}$ & $0.291_{\pm0.029}$ & $0.418_{\pm0.043}$ &$0.314_{\pm0.042}$  & $0.278_{\pm0.002}$ &  \\
 &  & 336 & $0.490_{\pm0.049}$  & $0.413_{\pm0.015}$  & $0.411_{\pm0.019}$ & $0.354_{\pm0.018}$ & $0.309_{\pm0.021}$  & $0.498_{\pm0.041}$ & $0.350_{\pm0.106}$  & $0.305_{\pm0.003}$ &  \\
 &  & 720 & $0.496_{\pm0.066}$ & $0.438_{\pm0.008}$  & $0.444_{\pm0.030}$ & $0.406_{\pm0.017}$ &$0.322_{\pm0.021}$  & $0.788_{\pm0.021}$ & $0.325_{\pm0.023}$ & $0.347_{\pm0.009}$ &  \\
\bottomrule[\thick pt]%
\end{tabular}
}
\end{table}

\section{Additional Background}
\label{app:additional_background}
\subsection{Reversible Instance Normalization: \texttt{RevIN}} 
\label{app:revin}
\paragraph{Overview.}
\citet{kim2021reversible} recently proposed \texttt{RevIN}, a reversible instance normalization to reduce the discrepancy between the distributions of training and test data. Indeed, statistical properties of real-world time series, e.g. mean and variance, can change over time, leading to non-stationary sequences. This causes a distribution shift between training and test sets for the forecasting task. The \texttt{RevIN} normalization scheme is now widespread in deep learning approaches for time series forecasting~\citep{chen2023tsmixer, nie2023patchtst}. The \texttt{RevIN} normalization involves trainable parameters $(\bm{\beta}, \bm{\gamma}) \in \mathbb{R}^K \times \mathbb{R}^K$ and consists of two parts: a normalization step and a symmetric denormalization step. Before presenting them, we introduce for a given input time series $\mbf{X}^{(i)} \in \mathcal{X}$ the empirical mean $\hat{\mu}[\mbf{X}^{(i)}_{k}]$ and empirical standard deviation $\hat{\sigma}^2[\mbf{X}^{(i)}_{k}]$ of its $k$-th feature $\mbf{X}^{(i)}_{k} \in \mathbb{R}^{1 \times L}$ as follows:
\begin{equation}
\label{eq:empirical_mean_std}
\begin{cases}
    &\hat{\mu}\mleft[\mbf{X}^{(i)}_{k}\mright] = \frac{1}{L}\sum_{t=1}^L\mbf{X}^{(i)}_{kj} \\
    &\hat{\sigma}^2\mleft[\mbf{X}^{(i)}_{k}\mright] = \frac{1}{L}\sum_{t=1}^L(\mbf{X}^{(i)}_{kj} - \hat{\mu}[\mbf{X}^{(i)}_{k}])^2\,.
\end{cases}
\end{equation}
The first one acts on the input sequence $\mbf{X}^{(i)}$ and outputs the corresponding normalized sequence $\mbf{\tilde{X}}^{(i)} \in \mathbb{R}^{K \times L}$ such that for all $k, t$, 
\begin{equation}
\label{eq:revin_first_step}
    \mbf{\tilde{X}}^{(i)}_{kt} = \bm{\gamma}_k\mleft( \frac{\mbf{X}^{(i)}_{kt} - \hat{\mu}\mleft[\mbf{X}^{(i)}_{k}\mright]}{\sqrt{\hat{\sigma}^2\mleft[\mbf{X}^{(i)}_{k}\mright] + \varepsilon}}\mright) + \bm{\beta}_k,
\end{equation}
where $\varepsilon > 0$ is a small constant to avoid dividing by $0$. The neural network's input is then $\mbf{\tilde{X}}^{(i)}$, instead of $\mbf{X}^{(i)}$. The second step is applied to the output of the neural network $\mbf{\tilde{Y}}^{(i)}$, such that the final output considered for the forecasting is the denormalized sequence $\mbf{\hat{Y}}^{(i)} \in \mathbb{R}^{K \times H}$ such that for all $k, t$,
\begin{equation}
\label{eq:revin_second_step}
    \mbf{\hat{Y}}^{(i)}_{kt} = \sqrt{\hat{\sigma}^2\mleft[\mbf{X}^{(i)}_{k}\mright] + \varepsilon} \cdot \mleft(\frac{\mbf{\tilde{Y}}^{(i)}_{kt} - \bm{\beta}_k}{\bm{\gamma}_k}\mright) + \hat{\mu}\mleft[\mbf{X}^{(i)}_{k}\mright].
\end{equation}
As stated in~\citet{kim2021reversible}, $\hat{\mu}, \hat{\sigma}^2, \bm{\beta}$ and $\bm{\gamma}$ contain the non-stationary information of the input sequences $\mbf{X}^{(i)}$.

\paragraph{End-to-end closed form with linear model and \texttt{RevIN}.} We consider a simple linear neural network. Formally, for any input sequence $\mbf{X} \in \mathbb{R}^{D \times L}$, the prediction of $f_\mathrm{lin} \colon \mathbb{R}^{D \times L} \to \mathbb{R}^{D \times H}$ simply writes
\begin{equation}
    \label{eq:linear_model}
    f_\mathrm{lin}(\mbf{X}) = \mbf{X}\mbf{W}.
\end{equation}
When combined with \texttt{RevIN}, the neural network $f_\mathrm{lin}$ is not directly applied to the input sequence but after the first normalization step of \texttt{RevIN} (Eq.~\eqref{eq:revin_first_step}). An interesting benefit of the simplicity of $f_\mathrm{lin}$ is that it enables us to write its prediction in closed form, even when with \texttt{RevIN}. The proof is deferred to Appendix~\ref{app:closed_form}.
\begin{boxprop}[Closed-form formulation]
\label{prop:closed_form}
    For any input sequence $\mbf{X} \in \mathbb{R}^{K \times L}$, the output of the linear model $\mbf{\hat{Y}} = f_\mathrm{lin}(\mbf{X}) \in \mathbb{R}^{K \times H}$ has entries
\begin{equation}
\label{eq:closed_form}
    \mbf{\hat{Y}}_{kt} = \hat{\mu}\mleft[\mbf{X}_{k}\mright] + \sum_{j=1}^L \mleft(\mbf{X}_{kj} - \hat{\mu}\mleft[\mbf{X}_{k}\mright]\mright)\mbf{W}_{jt} - \frac{\bm{\beta}_k}{\bm{\gamma}_k} \sqrt{\hat{\sigma}^2\mleft[\mbf{X}_{k}\mright] + \varepsilon} \mleft(1 - \sum_{j=1}^L\mbf{W}_{jt}\mright), 
\end{equation}
\end{boxprop}
Proposition~\ref{prop:closed_form} highlights the fact that the $k$-th variable of the outputs $\mbf{\hat{Y}}$ only depends on $k$-th variable of the input sequence $\mbf{X}$. It leads to channel-independent forecasting, although we did not explicitly enforce it. \eqref{eq:closed_form} can be seen as a linear interpolation around the mean $\hat{\mu}$ with a regularization term on the network parameters $\mbf{W}$ involving the non-stationary information $\hat{\sigma}^2, \bm{\beta}, \bm{\gamma}$. Moreover, the output sequence $\mbf{\hat{Y}}$ can be written in a more compact and convenient matrix formulation as follows
\begin{equation}
\label{eq:matrix_formulation}
    \mbf{\hat{Y}} = \mbf{X}\mbf{W} + \bm{\xi}^{(\mbf{X}, \mbf{W}, \bm{\beta}, \bm{\gamma})},
\end{equation}
where $\bm{\xi}^{(\mbf{X}, \mbf{W}, \bm{\beta}, \bm{\gamma})} \in \mathbb{R}^{K \times H}$ with entry $\mleft( \hat{\mu}\mleft[\mbf{X}_{k}\mright] - \frac{\bm{\beta}_k}{\bm{\gamma}_k} \sqrt{\hat{\sigma}^2\mleft[\mbf{X}_{k}\mright] + \varepsilon} \mright) \mleft(1 - \sum_{j=1}^L\mbf{W}_{jt}\mright)$ in the $k$-th row and $t$-th column. The proof is deferred to Appendix~\ref{app:matrix_formulation}. With this formulation, the predicted sequence can be seen as a sum of a linear term $ \mbf{X}\mbf{W}$ and a residual term $\bm{\xi}^{(\mbf{X}, \mbf{W}, \bm{\beta}, \bm{\gamma})}$ that takes into account the first and second moments of each variable $\mbf{X}_k$, which is reminiscent of the linear regression model. 

\subsection{Sharpness-aware minimization (SAM)}
\label{app:sam}
\paragraph{Regularizing with the sharpness.} 
Standard approaches consider a parametric family of models $f_{\bm{\omega}}$ and aim to find parameters $\bm{\omega}$ that minimize a training objective $\mathcal{L}_\mathrm{train}\mleft(\bm{\omega}\mright)$, used as a tractable proxy to the true generalization error $\mathcal{L}_\mathrm{test}\mleft(\bm{\omega}\mright)$. Most deep learning pipelines rely on first-order optimizers, e.g. SGD~\citep{nesterov1983sgd} or Adam~\citep{KingBa15}, that disregard higher-order information such as the curvature, despite its connection to generalization~\citep{dziugaite2017nonvacuous, chaudhari2017entropysgd, keskar2017sharpminima}. As $\mathcal{L}_\mathrm{train}$ is usually non-convex in $\bm{\omega}$, with multiple local or global minima, solving $\min_{\bm{\omega}}\mathcal{L}_\mathrm{train}\mleft(\bm{\omega}\mright)$ may still lead to high generalization error $\mathcal{L}_\mathrm{test}\mleft(\bm{\omega}\mright)$. To alleviate this issue,~\citet{foret2021sharpnessaware} propose to regularize the training objective with the sharpness, defined as follows
\begin{boxdef}[Sharpness,~\citet{foret2021sharpnessaware}]
    For a given $\rho \geq 0$, the sharpness of $\mathcal{L}_\mathrm{train}$ at $\bm{\omega}$ writes
    \begin{equation}
        \label{eq:def_sharpness}
        s\mleft(\bm{\omega}, \rho\mright) \coloneqq \max_{\lVert \bm{\epsilon} \rVert_2 \leq \rho} \mathcal{L}_\mathrm{train}\mleft(\bm{\omega} + \bm{\epsilon}\mright) - \mathcal{L}_\mathrm{train}\mleft(\bm{\omega}\mright).
    \end{equation}
\end{boxdef}
\begin{rmk}[Interpretation of $\rho$]
\label{rmk:sam_rho}
    Instead of simply minimizing the training objective $\mathcal{L}_\mathrm{train}$, SAM searches for parameters $\bm{\omega}$ achieving both low training loss and low curvature in a ball $\mathcal{B}(\bm{\omega}, \rho)$. The hyperparameter $\rho \geq 0$ corresponds to the size of the neighborhood on which the parameters search is done. In particular, taking $\rho=0$ is equivalent to the usual minimization of $\mathcal{L}_\mathrm{train}$.
\end{rmk}
In particular, SAM incorporates sharpness in the learning objective, resulting in the problem of minimizing w.r.t $\bm{\omega}$
\begin{equation}
    \label{eq:sam_objective}
    \mathcal{L}_\mathrm{train}^\textrm{SAM}(\bm{\omega}) \coloneqq \underbrace{\max_{\lVert \bm{\epsilon} \rVert_2 \leq \rho} \mathcal{L}_\mathrm{train}\mleft(\bm{\omega} + \bm{\epsilon}\mright)}_{=\mathcal{L}_\mathrm{train}\mleft(\bm{\omega}\mright) + s\mleft(\bm{\omega}, \rho\mright)}\,.
\end{equation}

\paragraph{Gradient updates.} As the exact solution to the inner maximization in Eq.~\eqref{eq:sam_objective} is hard to compute, the authors of~\citep{foret2021sharpnessaware} approximate it with the following first-order Taylor expansion 
\begin{align}
\label{eq:first_order_delta}
\bm{\epsilon}^*\mleft(\bm{\omega}\mright) &\coloneqq \argmax_{\lVert \bm{\epsilon} \rVert_2 \leq \rho} \mathcal{L}_\mathrm{train}\mleft(\bm{\omega} + \bm{\epsilon}\mright) \notag \\
    &\approx \argmax_{\lVert \bm{\epsilon} \rVert_2 \leq \rho} \mathcal{L}_\mathrm{train}\mleft(\bm{\omega}\mright) + \bm{\epsilon}^\top \nabla \mathcal{L}_\mathrm{train}\mleft(\bm{\omega}\mright) \notag \\
    & = \argmax_{\lVert \bm{\epsilon} \rVert_2 \leq \rho} \bm{\epsilon}^\top \nabla \mathcal{L}_\mathrm{train}\mleft(\bm{\omega}\mright)\,,
\end{align}
where the solution of \eqref{eq:first_order_delta} writes $\bm{\hat{\epsilon}}\mleft(\bm{\omega}\mright) = \rho \frac{\nabla \mathcal{L}_\mathrm{train}\mleft(\bm{\omega}\mright)}{\lVert \nabla \mathcal{L}_\mathrm{train}\mleft(\bm{\omega}\mright) \rVert_2}$. It leads to the following gradient update
\begin{equation*}
    \bm{\omega}_{t+1} = \bm{\omega}_t - \eta \nabla \mathcal{L}_\mathrm{train}\mleft(\bm{\omega}_t +  \rho \frac{\nabla \mathcal{L}_\mathrm{train}\mleft(\bm{\omega}\mright)}{\lVert \nabla \mathcal{L}_\mathrm{train}\mleft(\bm{\omega}\mright) \rVert_2}\mright)\,,
\end{equation*}
where $\eta$ is the learning rate.

\section{Proofs}
\label{app:proofs}
\subsection{Notations}
To ease the readability of the proofs, we recall the following notations. We denote scalar values by regular letters (e.g., parameter $\lambda$), vectors by bold lowercase letters (e.g., vector $\mbf{x}$), and matrices by bold capital letters (e.g., matrix $\mbf{M}$). For a matrix $\mbf{M} \in \mathbb{R}^{n \times m}$, we denote by $\mbf{M}_{i}$ its $i$-th row, by $\mbf{M}_{\cdot, j}$ its $j$-th column, by $m_{ij}$ its entries and by $\mbf{M}^\top$ its transpose. We denote the trace of a matrix $\mbf{M}$ by $\tr{\mbf{M}}$, its rank by $\rk{\mbf{M}}$ and its Frobenius norm by $\lVert \mbf{M} \rVert_\mathrm{F}$. We denote $\bm{\sigma}\mleft(\mbf{M}\mright) \coloneqq\mleft(\sigma_1(\mbf{M}), \dots, \sigma_{\tilde{n}}(\mbf{M})\mright)$ the vector of singular values of $\mbf{M}$ in non-decreasing order, with $\tilde{n} = \min\{n,m\}$ and the specific notation $\sigma_\mathrm{min}(\mbf{M}), \sigma_\mathrm{max}(\mbf{M})$ for the minimum and maximum singular values, respectively. We denote by $\lVert\mbf{M}\rVert_* = \sum_{i=1}^{\tilde{n}} \sigma_i(\mbf{M})$ its nuclear norm and by $\lVert\mbf{M}\rVert_2 = \sigma_\mathrm{max}(\mbf{M})$ its spectral norm. When $\mbf{M}$ is square with $n=m$, we denote $\bm{\lambda}\mleft(\mbf{M}\mright) \coloneqq\mleft(\lambda_1(\mbf{M}), \dots, \lambda_n(\mbf{M})\mright)$ the vector of singular values of $\mbf{M}$ in non-decreasing order and the specific notation $\lambda_\mathrm{min}(\mbf{M}), \lambda_\mathrm{max}(\mbf{M})$ for the minimum and maximum singular values, respectively. For a vector $\mbf{x}$, its transpose writes $\mbf{x}^\top$ and its usual Euclidean norm writes $\lVert \mbf{x} \rVert$. The identity matrix of size $n\times n$ is denoted by $\mbf{I}_n$. 
The vector of size $n$ with each entry equal to $1$ is denoted by $\mathbbm{1}_n$. The notation $\mbf{M} \succcurlyeq \mbf{0}$ indicates that $\mbf{M}$ is positive semi-definite.

\subsection{Proof of Proposition~\ref{prop:optimal_transformer}}
\label{app:optimal_transformer}
We first recall the following technical lemmas.
\begin{boxlem}
\label{lem:full_rank}
    Let $\mbf{S} \in \mathbb{R}^{n \times m}$ and $\mbf{B} \in \mathbb{R}^{m \times m}$. If $\mbf{B}$ has full rank, then
    \begin{equation*}
        \rk{\mbf{S}\mbf{B}} = \rk{\mbf{B}\mbf{S}} = \rk{\mbf{S}}.
    \end{equation*}
\end{boxlem}
\begin{proof}
Let $\mbf{F}_1 \coloneq \{\mbf{S}\mbf{u} | \mbf{u} \in \mathbb{R}^m\} \subset \mathbb{R}^n$ and $\mbf{F}_2 \coloneq \{\mleft(\mbf{S}\mbf{B}\mright)\mbf{u} | \mbf{u} \in \mathbb{R}^m\} \subset \mathbb{R}^n $ be the vector spaces generated by the columns of $\mbf{S}$ and $\mbf{S}\mbf{B}$ respectively. By definition, the rank of a matrix is the dimension of the vector space generated by its columns (equivalently by its rows). We will show that $\mbf{F}_1$ and $\mbf{F}_2$ coincides. Let $\mbf{v} \in \mbf{F}_1$, i.e., there exists $\mbf{u} \in \mathbb{R}^m$ such that $\mbf{v} = \mbf{S}\mbf{u}$. As $\mbf{B}$ is full rank, the operator $\mbf{x} \to \mbf{B}\mbf{x}$ is bijective. It follows that there always exists some $\mbf{z} \in \mathbb{R}^m$ such that $\mbf{u} = \mbf{B}\mbf{z}$. Then, we have
\begin{equation*}
    \mbf{v} = \mbf{S}\mbf{u} = \mbf{S}\mleft(\mbf{B}\mbf{z}\mright) = \mleft(\mbf{S}\mbf{B}\mright)\mbf{z},
\end{equation*}
which means that $\mbf{v} \in \mbf{F}_2$. As $\mbf{v}$ was taken arbitrarily in $\mbf{F}_1$, we have proved that $\mbf{F}_1 \subset \mbf{F}_2$. Conversely, consider $\mbf{y} \in \mbf{F}_2$, i.e., we can write $\mbf{y} = \mleft(\mbf{S}\mbf{B}\mright)\mbf{z}$ for some $\mbf{z} \in \mathbb{R}^m$. It can then be seen that
\begin{equation*}
    \mbf{y} = \mleft(\mbf{S}\mbf{B}\mright)\mbf{z} = \mbf{S}\mleft(\mbf{B}\mbf{z}\mright),
\end{equation*}
which means that $\mbf{y} \in \mbf{F}_1$. Again, as $\mbf{y}$ was taken arbitrarily, we have proved that $\mbf{F}_1 \subset \mbf{F}_2$. In the end, we demonstrated that $\mbf{F}_1$ and $\mbf{F}_2$ coincide, hence they have the same dimension. By definition of the rank, $\mbf{S}$ and $\mbf{S}\mbf{B}$ have the same rank. Similar arguments can be used to show that $\mbf{S}$ and $\mbf{B}\mbf{S}$ have the same rank, which concludes the proof.
\end{proof}

The next lemma is a well-known result in matrix analysis and can be found in \citet[Theorem 4.4.5]{Horn_Johnson_1991}. For the sake of self-consistency, we recall it below along with a sketch of the original proof. 
\begin{boxlem}{\citep[see][Theorem 4.4.5, p. 281]{Horn_Johnson_1991}.}
\label{lem:horn_johnson}
    Let $\mbf{S} \in \mathbb{R}^{n \times m}, \mbf{B} = \mathbb{R}^{p \times q}$ and $\mbf{C} \in \mathbb{R}^{n \times q}$. There exists matrices $\mbf{Y} \in \mathbb{R}^{m \times q}$ and $\mbf{Z} \in \mathbb{R}^{n \times p}$ such that $\mbf{S}\mbf{Y} - \mbf{Z}\mbf{B} = \mbf{C}$ if, and only if,
    \begin{equation*}
        \rk{
        \begin{bmatrix}
            \mbf{S} & \mbf{C} \\
            \mbf{0} & \mbf{B}
        \end{bmatrix}
        } = \rk{
        \begin{bmatrix}
            \mbf{S} & \mbf{0} \\
            \mbf{0} & \mbf{B}
        \end{bmatrix}
        }.
    \end{equation*}
\end{boxlem}
\begin{proof}
    Assume that there exists $\mbf{Y} \in \mathbb{R}^{m \times q}$ and $\mbf{Z} \in \mathbb{R}^{n \times p}$ such that $\mbf{S}\mbf{Y} - \mbf{Z}\mbf{B} = \mbf{C}$. Recall that the following equality holds
    \begin{equation}
    \label{eq:matrix_equality}
        \begin{bmatrix}
            \mbf{S} & \mbf{S}\mbf{Y} - \mbf{Z}\mbf{B} \\
            \mbf{0} & \mbf{B}
        \end{bmatrix} = 
        \begin{bmatrix}
            \mbf{I}_m & -\mbf{Y} \\
            \mbf{0} & \mbf{I}_q
        \end{bmatrix}
            \begin{bmatrix}
        \mbf{S} & \mbf{0} \\
        \mbf{0} & \mbf{B}
    \end{bmatrix}
            \begin{bmatrix}
        \mbf{I}_n & \mbf{Z} \\
        \mbf{0} & \mbf{I}_p
    \end{bmatrix}.
    \end{equation}
Using Lemma~\ref{lem:full_rank} on the right-hand-side of Eq.~\eqref{eq:matrix_equality}, we obtain
\begin{equation*}
        \rk{
        \begin{bmatrix}
            \mbf{S} & \mbf{S}\mbf{Y} - \mbf{Z}\mbf{B} \\
            \mbf{0} & \mbf{B}
        \end{bmatrix}
        } = \rk{
        \begin{bmatrix}
            \mbf{S} & \mbf{0} \\
            \mbf{0} & \mbf{B}
        \end{bmatrix}
        }.
\end{equation*}
Using $\mbf{S}\mbf{Y} - \mbf{Z}\mbf{B} = \mbf{C}$ concludes the proof for the first implication of the equivalence.\\
To prove the opposite direction, the authors of \citet{Horn_Johnson_1991} assume that 
    \begin{equation*}
        \rk{
        \begin{bmatrix}
            \mbf{S} & \mbf{C} \\
            \mbf{0} & \mbf{B}
        \end{bmatrix}
        } = \rk{
        \begin{bmatrix}
            \mbf{S} & \mbf{0} \\
            \mbf{0} & \mbf{B}
        \end{bmatrix}
        }.
    \end{equation*}
Since two matrices have the same rank if, and only if, they are equivalent, we know that there exists $\mbf{Q} \in \mathbb{R}^{(n+p) \times (n+p)}, \mbf{U} \in \mathbb{R}^{(m+q) \times (m+q)}$ non-singular such that 
\begin{equation}
\label{eq:eq_matrix}
            \begin{bmatrix}
            \mbf{S} & \mbf{C} \\
            \mbf{0} & \mbf{B}
        \end{bmatrix} = 
        \mbf{Q}
        \begin{bmatrix}
            \mbf{S} & \mbf{0} \\
            \mbf{0} & \mbf{B}
        \end{bmatrix}
        \mbf{U}.
\end{equation}
The rest of the proof in \citet{Horn_Johnson_1991} is constructive and relies on Eq.~\eqref{eq:eq_matrix} to exhibit $\mbf{Y} \in \mathbb{R}^{m \times q}$ and $\mbf{Z} \in \mathbb{R}^{n \times p}$ such that $\mbf{S}\mbf{Y} - \mbf{Z}\mbf{B} = \mbf{C}$. This concludes the proof of the equivalence.
\end{proof}
We now proceed to the proof of Proposition~\ref{prop:optimal_transformer}.
\begin{proof}
    Applying Lemma~\ref{lem:horn_johnson} with $\mbf{S}=\mbf{P}$, $\mbf{B}=\mbf{0}$, $\mbf{C} = \mbf{X}\mbf{W}_\mathrm{toy}$ and $\mbf{W}$ in the role of $\mbf{Y}$ ensures that there exists $\mbf{W} \in \mathbb{R}^{L \times H}$ such that $\mbf{P}\mbf{W} = \mbf{X}\mbf{W}_\mathrm{toy}$ if and only if $\rk{\left[\mbf{P}\quad \mbf{X}\mbf{W}_\mathrm{toy}\right]} = \rk{\mbf{P}}$, which concludes the proof.
\end{proof}

\subsection{Proof of Proposition~\ref{thm:upper_bound_nuclear_norm}}
\label{app:upper_bound_nuclear_norm}
We first prove the following technical lemmas. While these lemmas are commonly used and, for most of them, straightforward to prove, they are very useful to demonstrate Proposition~\ref{thm:upper_bound_nuclear_norm}.
\begin{boxlem}[Trace of a product of matrix]
\label{lem:ineq_trace}
    Let $\mbf{S}, \mbf{B} \in \mathbb{R}^{n \times n}$ be \textit{\textbf{symmetric}} matrices with $\mbf{B}$ positive semi-definite. We have
    \begin{equation*}
\lambda_\mathrm{min}\mleft(\mbf{S}\mright)\tr{\mbf{B}} \leq \tr{\mbf{S}\mbf{B}} \leq \lambda_\mathrm{max}\mleft(\mbf{S}\mright)\tr{\mbf{B}}.
    \end{equation*}
\end{boxlem}
\begin{proof}
 The spectral theorem ensures the existence of $\mbf{P} \in \mathbb{R}^{n \times n}$ orthogonal, i.e., $\mbf{P}^\top \mbf{P} = \mbf{P}\mbf{P}^\top = \mbf{I}_n$, and $\bm{\Lambda} \in \mathbb{R}^{n \times n}$ diagonal with the eigenvalues of $\mbf{S}$ as entries such that $\mbf{S} = \mbf{P}\bm{\Lambda}\mbf{P}^\top$. Benefiting from the properties of the trace operator, we have
 \begin{align*}
     \tr{\mbf{S}\mbf{B}} &= \tr{\mbf{I}_n\mbf{S}\mbf{B}} \\
     &= \tr{\underbrace{\mbf{P}\mbf{P}^\top}_{=\mbf{I}_n}\mbf{S}\mbf{B}} \tag{orthogonality of $\mbf{P}$} \\
     &= \tr{\mbf{P}^\top \mbf{S}\mbf{B}\mbf{P}} \tag{cyclic property of trace} \\
     &= \tr{\mbf{P}^\top\mbf{P}\bm{\Lambda}\mbf{P}^\top\mbf{B}\mbf{P}} \tag{Spectral theorem} \\
     &= \tr{\underbrace{\mbf{P}^\top\mbf{P}}_{=\mbf{I}_n}\bm{\Lambda}\mbf{P}^\top\mbf{B}\mbf{P}} \tag{orthogonality of $\mbf{P}$} \\
     &= \tr{\bm{\Lambda}\mbf{P}^\top\mbf{B}\mbf{P}}. \\
 \end{align*}
We introduce $\tilde{\mbf{B}} = \mbf{P}^\top\mbf{B}\mbf{P} = [\tilde{b}_{ij}]_{ij}$. It follows from the definition of $\bm{\Lambda}$ that
\begin{equation}
\label{eq:trace_sum_b_tilde}
    \tr{\mbf{S}\mbf{B}} = \tr{\bm{\Lambda}\mbf{P}^\top\mbf{B}\mbf{P}} = \tr{\bm{\Lambda} \tilde{\mbf{B}}} = \sum_i \lambda_i\mleft(\mbf{S}\mright) \tilde{b}_{ii}.
\end{equation}
We would like to write the $\tilde{b}_{ij}$ with respect to the $p_{ij}, b_{ij}$ the elements of $\mbf{P}, \mbf{B}$, respectively. As $\mbf{P}$ is orthogonal, we know that its columns $(\mbf{e}_i)_{i=0}^n$ form an orthonormal basis of $\mathbb{R}^n$. Hence, the entry $(i,j)$ of $\bm{\Lambda}\mbf{P}^\top\mbf{B}\mbf{P}$, writes as follows:
\begin{align*}
    \tilde{b}_{ij} &= \sum_{kl}p_{ki}b_{ij}p_{jk} \\ 
    &= \sum_{k}p_{ki}\underbrace{\mleft(\sum_lb_{ij}p_{jk}\mright)}_{\mleft[\mbf{B}\mbf{e}_j\mright]_k} \\ 
    &= \sum_{k}p_{ki}\mleft[\mbf{B}\mbf{e}_j\mright]_k \\
    &= e_i^\top \mbf{B}\mbf{e}_j \geq 0 \tag{$\mbf{B} \succcurlyeq \mbf{0}$}.
\end{align*}
Hence, as $\mbf{B}$ is positive semi-definite, the $\tilde{b}_{ij}$ are nonnegative. It follows that
\begin{equation}
\label{eq:ineq_sum_b_tilde}
    \lambda_\mathrm{min}\mleft(\mbf{S}\mright) \sum_i \tilde{b}_{ii} \leq \sum_i \lambda_i\mleft(\mbf{S}\mright) \underbrace{\tilde{b}_{ii}}_{\geq 0} \leq \lambda_\mathrm{max}\mleft(\mbf{S}\mright) \sum_i \tilde{b}_{ii}.
\end{equation}
Moreover, using the definition of $\tilde{\mbf{B}}$, the orthogonality of $\mbf{P}$ and the cyclic property of the trace operation, we have
\begin{equation*}
    \sum_i \tilde{b}_{ii} = \tr{\tilde{\mbf{B}}} = \tr{\mbf{P}^\top \mbf{B} \mbf{P}} = \tr{\underbrace{\mbf{P}\mbf{P}^\top}_{=\mbf{I}_n} \mbf{B}} = \tr{\mbf{B}}.
\end{equation*}
Combining this last equality with Eq.~\eqref{eq:trace_sum_b_tilde} and Eq.~\eqref{eq:ineq_sum_b_tilde} concludes the proof, i.e., 
\begin{equation}
    \lambda_\mathrm{min}\mleft(\mbf{S}\mright) \tr{\mbf{B}} \leq \tr{\mbf{S}\mbf{B}} \leq \lambda_\mathrm{max}\mleft(\mbf{S}\mright) \tr{\mbf{B}}.
\end{equation}
\end{proof}

\begin{boxlem}[Power of symmetric matrices]
\label{lem:power_n_psd}
    Let $\mbf{S} \in \mathbb{R}^{n \times n}$ be symmetric. The spectral theorem ensures the existence of $\mbf{P} \in \mathbb{R}^{n \times n}$ orthogonal, i.e., $\mbf{P}^\top \mbf{P} = \mbf{P}\mbf{P}^\top = \mbf{I}_n$, and $\bm{\Lambda} \in \mathbb{R}^{n \times n}$ diagonal with the eigenvalues of $\mbf{S}$ as entries such that $\mbf{S} = \mbf{P}\bm{\Lambda}\mbf{P}^\top$. For any integer $n \geq 1$, we have
    \begin{equation*}
\mbf{S}^n = \mbf{P}\bm{\Lambda}^n\mbf{P}^\top.
    \end{equation*}
In particular, the eigenvalues of $\mbf{S}^n$ are equal to the eigenvalues of $\mbf{S}$ to the power of $n$.
\end{boxlem}
\begin{proof}
Let $n \geq 1$ be an integer. We have
\begin{align*}
    \mbf{S}^n &= \mleft( \mbf{P}\bm{\Lambda}\mbf{P}^\top\mright)^n \\
    &= \underbrace{\mbf{P}\bm{\Lambda}\mbf{P}^\top \times \mbf{P}\bm{\Lambda}\mbf{P}^\top \times \dots \times \mbf{P}\bm{\Lambda}\mbf{P}^\top \times \mbf{P}\bm{\Lambda}\mbf{P}^\top}_{\times n} \\ 
    &= \underbrace{\mbf{P}\bm{\Lambda}\times\bm{\Lambda}\mbf{P}^\top\dots \mbf{P}\bm{\Lambda}\times\bm{\Lambda}\mbf{P}^\top}_{\times n} \tag{orthogonality of $\mbf{P}$} \\
    &= \mbf{P} \underbrace{\bm{\Lambda}\times\bm{\Lambda} \times \dots \times \bm{\Lambda}\times\bm{\Lambda}}_{\times n}\mbf{P}^\top \tag{orthogonality of $\mbf{P}$} \\
    &= \mbf{P}\bm{\Lambda}^n\mbf{P}^\top.
\end{align*}
The diagonality of $\bm{\Lambda}$ suffices to deduct the remark on the eigenvalues of $\mbf{S}^n$.
\end{proof}

\begin{boxlem}[Case of equality between eigenvalues and singular values]
\label{lem:eigenvalues_singular_values_psd}
    Let $\mbf{S} \in \mathbb{R}^{n \times n}$ be symmetric and positive semi-definite. Then the $i$-th eigenvalue and the $i$-th singular value of $\mbf{S}$ are equal, i.e., for all $i \in \llbracket 1,n \rrbracket$, we have 
    \begin{equation*}
\lambda_i\mleft(\mbf{S}\mright) = \sigma_i\mleft(\mbf{S}\mright).
    \end{equation*}
\end{boxlem}
\begin{proof}
Let $i \in \llbracket 1,n \rrbracket$. By definition of singular value, we have
\begin{align*}
    \sigma_i\mleft(\mbf{S}\mright) &\coloneqq \sqrt{\lambda_i\mleft(\mbf{S}^\top \mbf{S}\mright)} \\
    &= \sqrt{\lambda_i\mleft(\mbf{S}^2\mright)} \tag{$\mbf{S}$ is symmetric} \\
    &= \sqrt{\lambda_i\mleft(\mbf{S}\mright)^2} \tag{Lemma~\ref{lem:power_n_psd}} \\
    &= \lvert \lambda_i\mleft(\mbf{S}\mright) \rvert \\
    &= \lambda_i\mleft(\mbf{S}\mright) \tag{$\mbf{S} \succcurlyeq \mbf{0}$}. \\
\end{align*}
\end{proof}

\begin{boxlem}
\label{lem:psd}
    Let $\mbf{X} \in \mathbb{R}^{D \times L}$ be an input sequence and $\mbf{S} \in \mathbb{R}^{L \times L}$ be a positive semi-definite matrix. Then, $\mbf{X}\mbf{S}\mbf{X}^\top$ is positive semi-definite.
\end{boxlem}
\begin{proof}
    It is clear that $\mbf{X}\mbf{S}\mbf{X}^\top \in \mathbb{R}^{L \times L}$ is symmetric. Let $\mbf{u} \in \mathbb{R}^L$. We have:
    \begin{align*}
        \mbf{u}^\top \mbf{X}\mbf{S}\mbf{X}^\top \mbf{u} &= \mleft(\mbf{X}^\top \mbf{u}\mright)^\top \mbf{S} \mleft(\mbf{X}^\top\mbf{u}\mright) \geq 0 \tag{$\mbf{S} \succcurlyeq \mbf{0}$}.
    \end{align*}
    As $\mbf{u}$ was arbitrarily chosen, we have proved that $\mbf{X}\mbf{S}\mbf{X}^\top$ is positive semi-definite.
\end{proof}
We now proceed to the proof of Theorem~\ref{thm:upper_bound_nuclear_norm}.
\begin{proof}
We recall that $\mbf{W}_Q\mbf{W}_K^\top$ is symmetric and positive semi-definite, we have
\begin{align*}
       \lVert \mbf{X}\mbf{W}_Q\mbf{W}_K^\top\mbf{X}^\top\rVert_*
       &= \tr{\sqrt{\mleft(\mbf{X}\mbf{W}_Q\mbf{W}_K^\top\mbf{X}^\top\mright)^\top \mbf{X}\mbf{W}_Q\mbf{W}_K^\top\mbf{X}^\top}} \\
       &= \tr{\sqrt{\mbf{X}\mbf{W}_K\mbf{W}_Q^\top\mbf{X}^\top\mbf{X}\mbf{W}_Q\mbf{W}_K^\top\mbf{X}^\top}} \\
       &= \tr{\sqrt{\mbf{X}\mbf{W}_Q\mbf{W}_K^\top\mbf{X}^\top\mbf{X}\mbf{W}_Q\mbf{W}_K^\top\mbf{X}^\top}} \tag{symmetry} \\
       &= \tr{\sqrt{\mleft(\mbf{X}\mbf{W}_Q\mbf{W}_K^\top\mbf{X}^\top\mright)^2}} \\
       &= \tr{\mbf{X}\mbf{W}_Q\mbf{W}_K^\top\mbf{X}^\top} \tag{Lemma~\ref{lem:psd} with $\mbf{S} = \mbf{W}_Q\mbf{W}_K^\top$} \\
       &= \tr{\mbf{X}^\top\mbf{X}\mbf{W}_Q\mbf{W}_K^\top} \tag{cyclic property of the trace}.
   \end{align*}
Using the fact that $\mbf{X}^\top\mbf{X}$ is positive semi-definite (Lemma~\ref{lem:psd} with $\mbf{S} = \mbf{I}_L$), and that $\mbf{W}_Q\mbf{W}_K^\top$ is symmetric, Lemma~\ref{lem:ineq_trace} can be applied with $\mbf{M} = \mbf{W}_Q\mbf{W}_K^\top$ and $\mbf{B} = \mbf{X}^\top\mbf{X}$. It leads to:
\begin{align*}
    \lVert \mbf{X}\mbf{W}_Q\mbf{W}_K^\top\mbf{X}^\top\rVert_* = \tr{\mbf{X}^\top\mbf{X}\mbf{W}_Q\mbf{W}_K^\top} & \leq \lambda_\mathrm{max}\mleft(\mbf{W}_Q\mbf{W}_K^\top \mright)\tr{\mbf{X}^\top\mbf{X}}\,.
    \tag{Lemma~\ref{lem:ineq_trace}} 
\end{align*}
As $\mbf{W}_Q\mbf{W}_K^\top$ is positive semi-definite, Lemma~\ref{lem:eigenvalues_singular_values_psd} ensure 
\begin{equation*}
\lambda_\mathrm{max}\mleft(\mbf{W}_Q\mbf{W}_K^\top \mright) = \sigma_\mathrm{max}\mleft(\mbf{W}_Q\mbf{W}_K^\top \mright) = \lVert \mbf{W}_Q\mbf{W}_K^\top\rVert_2
\end{equation*}
by definition of the spectral norm $\lVert \cdot \rVert_2$. Recalling that by definition, $\tr{\mbf{X}^\top \mbf{X}} = \lVert \mbf{X}\rVert_\mathrm{F}^2$ concludes the proof, i.e.,
\begin{equation*}
    \lVert \mbf{X}\mbf{W}_Q\mbf{W}_K^\top\mbf{X}^\top\rVert_* \leq \lVert \mbf{W}_Q\mbf{W}_K^\top\rVert_2 \lVert \mbf{X}\rVert_\mathrm{F}^2.
\end{equation*}
\end{proof}

\subsection{Proof of Proposition~\ref{prop:closed_form}}
\label{app:closed_form}
\begin{proof}
Let $k \in \llbracket 1,K \rrbracket$ and $t \in \llbracket 1,H \rrbracket$. We have
\begin{align*}
    \mbf{\hat{Y}}_{kt} &= \sqrt{\hat{\sigma}^2\mleft[\mbf{X}_{k}\mright] + \varepsilon} \cdot \mleft(\frac{\mbf{\tilde{y}}_{kt} - \bm{\beta}_k}{\bm{\gamma}_k}\mright) + \hat{\mu}\mleft[\mbf{X}_{k}\mright], \tag{from \eqref{eq:revin_second_step}}\\
    &= \sqrt{\hat{\sigma}^2\mleft[\mbf{x}_{k}\mright] + \varepsilon} \cdot \mleft(\frac{\sum_{j=1}^L \mbf{\tilde{X}}_{kj} \mbf{W}_{jt} - \bm{\beta}_k}{\bm{\gamma}_k}\mright) + \hat{\mu}\mleft[\mbf{X}_{k}\mright], \tag{from \eqref{eq:linear_model}} \\
    &= \frac{\sqrt{\hat{\sigma}^2\mleft[\mbf{X}_{k}\mright] + \varepsilon}}{\bm{\gamma}_k} \cdot \sum_{j=1}^L \mbf{\tilde{X}}_{kj} \mbf{W}_{jt} - \frac{\bm{\beta}_k}{\bm{\gamma}_k} \sqrt{\hat{\sigma}^2\mleft[\mbf{X}_{k}\mright] + \varepsilon} + \hat{\mu}\mleft[\mbf{X}_{k}\mright] \\
    &= \frac{\sqrt{\hat{\sigma}^2\mleft[\mbf{X}_{k}\mright] + \varepsilon}}{\bm{\gamma}_k} \cdot \sum_{j=1}^L \mleft(\bm{\gamma}_k\mleft( \frac{\mbf{X}_{kj} - \hat{\mu}\mleft[\mbf{x}_{k}\mright]}{\sqrt{\hat{\sigma}^2\mleft[\mbf{X}_{k}\mright] + \varepsilon}}\mright) + \bm{\beta}_k \mright)\mbf{W}_{jt} - \frac{\bm{\beta}_k}{\bm{\gamma}_k} \sqrt{\hat{\sigma}^2\mleft[\mbf{x}_{k}\mright] + \varepsilon} + \hat{\mu}\mleft[\mbf{X}_{k}\mright] , \tag{from \eqref{eq:revin_first_step}} \\
    &= \sum_{j=1}^L \mleft(\mbf{X}_{kj} - \hat{\mu}\mleft[\mbf{X}_{k}\mright]\mright)\mbf{W}_{jt} + \frac{\bm{\beta}_k}{\bm{\gamma}_k} \sqrt{\hat{\sigma}^2\mleft[\mbf{X}_{k}\mright] + \varepsilon} \mleft(\sum_{j=1}^L\mbf{W}_{jt}-1\mright) + \hat{\mu}\mleft[\mbf{X}_{k}\mright] \\
    &= \hat{\mu}\mleft[\mbf{X}_{k}\mright]   + \sum_{j=1}^L \mleft(\mbf{X}_{kj} - \hat{\mu}\mleft[\mbf{X}_{k}\mright]\mright)\mbf{W}_{jt}  - \frac{\bm{\beta}_k}{\bm{\gamma}_k} \sqrt{\hat{\sigma}^2\mleft[\mbf{X}_{k}\mright] + \varepsilon}\mleft(1-\sum_{j=1}^L\mbf{W}_{jt}\mright).
\end{align*}
\end{proof}

\subsection{Matrix formulation of $\mbf{\hat{Y}}$ in Eq.~\eqref{eq:matrix_formulation}}
\label{app:matrix_formulation}
\begin{proof}
Let $k \in \llbracket 1,K \rrbracket$ and $t \in \llbracket 1,H \rrbracket$. From Proposition~\ref{prop:closed_form}, we have
\begin{align*}
    \mbf{\hat{Y}}_{kt} &= \hat{\mu}\mleft[\mbf{X}_{k}\mright]   + \sum_{j=1}^L \mleft(\mbf{X}_{kj} - \hat{\mu}\mleft[\mbf{X}_{k}\mright]\mright)\mbf{W}_{jt} - \frac{\bm{\beta}_k}{\bm{\gamma}_k} \sqrt{\hat{\sigma}^2\mleft[\mbf{X}_{k}\mright] + \varepsilon} \mleft(1-\sum_{j=1}^L\mbf{W}_{jt}\mright) \\
    &= \sum_{j=1}^L\mbf{X}_{kj}\mbf{W}_{jt} + \mleft(\hat{\mu}\mleft[\mbf{X}_{k}\mright] - \frac{\bm{\beta}_k}{\bm{\gamma}_k} \sqrt{\hat{\sigma}^2\mleft[\mbf{X}_{k}\mright] + \varepsilon}\mright) \cdot  \mleft(1-\sum_{j=1}^L\mbf{W}_{jt}\mright).
\end{align*}
Gathering in matrix formulation concludes the proof.
\end{proof}

\end{document}